\documentclass{article}

\usepackage[preprint]{neurips_2026}

\usepackage[utf8]{inputenc} 
\usepackage[T1]{fontenc}    
\usepackage{hyperref}       
\usepackage{url}            
\usepackage{booktabs}       
\usepackage{amsfonts}       
\usepackage{nicefrac}       
\usepackage{microtype}      
\usepackage{xcolor}         

\usepackage{etoc}           
\usepackage{tocloft}        

\newcommand{\ie}{\textit{i.e.}}
\newcommand{\eg}{\textit{e.g.}}
\newcommand{\vs}{\textit{v.s.}}

\usepackage{tikz}
\usepackage{pgfplots}
\usepackage{subcaption}
\usepackage{xcolor}

\usepackage{graphicx}
\usepackage{amsmath}
\usepackage{cleveref}
\usepackage{amssymb}
\usepackage{booktabs}
\usepackage{tikz}
\usetikzlibrary{spy}
\usepackage{multirow}
\usepackage{subcaption}
\usepackage{overpic}
\usepackage{wrapfig} 

\usepackage{colortbl}  %
\usepackage{xcolor}    %
\definecolor{color3}{rgb}{0.95,0.95,0.95}

\newcommand{\myname}[0]{$\alpha$Depth} %

\title{$\alpha$Depth: Learning Scene-Level Soft Boundary Decomposition for Stereo Conversion}

\author{%
  Xiang~Zhang$^{1,2}$, Yang~Zhang$^{2}$, Lukas~Mehl$^{2}$, \\ \textbf{Karlis~Martins~Briedis}$^{2}$, 
  \textbf{Markus~Gross}$^{1,2}$,  \textbf{Christopher~Schroers}$^{2}$ \\
  $^{1}$ETH Zürich,\quad $^{2}$DisneyResearch|Studios\\
}

\begin{document}

\maketitle

\begin{figure}[th]
    \centering
    \includegraphics[width=\linewidth]{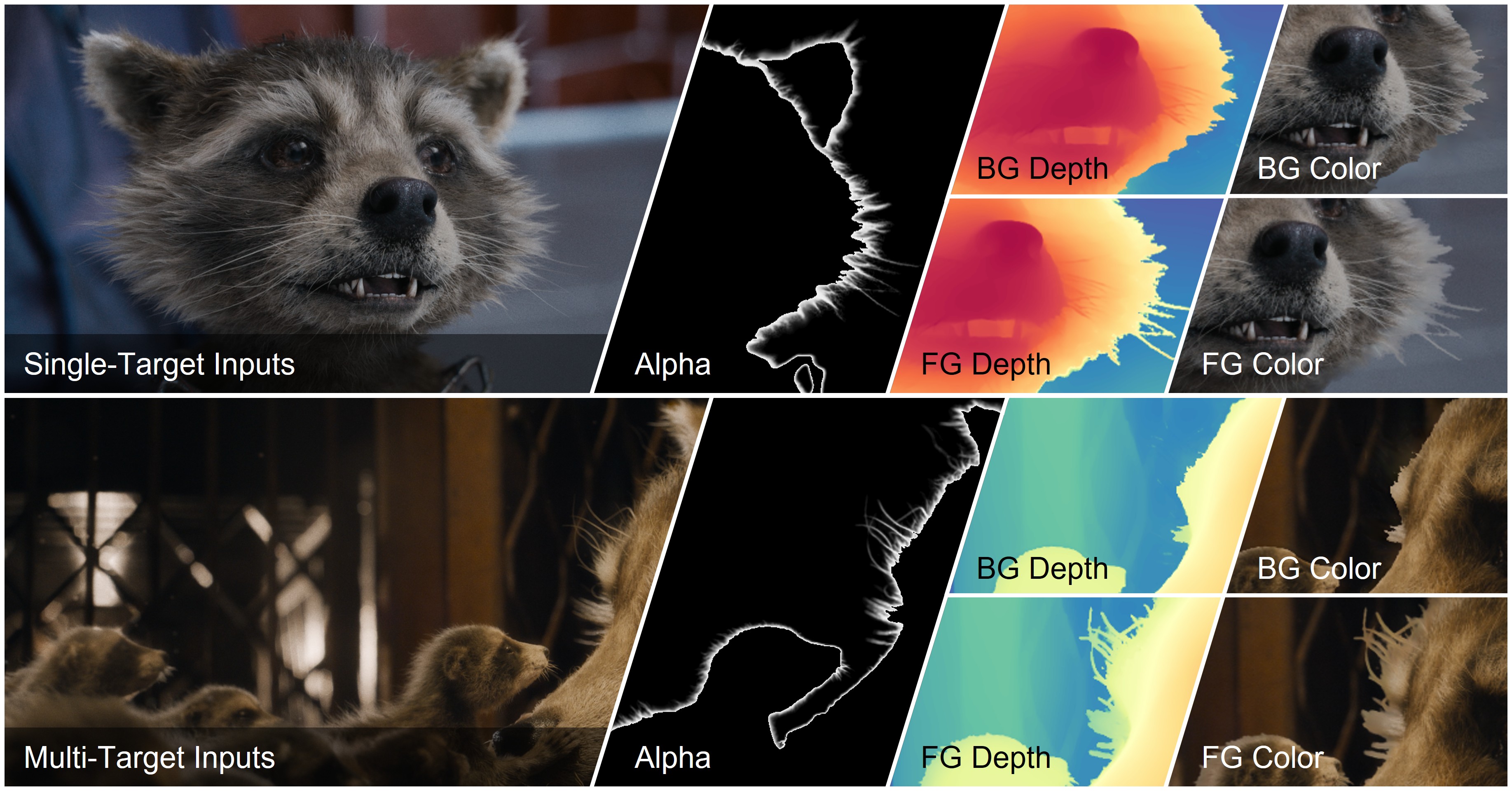}
    \caption{\textbf{Layered \myname\ Representation}. We introduce \myname\ to decompose soft boundaries (\eg, hair, thin structures, and defocus blur) for high-fidelity stereo conversion. 
    Given an image and its depth map as inputs, our approach estimates layered information, \ie, alpha, foreground/background (FG/BG) colors and depths, at \textit{local soft boundaries} (see non-zero alpha regions), enabling scene-level inference of multiple/overlapping targets in a single forward pass without user intervention.}
    \label{fig:main-teaser}
\end{figure}
\begin{abstract}

    Accurately modeling soft boundaries, \eg, hair and defocus blur, is a fundamental challenge in stereo conversion due to the ambiguous blending of foreground and background. Existing depth models primarily predict single-layer depth, leading to ambiguity in depth correspondence at soft boundaries. While matting techniques can capture opacity for layered modeling, they often struggle in complex scenes with multiple targets and usually require user intervention. This paper introduces \textbf{\myname}, a layered representation that decomposes soft boundaries for high-fidelity stereo conversion. Specifically, we first resolve mixed color and depth ambiguity by estimating layered color and depth values at soft boundaries. Considering complex multi-target scenes, we design a Circular Alpha Representation (CAR) that shifts the paradigm from global target extraction to local boundary decomposition. Unlike prior matting methods restricted to a single foreground/background, CAR enables efficient scene-level inference without manual guidance. Extensive evaluations demonstrate that {\myname} achieves state-of-the-art performance in stereo conversion, eliminating background bleeding and structural distortions at soft boundaries.

\end{abstract}

\begin{figure}[t]
    \centering
    \includegraphics[width=\linewidth]{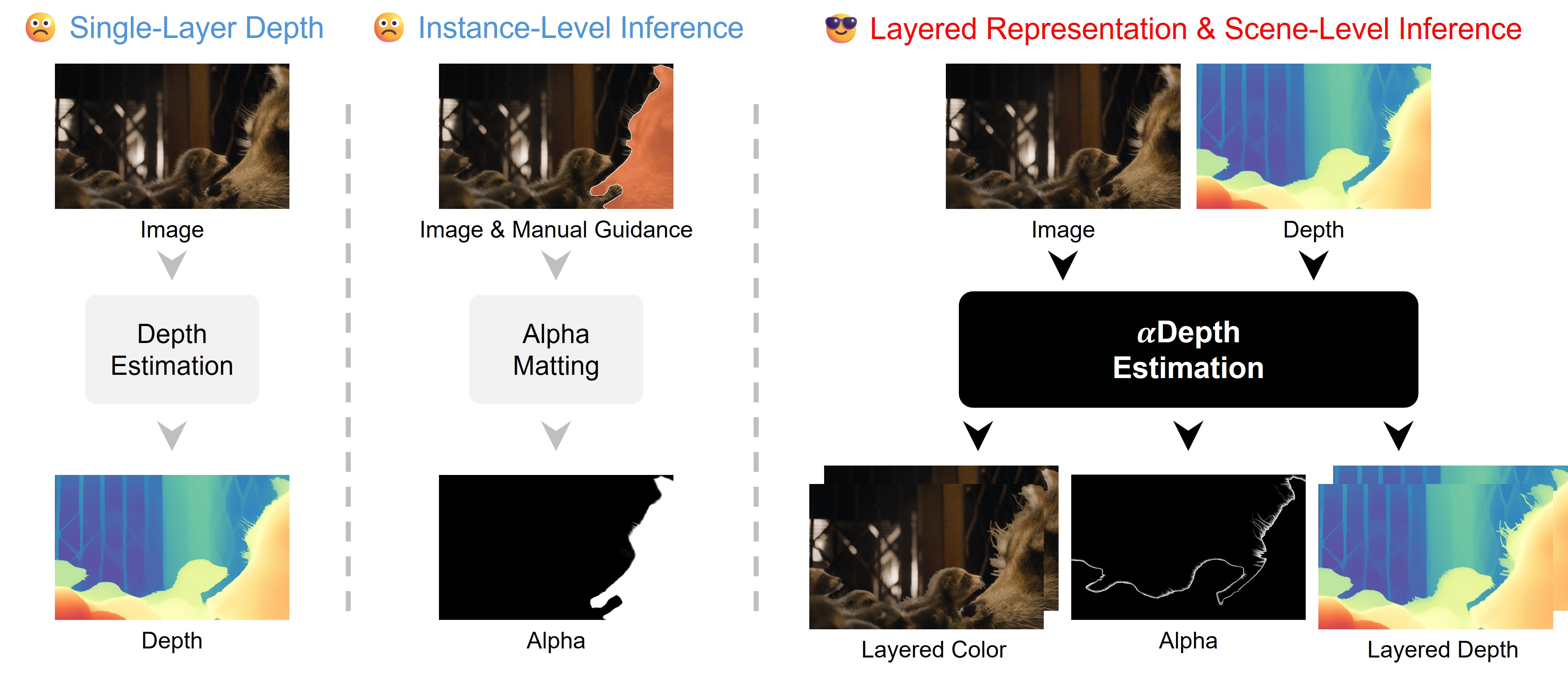}
    \caption{\textbf{Comparison with existing paradigms.} Depth estimation models typically assign a single depth value per pixel, struggling with mixed colors at soft boundaries and suffering from depth ambiguity. While conventional matting approaches extract instance-level soft boundaries, they usually require manual guidance (\eg, trimaps). In contrast, our layered \myname\ representation enables automatic scene-level decomposition. Given an image and its depth map, \myname\ explicitly decomposes soft boundaries into layered color, alpha, and layered depth in a single forward pass.}
    \label{fig:main-task-comparison}
\end{figure}
\section{Introduction}

By lifting 2D images into 3D content, stereo conversion is critical for various immersive applications, including Virtual/Augmented Reality (VR/AR) and movie production~\cite{mehl2024stereoconv,hairguard,yu2024viewcrafter,gao2024cat3d}. Recent methods have demonstrated remarkable progress in stereo conversion by leveraging foundation diffusion models to synthesize realistic novel views from monocular inputs~\cite{mono2stereo,stereodiffusion,shen2025stereopilot}. For instance, Eye2eye \cite{geyer2025eye2eye} effectively utilizes diffusion priors to simulate complex view-dependent effects such as specular reflections in stereo conversion. Approaches like SplatDiff \cite{splatdiff} and Elastic3D \cite{metzger2025elastic3d} have focused on improving the overall visual fidelity and structural consistency of the generated stereo pairs. Despite these advances, generating high-quality stereo content from monocular inputs remains challenging, particularly when dealing with intricate details in complex scenes.

\par 
A key challenge in stereo conversion is the accurate handling of \textit{soft boundaries}, which naturally occur across diverse subjects (\eg, humans, animals, or computer-generated characters as shown in Fig.~\ref{fig:main-teaser}) and camera effects (\eg, defocus blur). 
At soft boundaries, foreground and background colors inherently mix within a single pixel, creating regions of partial transparency and resulting in ambiguous depth correspondence. Consequently, previous stereo conversion methods often struggle to recover soft boundary details at high fidelity and produce stereo results with visual artifacts, such as halo effects, background bleeding, or unnatural floating textures around object silhouettes~\cite{hairguard}.

\par 
To achieve high-quality soft boundary recovery in stereo conversion, two fundamental challenges exist: \textbf{(i) Mixed Foreground and Background}: Many stereo conversion pipelines rely on monocular depth estimation for view synthesis~\cite{mono2stereo,shvetsova2025m2svid,splatdiff,stereocrafter}. However, conventional depth estimators typically assign only a single depth value per pixel~\cite{yang2024depthanything,depthpro,zhang2024betterdepth,ke2024marigold}, failing to model the layered characteristics of soft boundaries (Fig.~\ref{fig:main-task-comparison}). Although the recent work HairGuard~\cite{hairguard} better captures soft boundary structures via depth refinement, it still suffers from depth ambiguity due to the single-layer depth representation. \textbf{(ii) Complex Scenes with Multiple Targets}: Alpha matting techniques can extract alpha mattes for soft boundary decomposition~\cite{yao2024vitmatte,li2024mattinganything,kim2025zim,yang2025matanyone2}. However, existing matting methods generally rely on manual guidance, such as trimaps, visual prompts (\eg, points or boxes), and segmentation masks, for instance-level inference (Fig.~\ref{fig:main-task-comparison}). 
Consequently, these techniques necessitate user intervention for each target or repeated forward passes, rendering them impractical for automated stereo conversion pipelines when handling complex multi-target scenes.
While auxiliary-free matting approaches like GVM~\cite{ge2025GVM} are emerging, they are typically designed for specific target categories (\eg, humans or animals) and thus struggle to generalize to the diverse types of soft boundaries present in complex stereo conversion scenarios (\eg, defocus blur in movie shots).

\par
We introduce \textbf{\myname}, a novel layered representation designed to explicitly decompose local foreground and background at soft boundaries for high-fidelity stereo conversion, as shown in Fig.~\ref{fig:main-task-comparison}. To address the challenge of layer mixing, our \myname\ jointly estimates foreground and background information to model soft boundary regions. This explicitly disentangles the mixed colors and resolves depth ambiguity by allocating distinct depth and color values to the overlapping layers at soft boundaries. 
To enable efficient scene-level inference in complex scenarios, we introduce the Circular Alpha Representation (CAR). Unlike vanilla alpha representations that rely on global foreground and background definitions, CAR shifts the focus to modeling opacity exclusively at local soft boundaries. It treats all opaque regions (whether foreground or background) as a single unified class (\eg, see the black regions in Fig.~\ref{fig:main-task-comparison}). 
Furthermore, while resolving all occlusions might theoretically demand numerous layers, we observe that \textit{a two-layer formulation is effective in locally separating foreground and background at soft boundaries, even in complex scenes with many layers.}
This localized approach allows \myname\ to automatically decompose multiple overlapping targets across the entire scene in a single forward pass without user intervention.
Leveraging the estimated \myname\ representation, we perform layered warping to synthesize initial novel views, which are subsequently refined by off-the-shelf inpainting models to generate high-fidelity stereo pairs.
Finally, we introduce an efficient training data curation method that leverages existing image datasets and matting datasets to construct training pairs for \myname\ estimation. In summary, our main contributions are:
\begin{itemize}
    \item We propose a novel layered \myname\ representation that explicitly disentangles mixed colors and resolves depth ambiguity at soft boundaries for high-fidelity stereo conversion.
    \item We design the Circular Alpha Representation (CAR) to model local soft boundary transitions rather than extracting global foregrounds, enabling efficient scene-level inference in a single forward pass without user intervention.
\end{itemize}

\section{Related Work}

\textbf{Stereo Conversion.}
Stereo conversion aims to synthesize high-quality stereo pairs from monocular inputs~\cite{xie2016deep3d,geyer2025eye2eye,dai2024svg}, a crucial technique for immersive media applications~\cite{mehl2024stereoconv,mono2stereo}. Recent methods often leverage generative foundation models: image-based approaches~\cite{stereodiffusion,mono2stereo} utilize diffusion priors, \eg, Stable Diffusion~\cite{rombach2022stablediffusion}, to generate realistic stereo views, while video-based methods~\cite{stereocrafter,geyer2025eye2eye,shvetsova2025m2svid} design spatio-temporal mechanisms, \eg, tiled diffusion strategy~\cite{stereocrafter} and global spatial attention~\cite{shvetsova2025m2svid}, for improved temporal coherence. To mitigate the texture hallucinations and geometric distortions inherent to diffusion models, several works incorporate explicit guidance mechanisms, such as texture bridge~\cite{splatdiff}, guided decoding~\cite{metzger2025elastic3d}, and color fuser~\cite{hairguard}, for enhanced stereo conversion fidelity. However, due to the mixed foreground and background, existing approaches often struggle with soft boundaries and produce stereo results with visual artifacts like background bleeding. 

\par 
\textbf{Depth Estimation.}
Monocular depth estimation predicts dense scene geometry from a single image, serving as a fundamental pillar for 3D vision tasks like stereo conversion~\cite{ranftl2021dpt,yang2024depthanything,mono2stereo,splatdiff}. To achieve robust zero-shot generalization, a dominant paradigm is to train depth models on large-scale datasets~\cite{li2018megadepth,yin2020diversedepth,ranftl2021dpt,yang2024depthanything,wang2025moge,piccinelli2024unidepth}. For instance, MiDaS~\cite{ranftl2020midas} proposes a family of depth losses to enable model training on diverse datasets. Recently, a variety of techniques have been developed to combine real-world and synthetic datasets for enhanced detail extraction, such as teacher-student distillation~\cite{yang2024depthanythingv2,lin2025depthanything3}, edge-guided loss~\cite{unidepthv2}, real data refinement~\cite{wang2025moge2}, and training protocols for fine boundary preservation~\cite{depthpro}.
Alternatively, harnessing geometric priors from pre-trained generative models has emerged as a compelling trajectory. Methods like Marigold~\cite{ke2024marigold}, BetterDepth~\cite{zhang2024betterdepth}, and Pixel-Perfect Depth~\cite{xu2025ppd} cast depth estimation as an iterative denoising process to achieve remarkable detail recovery. Nevertheless, existing depth estimators generally assign a single depth value per pixel. This single-layer representation fundamentally fails at handling soft boundaries, suffering from depth ambiguity where foreground and background overlap.

\par 
\textbf{Alpha Matting.}
Alpha matting estimates the opacity of foreground targets, \ie, alpha mattes, to extract foreground objects from their backgrounds~\cite{xu2017composition_1k,yao2024vitmatte,yang2025matanyone2}, which potentially benefits soft boundary modeling in stereo conversion. 
To resolve semantic ambiguities, previous matting models usually rely on manual guidance, \eg, trimaps, to explicitly define foreground and background~\cite{xu2017composition_1k,park2022matteformer,yao2024vitmatte}. Recent approaches have developed more flexible guidance, \eg, segmentation masks or visual prompts, to facilitate matting for video sequences~\cite{yang2025matanyone,yang2025matanyone2}. To reduce human effort, auxiliary-free methods, such as MODNet~\cite{ke2022modnet}, RVM~\cite{lin2022rvm}, and GVM~\cite{ge2025GVM}, leverage pre-trained model priors, \eg, diffusion priors, or temporal consistency to infer alpha mattes directly from input images. However, both paradigms exhibit critical limitations in complex scenes. Guidance-based methods necessitate user intervention or repeated forward passes for each subject, rendering them impractical for integration into automated stereo conversion pipelines. Conversely, auxiliary-free methods are generally restricted to specific semantic categories (\eg, humans) and struggle to generalize to diverse soft boundaries (\eg, defocus blur). To overcome these bottlenecks, we introduce the Circular Alpha Representation (CAR). By shifting the paradigm from global foreground extraction to local soft boundary decomposition, CAR enables \myname\ to automatically infer opacity across complex scenes in a single forward pass.

\begin{figure}[t]
    \centering
    \begin{subfigure}{\linewidth}
        \centering
        \includegraphics[width=\linewidth]{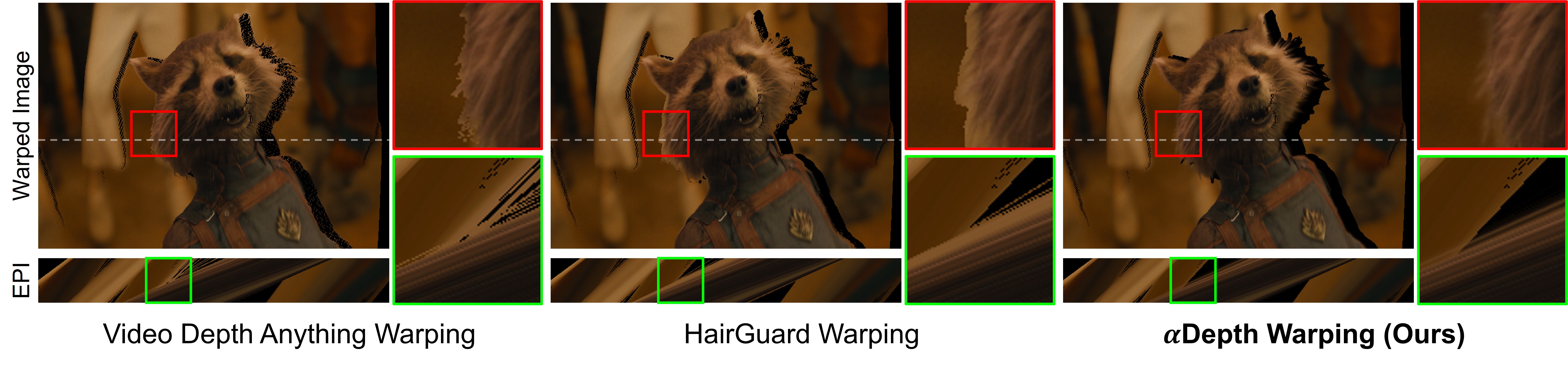}
        \vspace{-1.5em}
        \caption{Visual comparison of warping performance and Epipolar Plane Images (EPIs)}
        \label{fig:main-epi}
    \end{subfigure}
    \begin{subfigure}{\linewidth}
        \centering
        \includegraphics[width=\linewidth]{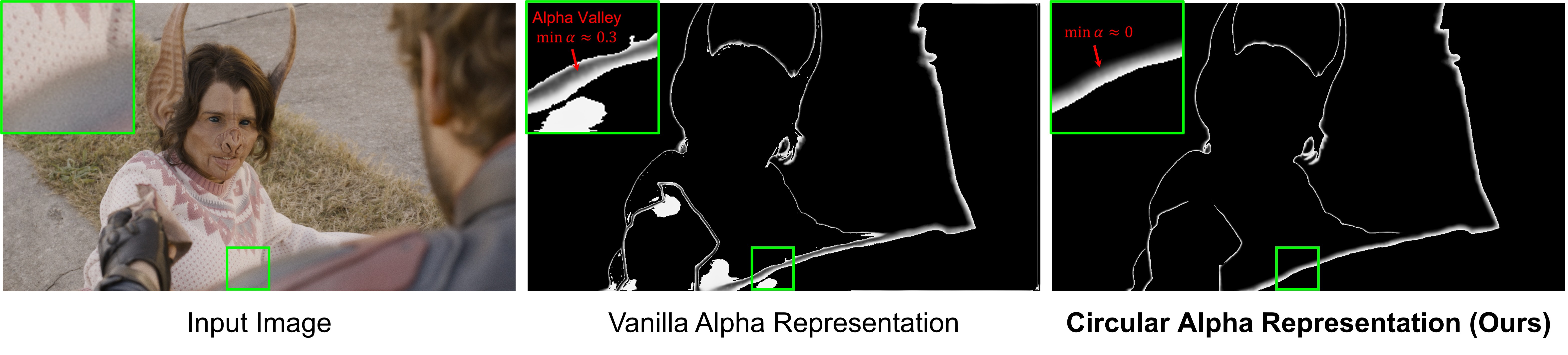}
        \vspace{-1.5em}
        \caption{Alpha valley issue (Alpha is shown only at soft regions, \ie, $\alpha\in[0.02,0.98]$, for comparison)}
        \label{fig:main-alpha-valley}
    \end{subfigure}
    \caption{\textbf{Challenges of soft boundary recovery in stereo conversion.} (a) We evaluate warping performance via Epipolar Plane Images (EPIs) extracted along the gray dashed line under uniform rightward camera motion. Direct warping with Video Depth Anything~\cite{chen2025videodepthanything} struggles with depth ambiguity at soft boundaries, causing broken edges and flying pixels. Although HairGuard~\cite{hairguard} captures better details, its single-layer depth representation often results in background bleeding and aliasing (\eg, see EPIs). By contrast, our layered representation effectively models soft boundaries to achieve superior warping and multi-view consistency. (b) Vanilla alpha representations often suffer from ``alpha valleys'' (\ie, alpha estimation errors at intersecting boundaries) due to their reliance on explicit global foreground and background definitions. By modeling local transitions rather than global separation, our Circular Alpha Representation (CAR) robustly handles overlapping targets to produce accurate alpha values.}
    \label{fig:main-problem-analysis}
\end{figure}
\begin{figure}[t]
    \centering
    \includegraphics[width=0.8\linewidth]{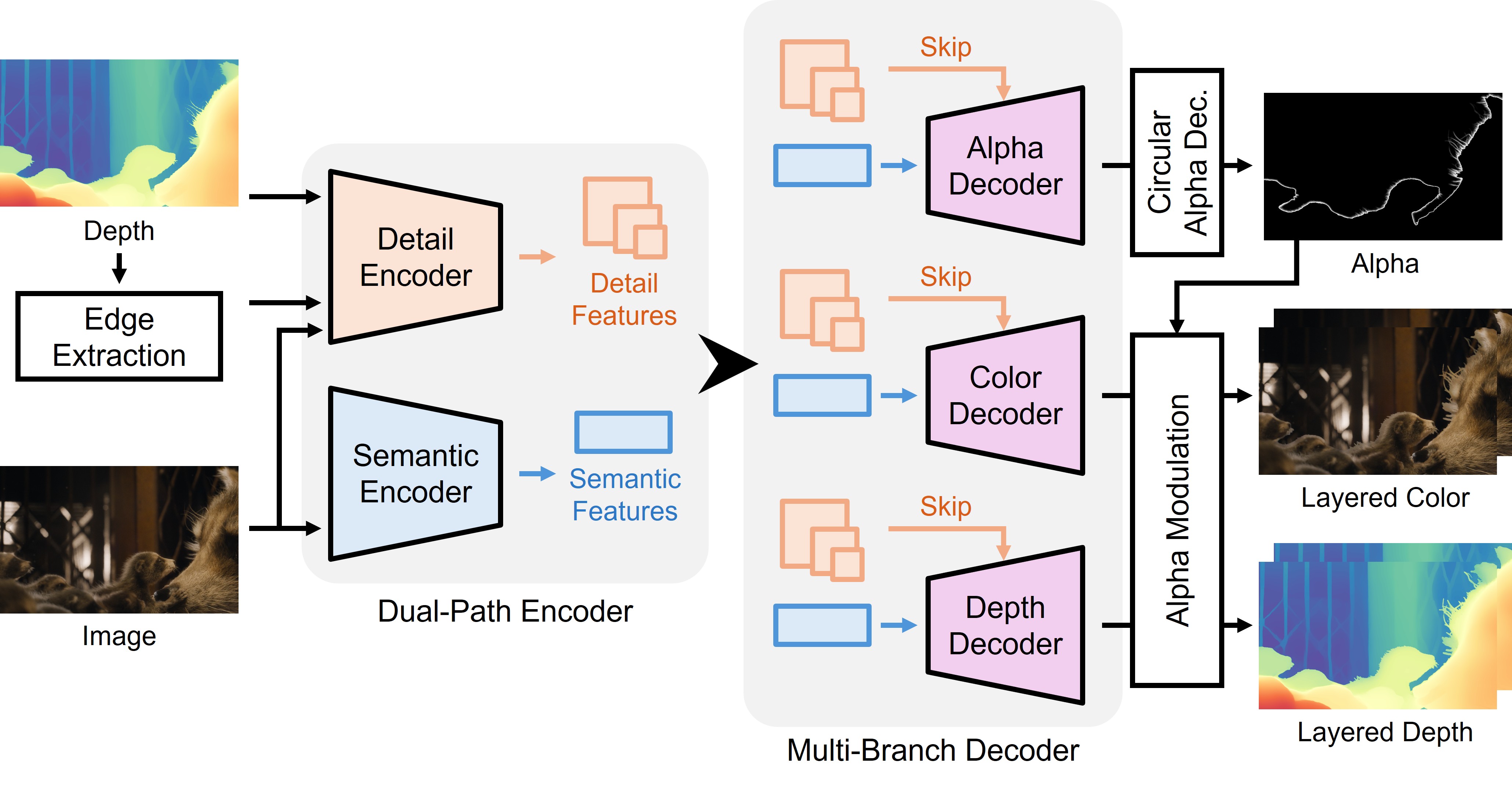}
    \caption{\textbf{\myname\ estimation pipeline.} Given an image and its corresponding depth map (\eg, from a pre-trained depth model), we employ a dual-path encoder to extract both semantic and detail features. A multi-branch decoder then processes these features for task-specific predictions. Finally, we apply circular alpha decoding to generate the estimated alpha map, which subsequently modulates and constrains the layered color and depth predictions on soft boundary regions.}
    \label{fig:main-network}
\end{figure}
\section{Method}~\label{main-sec:method}
We first analyze the main challenges of soft boundary recovery in stereo conversion (Sec.~\ref{subsec:prob_analysis}), and then propose the layered \myname\ representation for efficient soft boundary decomposition (Sec.~\ref{subsec:alphadepth}).


\subsection{Problem Analysis}~\label{subsec:prob_analysis}
Stereo conversion in real-world scenarios frequently encounters complex scenes characterized by multiple overlapping targets and intricate soft boundaries (\eg, hair, fur, and defocus blur). At these soft boundaries, the observed color $I$ is inherently a mixture of the foreground color $I_\mathrm{FG}$ and the background color $I_\mathrm{BG}$, modulated by an opacity value $\alpha \in [0, 1]$. Following alpha matting~\cite{yao2024vitmatte,yang2025matanyone2}, the color blending at soft boundaries can be defined as:
\begin{equation}\label{eq:alpha_composition}
    I = \alpha I_{\mathrm{FG}} + (1 - \alpha) I_{\mathrm{BG}}.
\end{equation}
Mixed colors at soft boundaries, coupled with multiple targets, introduce significant challenges:

\par 
\textbf{Depth Ambiguity.} 
Based on Eq.~(\ref{eq:alpha_composition}), pixels within a soft boundary mix color contributions from both foreground and background layers. However, most monocular depth estimators predict only a single depth value per pixel~\cite{yang2024depthanything,ke2024marigold,wang2025moge,xu2025ppd}. This single-layer representation inherently results in depth ambiguity at soft boundaries. Consequently, view transformation techniques like softmax splatting~\cite{niklaus2020softmaxsplat} struggle to project these pixels accurately, leading to broken boundaries and severe flying pixels (Fig.~\ref{fig:main-epi}). Although recent methods like HairGuard~\cite{hairguard} capture finer boundary details, their single-layer depth often leads to background bleeding and aliasing artifacts (Fig.~\ref{fig:main-epi}).


\par
\textbf{Scene-level Inference.} Boundaries at depth discontinuities are crucial for high-fidelity stereo conversion, as they govern the realistic rendering of occlusions and binocular parallax~\cite{mono2stereo,hairguard}. 
However, complex scenes often feature multiple overlapping targets at varying depths (\eg, Fig.~\ref{fig:main-alpha-valley}),
and even individual targets may possess intricate soft boundaries within their own structures (\eg, see alpha in the top example of Fig.~\ref{fig:main-teaser}).
Existing matting methods typically rely on a global definition of foreground and background, \eg, foreground defined by user inputs like trimaps or visual prompts~\cite{yao2024vitmatte,yang2025matanyone2}. Consequently, they either require iterative user guidance for each instance or indiscriminately merge multiple targets into a single monolithic foreground, discarding vital inter-object boundary details.
\par
Furthermore, learning a vanilla alpha representation for multiple overlapping targets often suffers from the ``\textit{alpha valley}'' issue (Fig.~\ref{fig:main-alpha-valley}).
To preserve local boundary details of overlapping foreground instances, alpha discontinuities emerge at their intersections (\eg, see alpha in Fig.~\ref{fig:main-circular-alpha}). 
Since neural networks are usually biased toward smooth predictions, they often struggle to resolve these sharp transitions, yielding inaccurate alpha values (Fig.~\ref{fig:main-alpha-valley}). These estimation errors ultimately propagate through the stereo conversion pipeline, yielding inconsistent soft boundaries in the stereo pairs.

\subsection{\myname}~\label{subsec:alphadepth}
We propose a layered \myname\ representation to address depth ambiguity and alpha valley issues for efficient soft boundary decomposition. Instead of using manual guidance like trimaps, we directly utilize image semantics and scene geometry for \myname\ estimation, enabling scene-level inference without user intervention (Fig.~\ref{fig:main-network}). Given an input image $I_{\mathrm{IN}}$ and its depth map $D_{\mathrm{IN}}$, we first extract depth gradients to facilitate soft boundary localization. Then, we employ a dual-path encoder to capture both high-level contextual cues and fine-grained structural details. Specifically, a detail encoder built upon UNet encoder~\cite{ronneberger2015unet} extracts multi-scale features to preserve rich textural and structural details. Concurrently, a semantic encoder (based on DINOv2~\cite{oquab2023dinov2} with Depth Anything V2 weights initialization~\cite{yang2024depthanythingv2}) extracts deep semantic features for soft boundary reasoning. Finally, these features are fed into a multi-branch decoder (based on UNet decoder~\cite{ronneberger2015unet}) to predict the \myname\ representation, which jointly estimates alpha, layered color, and layered depth at soft boundaries.

\begin{figure}[t]
    \centering
    \includegraphics[width=\linewidth]{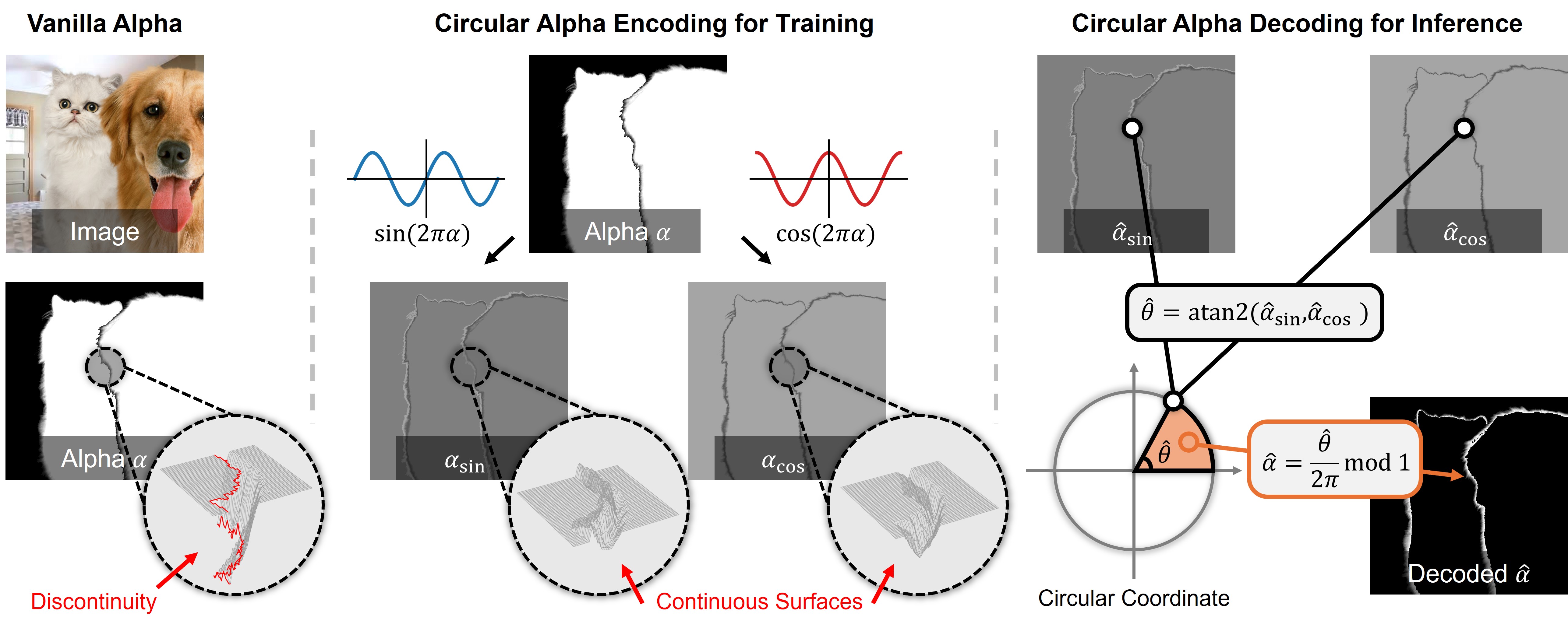}
    \caption{\textbf{Circular Alpha Representation (CAR).} The vanilla alpha representation inherently suffers from sharp discontinuities at the intersecting boundaries of multiple overlapping instances. By contrast, CAR encodes the ground-truth alpha into continuous trigonometric space during training, benefiting model optimization and eliminating alpha valley issues (Fig.~\ref{fig:main-alpha-valley}). During inference, the predicted trigonometric components are decoded back into an alpha matte $\hat{\alpha}$ at soft boundaries. }
    \label{fig:main-circular-alpha}
\end{figure}

\par 
\textbf{Circular Alpha Representation (CAR).}
Traditional matting paradigms rely on a global definition of foreground and background~\cite{yao2024vitmatte,yang2025matanyone2}. In scenarios featuring multiple overlapping objects at varying depths, this global assignment forces sharp discontinuities at inter-object intersecting boundaries (Fig.~\ref{fig:main-circular-alpha}), leading to the alpha valley issue (see Fig.~\ref{fig:main-alpha-valley}).
By contrast, the proposed CAR reformulates the task by shifting from global foreground extraction to local soft boundary decomposition. We treat all opaque regions, whether foreground or background, as a single unified class and focus on estimating opacity at semi-transparent boundaries. To achieve this, we project the ground-truth alpha map $\alpha \in [0, 1]$ into a continuous trigonometric space via circular alpha encoding (Fig.~\ref{fig:main-circular-alpha}):
\begin{equation}\label{eq:circular_alpha_encoding}
\alpha_{\mathrm{sin}} = \sin(2\pi \alpha), \quad \alpha_{\mathrm{cos}} = \cos(2\pi \alpha).
\end{equation}
By leveraging the periodicity of trigonometric functions, we wrap the linear alpha scale around a circle, mapping background ($\alpha=0$) and opaque foreground ($\alpha=1$) to the same coordinate $(0, 1)$ in the $(\alpha_{\mathrm{sin}}, \alpha_{\mathrm{cos}})$ space.
This transformation collapses the discrete jump at intersecting boundaries into a continuous manifold (Fig.~\ref{fig:main-circular-alpha}). By bridging the gap between $\alpha=0$ and $\alpha=1$, it eliminates boundary discontinuities and facilitates model optimization.

\par 
During inference, the alpha decoder directly estimates the continuous trigonometric components, \ie, $\hat{\alpha}_{\mathrm{sin}}$ and $\hat{\alpha}_{\mathrm{cos}}$. We then apply circular alpha decoding to reconstruct alpha values $\hat{\alpha}\in[0, 1)$ via the four-quadrant inverse tangent function, \ie, 
\begin{equation}\label{eq:circular_alpha_decoding}
    \hat{\alpha} = \frac{\hat{\theta}}{2\pi} \bmod 1, \quad \text{where} \quad \hat{\theta} = \operatorname{atan2}(\hat{\alpha}_{\mathrm{sin}}, \hat{\alpha}_{\mathrm{cos}}).
\end{equation}
As shown in Fig.~\ref{fig:main-circular-alpha}, CAR projects opaque regions into a unified class with $\hat{\alpha}=0$ and focuses on estimating opacity for soft boundaries. 
By navigating this circular space, CAR circumvents alpha valley issues and enables scene-level soft boundary decomposition in a single forward pass.

\par 
\textbf{Layered Representation.} 
To resolve depth ambiguity and color mixing, we explicitly decouple soft boundaries into local foreground (FG) and background (BG) representations. 
While complex scenes may theoretically require numerous global layers to account for all occlusions, we observe that \textit{local soft boundaries can be effectively modeled using a two-layer decomposition}, \ie, locally differentiating foreground and background for each soft boundary. Thus, we adopt a two-layer representation for both color and depth at soft boundary regions. Specifically, we estimate the layered color $\bar{I}_{\mathrm{FG}}, \bar{I}_{\mathrm{BG}}\in\mathbb{R}^{3\times h\times w}$ and color blending weights $W_{\mathrm{FG}}^{I}, W_{\mathrm{BG}}^{I}\in\mathbb{R}^{h\times w}$ via the color decoder. Similarly, the depth decoder predicts the layered depth  $\bar{D}_{\mathrm{FG}}, \bar{D}_{\mathrm{BG}}\in\mathbb{R}^{h\times w}$ and depth blending weights $W_{\mathrm{FG}}^{D}, W_{\mathrm{BG}}^{D}\in\mathbb{R}^{h\times w}$. In addition, we estimate soft boundary regions $\hat{S}$ by thresholding the estimated $\hat{\alpha}$, \ie, $\hat{S} = \mathbb{I}(\alpha_{\mathrm{th}} \le \hat{\alpha} \le 1 - \alpha_{\mathrm{th}})$ with $\alpha_{\mathrm{th}}=0.02$ and $\mathbb{I}(\cdot)$ denoting the indicator function. The layered prediction is then formulated via alpha-modulated blending, \ie,
\begin{equation}\label{eq:alpha_modulated_blending}
\hat{X}_{\star} = W_{\star}^{'X} \odot \bar{X}_{\star} + (1 - W_{\star}^{'X}) \odot X_{\mathrm{IN}}, \quad \text{with} \quad W_{\star}^{'X} = \hat{S} \odot (1 - W_{\star}^{X}),
\end{equation}
where $X \in \{I, D\},\ \star \in \{\mathrm{FG}, \mathrm{BG}\},$ and $\odot$ denotes spatial element-wise multiplication. Eq.~(\ref{eq:alpha_modulated_blending}) restricts the layered modeling exclusively to the soft boundary regions $\hat{S}$. For opaque regions, the original input colors and depths are retained, which preserves high-fidelity textures from $I_{\mathrm{IN}}$ and facilitates plug-and-play integration with state-of-the-art depth estimation models. 
We then project the estimated \myname\ representation to the target viewpoint via layered warping (see Sec.~\ref{supp-sec:layered-warping} for more details). To generate the final stereo results, we employ the pretrained scene painter and color fuser from HairGuard~\cite{hairguard} for disocclusion inpainting and texture enhancement.




\begin{figure}[t]
    \centering
    \includegraphics[width=\linewidth]{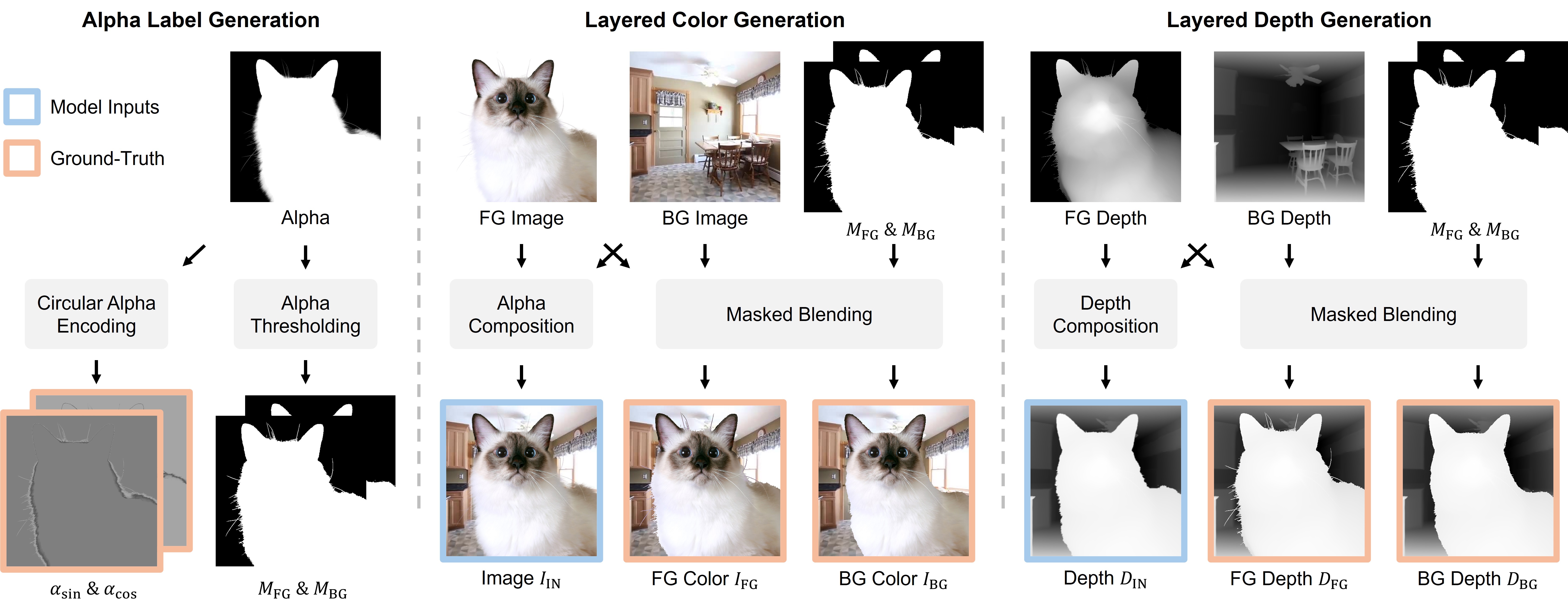}
    \caption{\textbf{Training Data Curation.} Firstly, the alpha map is processed via circular alpha encoding to yield continuous alpha labels ($\alpha_{\sin}, \alpha_{\cos}$) and thresholded to produce layered masks ($M_{\mathrm{FG}}, M_{\mathrm{BG}}$). In layered color/depth generation, foreground and background assets are composited to form the synthesized input image ($I_{\mathrm{IN}}$) and depth ($D_{\mathrm{IN}}$). Concurrently, masked blending is applied to generate ground-truth color layers ($I_{\mathrm{FG}}, I_{\mathrm{BG}}$) and depth layers ($D_{\mathrm{FG}}, D_{\mathrm{BG}}$) for soft boundary regions.}
    \label{fig:main-train-pipeline}
\end{figure}

\par
\textbf{Model Training.}
We propose an efficient data curation strategy to utilize existing matting datasets for \myname\ training (Fig.~\ref{fig:main-train-pipeline}). Given a foreground RGBA image from matting datasets, we first project the ground-truth alpha matte $\alpha$ into continuous trigonometric labels $\alpha_{\mathrm{sin}}, \alpha_{\mathrm{cos}}$ via circular alpha encoding Eq.~(\ref{eq:circular_alpha_encoding}). We also derive the binary foreground mask $M_{\mathrm{FG}} = \mathbb{I}(\alpha \ge \alpha_{\mathrm{th}})$ and background mask $M_{\mathrm{BG}} = \mathbb{I}(\alpha \ge 1 - \alpha_{\mathrm{th}})$ for layered label generation. $M_{\mathrm{FG}}$ captures soft boundary regions for synthesizing foreground color/depth labels, whereas $M_{\mathrm{BG}}$ helps preserve background information at soft boundaries. For layered color generation, we composite the foreground and background (sampled from image datasets) via standard alpha blending Eq.~(\ref{eq:alpha_composition}) to synthesize the input image $I_{\mathrm{IN}}$. We then apply masked blending using $M_{\mathrm{FG}}$ and $M_{\mathrm{BG}}$ to generate the ground-truth layered colors $I_{\mathrm{FG}}$ and $I_{\mathrm{BG}}$ at soft boundary regions. For layered depth generation, we synthesize the input depth map $D_{\mathrm{IN}}$ following the depth composition protocol of HairGuard~\cite{hairguard}. Analogously, masked blending is applied to generate ground-truth layered depths $D_{\mathrm{FG}}$ and $D_{\mathrm{BG}}$ (please see Sec.~\ref{supp-sec:train-data-curation} for more details).
\par 
The \myname\ network is trained by jointly optimizing color, depth, and alpha representations, \ie,
\begin{equation}\label{eq:total_loss}
\mathcal{L}_{\mathrm{total}} = \mathcal{L}_{I} + \mathcal{L}_{D} + \mathcal{L}_{\alpha},
\end{equation}
where the color loss $\mathcal{L}_{I} = \mathcal{L}(\hat{I}_{\mathrm{FG}}, I_{\mathrm{FG}}) + \mathcal{L}(\hat{I}_{\mathrm{BG}}, I_{\mathrm{BG}})$, the depth loss $\mathcal{L}_{D} = \mathcal{L}(\hat{D}_{\mathrm{FG}}, D_{\mathrm{FG}}) + \mathcal{L}(\hat{D}_{\mathrm{BG}}, D_{\mathrm{BG}})$, and the alpha loss $\mathcal{L}_{\alpha} = \mathcal{L}(\hat{\alpha}_{\mathrm{sin}}, \alpha_{\mathrm{sin}}) + \mathcal{L}(\hat{\alpha}_{\mathrm{cos}}, \alpha_{\mathrm{cos}})$. To facilitate stable multi-task learning, we apply the same loss function $\mathcal{L}(\cdot)$ across all modalities and follow a two-stage training scheme. 
The first-stage training focuses on recovering fine-grained details at local soft boundaries with $\mathcal{L}(\cdot)$ defined as
$\mathcal{L}(\hat{X}, X) = \mathcal{L}_{1}(\hat{X}, X) + \mathcal{L}_{\mathrm{m}}(S \odot \hat{X}, S \odot X),$
where $\mathcal{L}_{1}$ is $\ell_1$ loss, $\mathcal{L}_{\mathrm{m}}$ denotes the matting loss from ViTMatte~\cite{yao2024vitmatte}, and $S = \mathbb{I}(\alpha_{\mathrm{th}} \le \alpha \le 1 - \alpha_{\mathrm{th}})$ is the soft boundary mask. In the second stage, we apply the matting loss $\mathcal{L}_{\mathrm{m}}$ across the entire image space for global refinement, \ie, $\mathcal{L}(\hat{X}, X) = \mathcal{L}_{\mathrm{m}}(\hat{X}, X)$.

\section{Experiments and Analysis}\label{main-sec:exp}

\subsection{Experimental Settings}
\textbf{Implementation}. We train the \myname\ network with AdamW optimizer~\cite{loshchilov2018adamw} under $448\times 448$ patches, batch size 32, and $1\times 10^{-5}$ learning rate. To curate training pairs, we sample background images from RealEstate10K~\cite{re10k} and DL3DV-10K~\cite{ling2024dl3dv}, and use foreground images from matting datasets: {AM-2K}~\cite{li2020am2k}, {Distinctions-646}~\cite{Qiao_2020Distinctions_646}, and {Composition-1K}~\cite{xu2017composition_1k}. We train the model for 50 epochs per stage, which takes approximately 6 days in total on an NVIDIA RTX A6000 GPU.

\par 
\textbf{Evaluation}. We employ the Mono2Stereo~\cite{mono2stereo} and Marvel-10K~\cite{hairguard} datasets for stereo image/video conversion. For fair comparisons, all baselines use the same depth models: Depth Anything V2 (DAv2)~\cite{yang2024depthanythingv2} for Mono2Stereo and Video Depth Anything (VDA)~\cite{chen2025videodepthanything} for Marvel-10K. We also employ two natural image matting datasets {AIM-500}~\cite{aim_500} and {P3M-10K}~\cite{li2021p3m_10k} to evaluate the performance of our circular alpha representation.

\def\imgWidth{0.32\linewidth} %
\def\depthWidth{0.19\linewidth} %
\def\pointWidth{0.24\linewidth} %

\def\scc{(-1.9,-1.4)}

\def\rebigone{(-0.5, -0.32)} %
\def\rebigtwo{(1.3, -0.32)} %

\def\zoomone{(-0.2,0.44)} %
\def\zoomtwo{(-0.25,0.44)} %

\def\zoomthree{(-0.8,0.3)} %
\def\zoomfour{(-0.94,0.3)} %

\def\ssizz{0.8cm} %
\def\ssmag{5}

\begin{figure}[t]
\centering
\tikzstyle{img} = [rectangle, minimum width=\imgWidth]
    \centering
    \begin{subfigure}{\linewidth}
    \centering
    \begin{subfigure}{\depthWidth}
        \begin{tikzpicture}[spy using outlines={green,magnification=\ssmag,size=\ssizz},inner sep=0]
            \node [align=center, img] {\includegraphics[width=\textwidth]{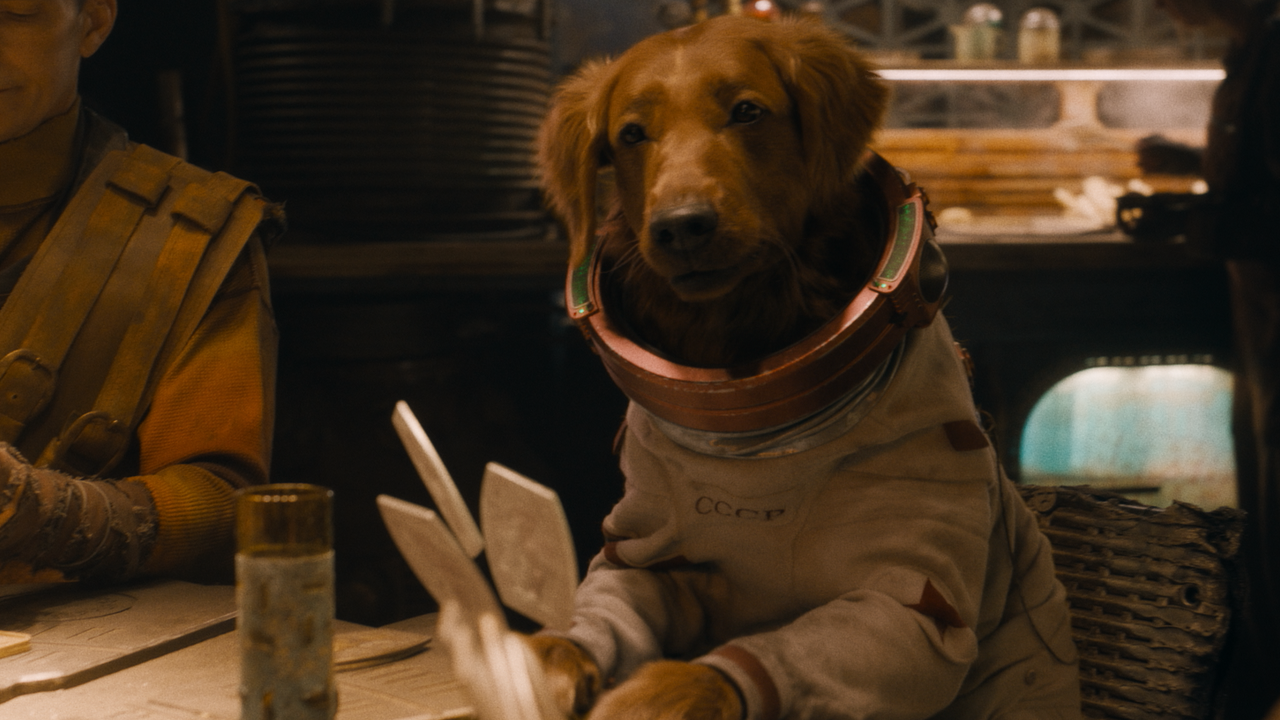}};
            \spy on \zoomone in node [left] at \rebigone;
    	\end{tikzpicture}
    \end{subfigure}
    \begin{subfigure}{\depthWidth}
		\begin{tikzpicture}[spy using outlines={green,magnification=\ssmag,size=\ssizz},inner sep=0]
            \node [align=center, img] {\includegraphics[width=\textwidth]{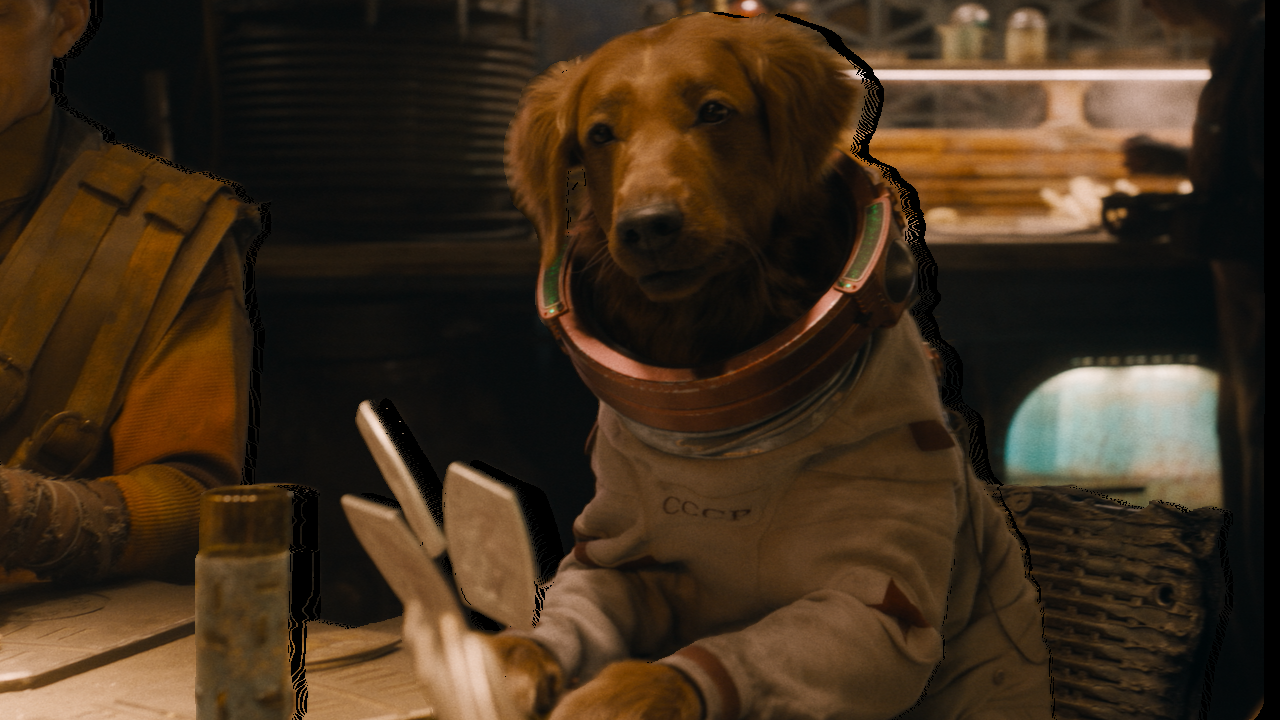}};
            \spy on \zoomtwo in node [left] at \rebigone;
    	\end{tikzpicture}
    \end{subfigure}
    \begin{subfigure}{\depthWidth}
        \begin{tikzpicture}[spy using outlines={green,magnification=\ssmag,size=\ssizz},inner sep=0]
            \node [align=center, img] {\includegraphics[width=\textwidth]{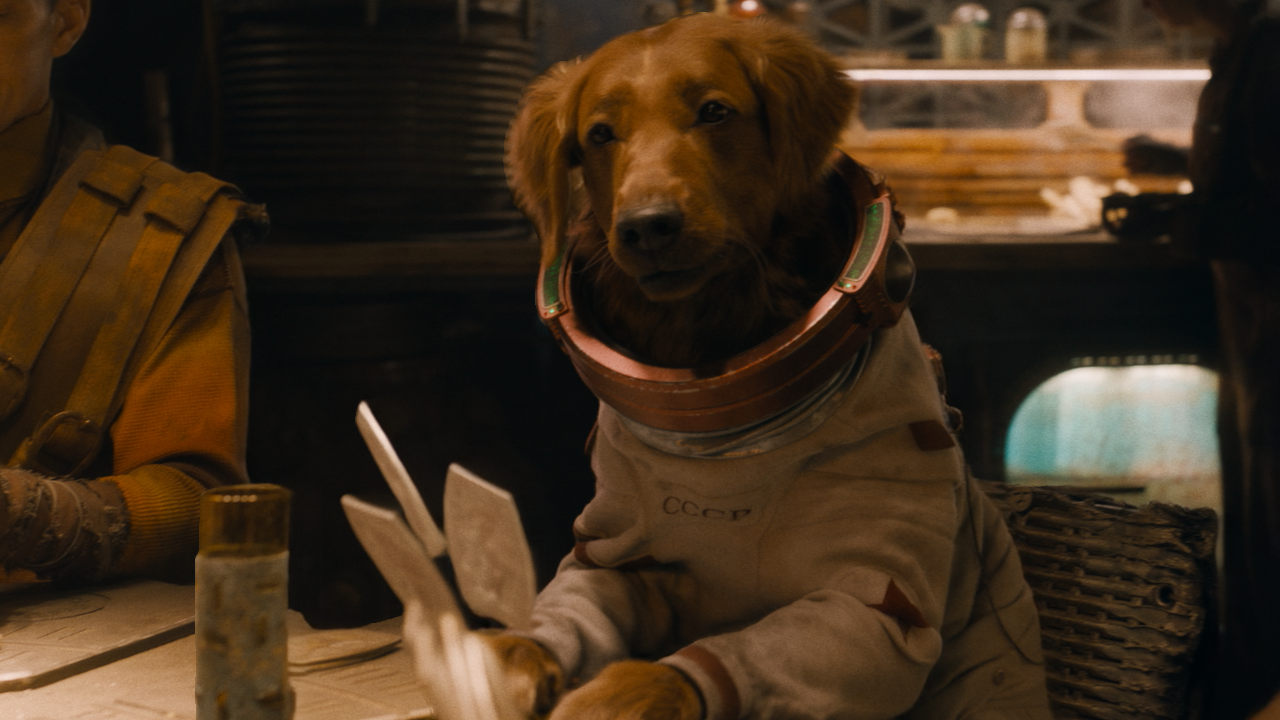}};
            \spy on \zoomtwo in node [left] at \rebigone;
    	\end{tikzpicture}
      \end{subfigure}
    \begin{subfigure}{\depthWidth}
        \begin{tikzpicture}[spy using outlines={green,magnification=\ssmag,size=\ssizz},inner sep=0]
            \node [align=center, img] {\includegraphics[width=\textwidth]{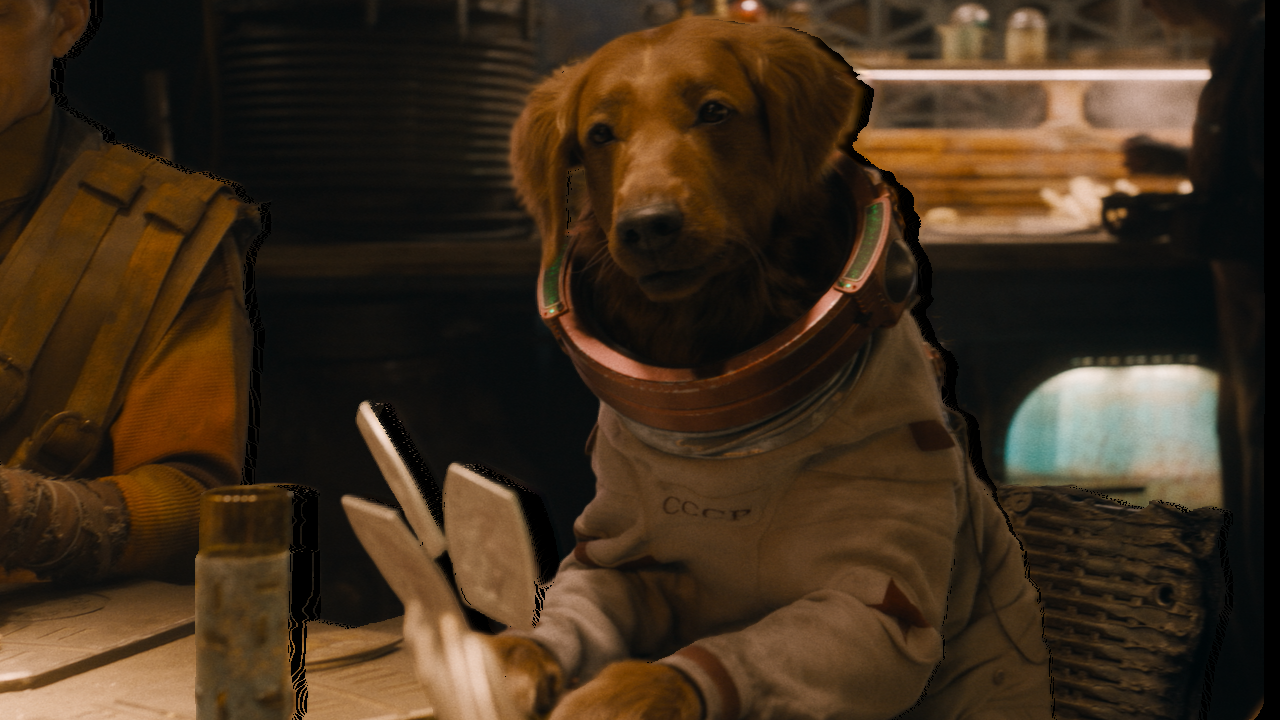}};
            \spy on \zoomtwo in node [left] at \rebigone;
    	\end{tikzpicture}
      \end{subfigure}
    \begin{subfigure}{\depthWidth}
        \begin{tikzpicture}[spy using outlines={green,magnification=\ssmag,size=\ssizz},inner sep=0]
            \node [align=center, img] {\includegraphics[width=\textwidth]{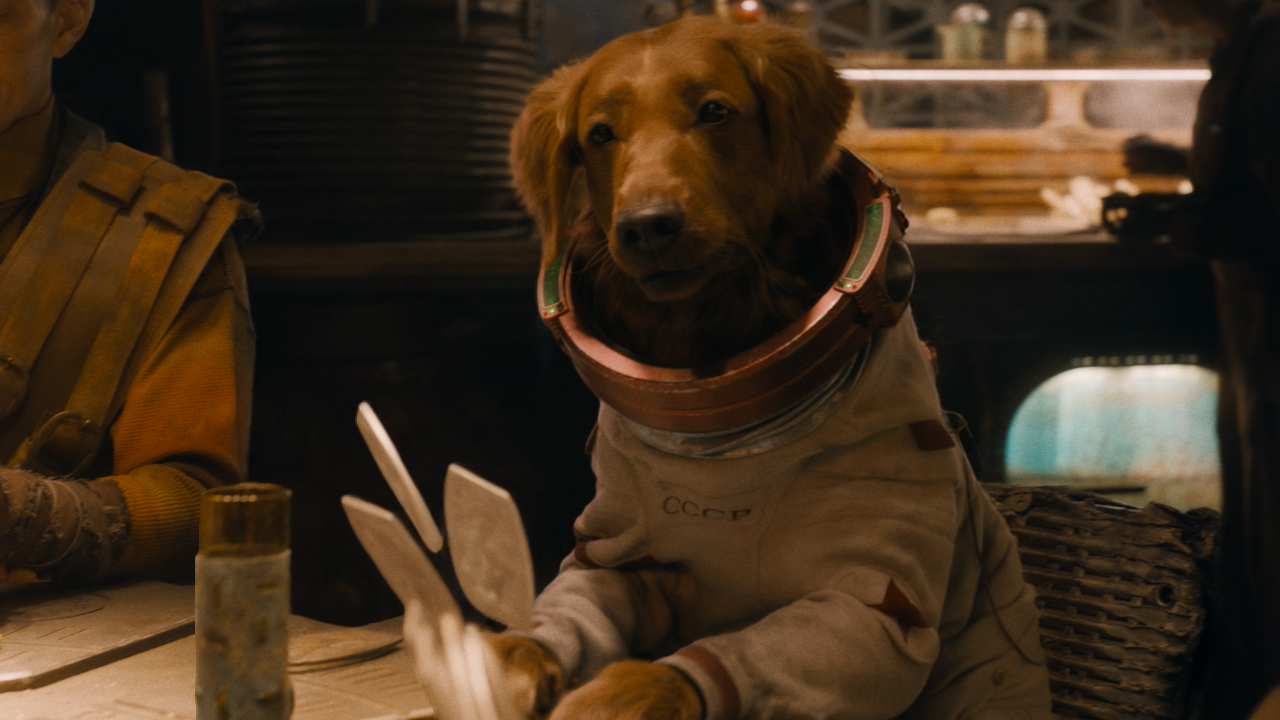}};
            \spy on \zoomtwo in node [left] at \rebigone;
    	\end{tikzpicture}
      \end{subfigure}
    \end{subfigure}
    \begin{subfigure}{\linewidth}
    \centering
    \begin{subfigure}{\depthWidth}
        \begin{tikzpicture}[spy using outlines={green,magnification=\ssmag,size=\ssizz},inner sep=0]
            \node [align=center, img] {\includegraphics[width=\textwidth]{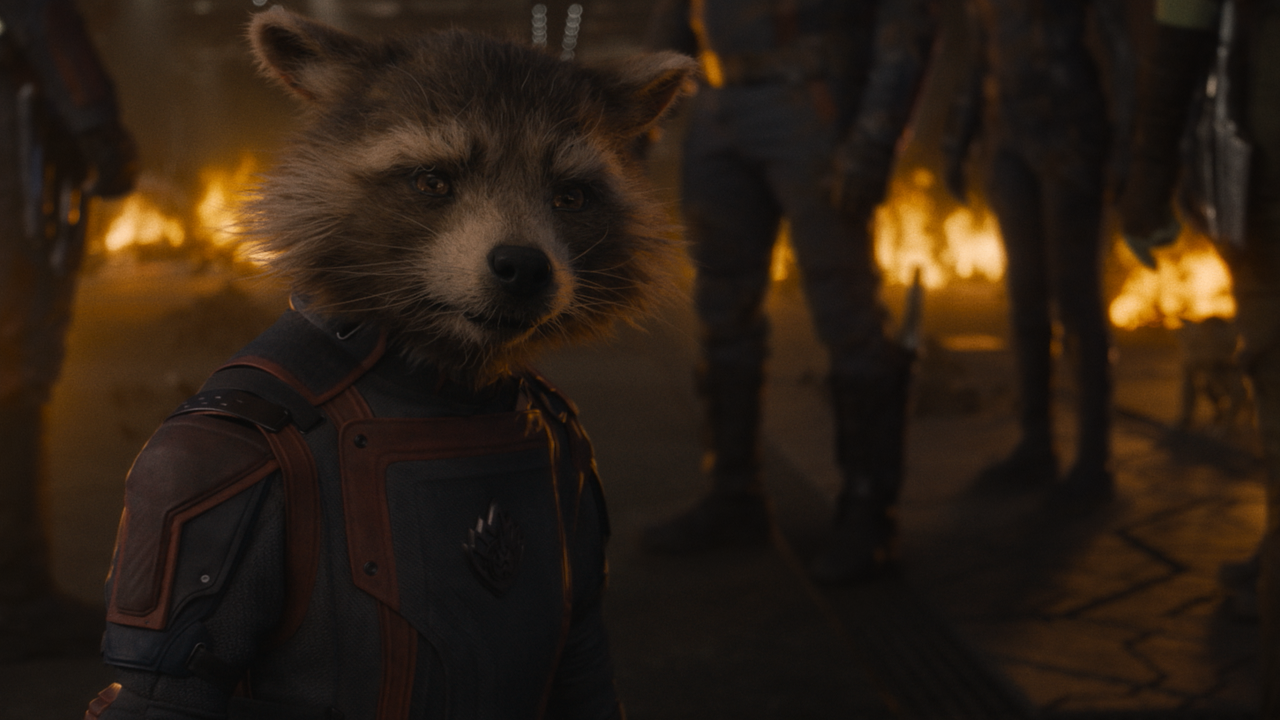}};
            \spy on \zoomthree in node [left] at \rebigtwo;
    	\end{tikzpicture}
        \caption*{Input Image}
    \end{subfigure}
    \begin{subfigure}{\depthWidth}
		\begin{tikzpicture}[spy using outlines={green,magnification=\ssmag,size=\ssizz},inner sep=0]
            \node [align=center, img] {\includegraphics[width=\textwidth]{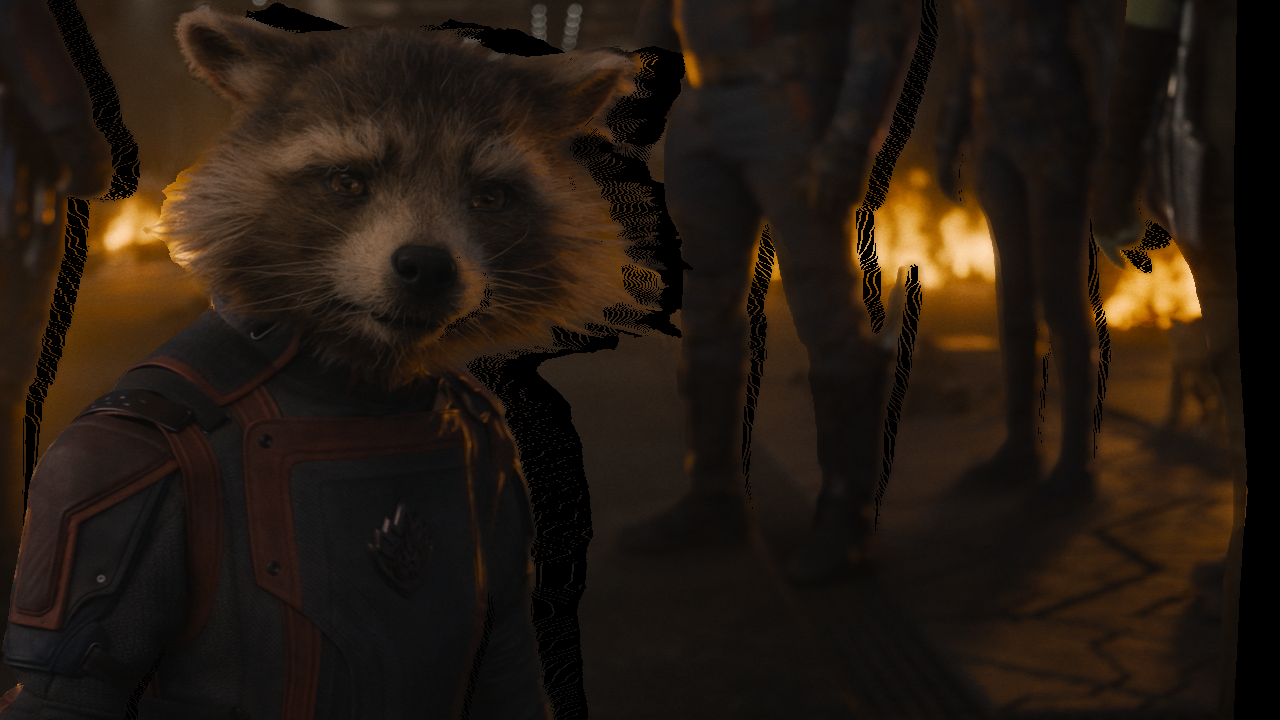}};
            \spy on \zoomfour in node [left] at \rebigtwo;
    	\end{tikzpicture}
        \caption*{HairGuard Warping}
    \end{subfigure}
    \begin{subfigure}{\depthWidth}
        \begin{tikzpicture}[spy using outlines={green,magnification=\ssmag,size=\ssizz},inner sep=0]
            \node [align=center, img] {\includegraphics[width=\textwidth]{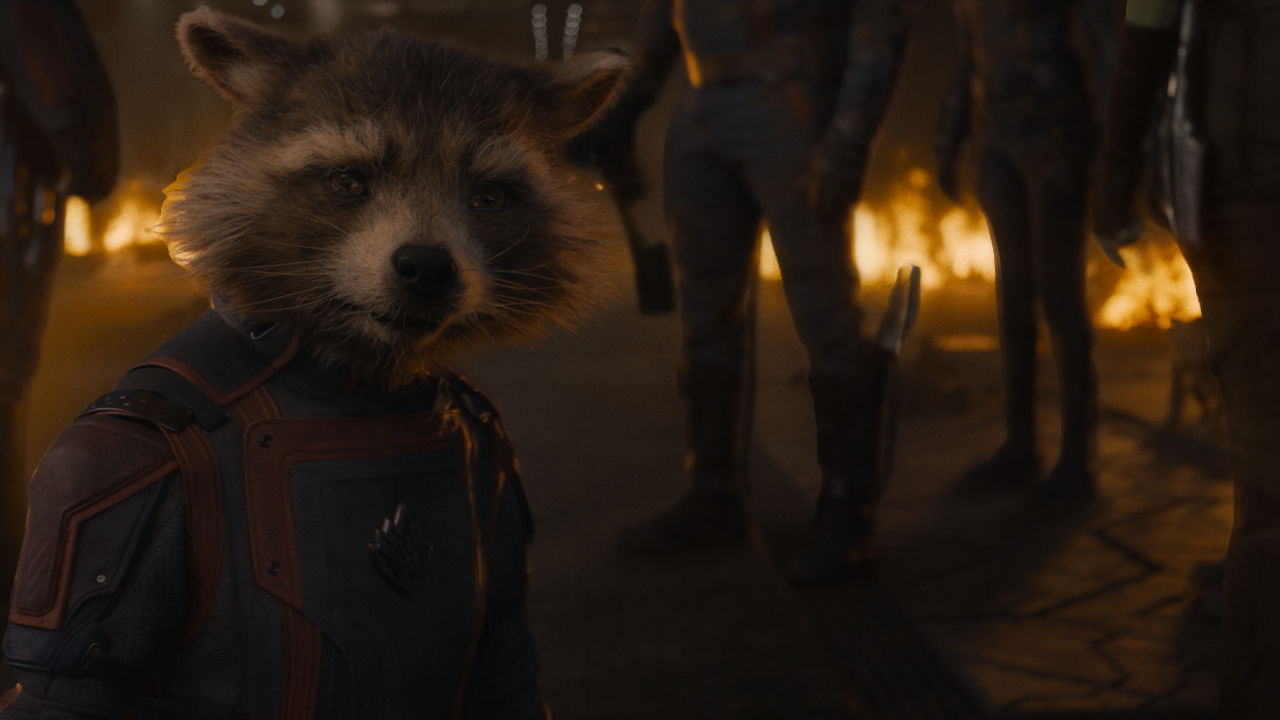}};
            \spy on \zoomfour in node [left] at \rebigtwo;
    	\end{tikzpicture}
        \caption*{HairGuard Result}
      \end{subfigure}
    \begin{subfigure}{\depthWidth}
        \begin{tikzpicture}[spy using outlines={green,magnification=\ssmag,size=\ssizz},inner sep=0]
            \node [align=center, img] {\includegraphics[width=\textwidth]{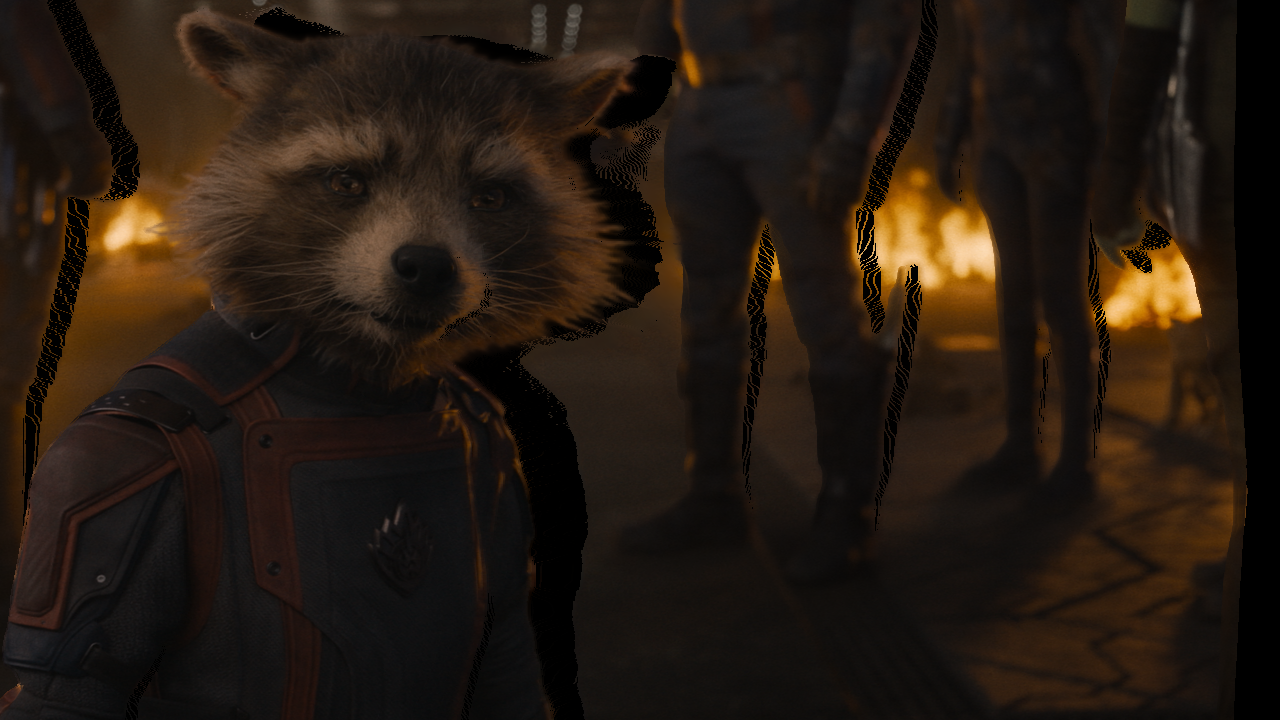}};
            \spy on \zoomfour in node [left] at \rebigtwo;
    	\end{tikzpicture}
        \caption*{\textbf{Our Warping}}
      \end{subfigure}
    \begin{subfigure}{\depthWidth}
        \begin{tikzpicture}[spy using outlines={green,magnification=\ssmag,size=\ssizz},inner sep=0]
            \node [align=center, img] {\includegraphics[width=\textwidth]{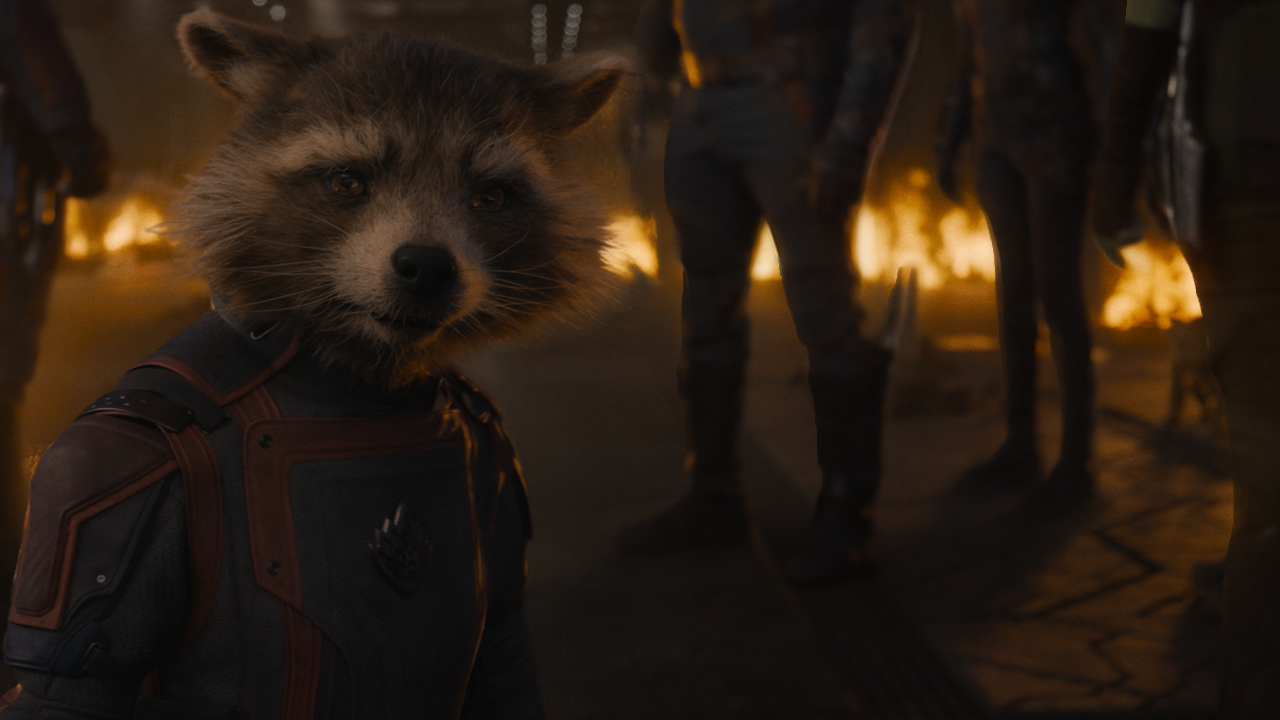}};
            \spy on \zoomfour in node [left] at \rebigtwo;
    	\end{tikzpicture}
        \caption*{\textbf{Our Result}}
      \end{subfigure}
    \end{subfigure}
    \caption{\textbf{Visual comparisons with HairGuard~\cite{hairguard}} in warping and stereo conversion. }
    \label{fig:main-visual-stereo}
\end{figure}

\begin{table*}[t]
\centering
\caption{\textbf{Stereo image/video conversion performance}. The \textcolor{red}{best} and \textcolor{blue}{second best} results are marked.}
\label{tab:main-stereo-conv-full}
\footnotesize
\setlength\tabcolsep{2.4pt}
\begin{tabular}{lcccc c ccccc}
\toprule[0.15em]
\rowcolor{color3} & \multicolumn{4}{c}{\textbf{Stereo Image Conversion (Mono2Stereo)}} &  & \multicolumn{5}{c}{\textbf{Stereo Video Conversion (Marvel-10K)}} \\
\cline{2-5} \cline{7-11}
\rowcolor{color3} \multirow{-2}{*}{\textbf{Method}} & S-PSNR $\uparrow$ & S-SSIM $\uparrow$ & LPIPS $\downarrow$ & DISTS $\downarrow$ &  & S-PSNR $\uparrow$ & S-SSIM $\uparrow$ & LPIPS $\downarrow$ & DISTS $\downarrow$ & FVD $\downarrow$ \\ \midrule
StereoDiffusion~\cite{stereodiffusion} & 22.95 & 0.6407 & 0.2098 & 0.0747 &  & 23.53 & 0.6805 & 0.2173 & 0.0694 & 18.41 \\
Mono2Stereo~\cite{mono2stereo} & 24.48 & 0.6933 & 0.1990 & 0.0751 &  & 25.19 & 0.7309 & 0.1979 & 0.0690 & 19.81 \\
StereoCrafter~\cite{stereocrafter} & 24.24 & 0.7073 & 0.2334 & 0.1198 &  & 25.96 & 0.7619 & 0.2287 & 0.1231 & 10.05 \\
SplatDiff~\cite{splatdiff} & 24.78 & 0.7339 & 0.1235 & {\color{blue} 0.0439} &  & 26.57 & 0.8006 & 0.0831 & 0.0364 & 2.18 \\
HairGuard~\cite{hairguard} & {\color{blue} 25.05} & {\color{blue} 0.7445} & {\color{blue} 0.1212} & 0.0444 &  & {\color{blue} 27.10} & {\color{blue} 0.8239} & {\color{blue} 0.0734} & {\color{blue} 0.0322} & {\color{blue} 2.14} \\
\rowcolor{color3} \textbf{\myname\ (Ours)} & {\color{red} 25.60} & {\color{red} 0.7554} & {\color{red} 0.1181} & {\color{red} 0.0417} &  & {\color{red} 28.46} & {\color{red} 0.8373} & {\color{red} 0.0699} & {\color{red} 0.0297} & {\color{red} 1.72} \\ \bottomrule[0.15em]
\end{tabular}
\end{table*}


\begin{wraptable}{r}{0.52\columnwidth}
\vspace{-\intextsep}
\centering
\caption{\textbf{Warping performance} on Marvel-10K~\cite{hairguard}. \textcolor{red}{Best} and \textcolor{blue}{second best} results are marked.}
\label{tab:main-warping-performance}
\footnotesize
\setlength\tabcolsep{1.5pt}
\begin{tabular}{lccccc}
\toprule[0.15em]
\rowcolor{color3} {\textbf{Method}} & S-PSNR & S-SSIM & LPIPS & DISTS & FVD \\ \midrule
VDA~\cite{chen2025videodepthanything} & 26.14 & 0.6718 & 0.1337 & 0.0910 & {\color{blue} 7.21} \\
VDA + HairGuard~\cite{hairguard} & {\color{blue} 27.90} & {\color{blue} 0.7328} & {\color{blue} 0.1277} & {\color{blue} 0.0839} & 9.39 \\
\rowcolor{color3} \textbf{VDA + \myname\ (Ours)} & {\color{red} 28.68} & {\color{red} 0.7636} & {\color{red} 0.1190} & {\color{red} 0.0738} & {\color{red} 7.13} \\ \bottomrule[0.15em]
\end{tabular}
\end{wraptable}

\subsection{Stereo Conversion}
Tab.~\ref{tab:main-stereo-conv-full} compares \myname\ with state-of-the-art methods for stereo image/video conversion. Since soft boundaries usually occupy a small fraction of the image, we compute pixel-level metrics exclusively on soft regions $\hat{S}$ (denoted by S-PSNR and S-SSIM), alongside whole-image perceptual metrics (LPIPS~\cite{zhang2018lpips} and DISTS~\cite{dists}). 
Benefiting from soft boundary decomposition, our \myname\ consistently outperforms previous methods and eliminates artifacts like background bleeding (Fig.~\ref{fig:main-visual-stereo}), yielding the best video consistency (FVD~\cite{unterthiner2018FVD} in Tab.~\ref{tab:main-stereo-conv-full}). Finally, Tab.~\ref{tab:main-warping-performance} compares our warping performance against baselines using the original depth from VDA~\cite{chen2025videodepthanything} and refined depth from HairGuard~\cite{hairguard}. While HairGuard improves soft boundary details, its single-layer depth fails to resolve depth ambiguity (\eg, mixed colors in Fig.~\ref{fig:main-visual-stereo}). By contrast, our approach performs layered modeling on local soft boundaries and achieves the best warping performance.


\def\imgWidth{0.32\linewidth} %
\def\depthWidth{0.19\linewidth} %
\def\pointWidth{0.24\linewidth} %

\def\scc{(-1.9,-1.4)}

\def\rebigone{(-0.5, -0.5)} %
\def\rebigtwo{(1.3, -0.5)} %

\def\zoomeight{(0.2,0.2)} %
\def\zoomnine{(-1.02,-0.3)} %

\def\ssizz{0.8cm} %
\def\ssmag{3}

\begin{figure}[t]
\centering
\tikzstyle{img} = [rectangle, minimum width=\imgWidth]
    \centering
      \begin{subfigure}{\linewidth}
    \centering
    \begin{subfigure}{\depthWidth}
        \begin{tikzpicture}[spy using outlines={green,magnification=\ssmag,size=\ssizz},inner sep=0]
            \node [align=center, img] {\includegraphics[width=\textwidth]{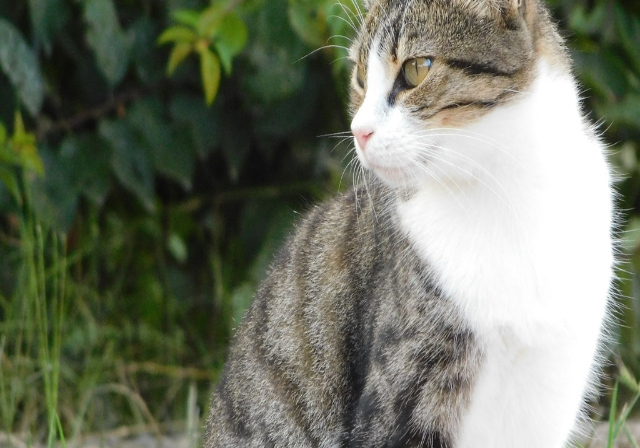}};
            \spy on \zoomeight in node [left] at \rebigone;
    	\end{tikzpicture}
    \end{subfigure}
    \begin{subfigure}{\depthWidth}
		\begin{tikzpicture}[spy using outlines={green,magnification=\ssmag,size=\ssizz},inner sep=0]
            \node [align=center, img] {\includegraphics[width=\textwidth]{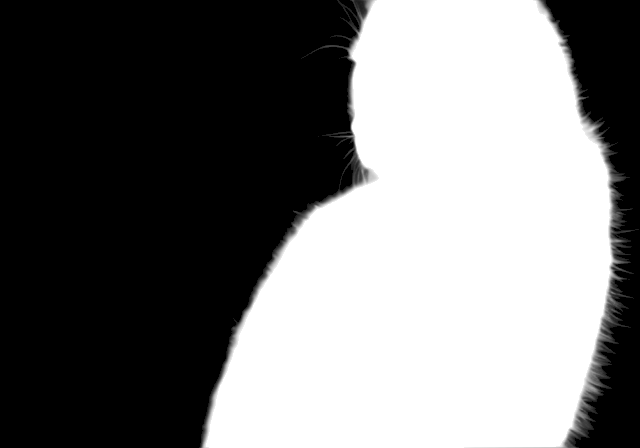}};
            \spy on \zoomeight in node [left] at \rebigone;
    	\end{tikzpicture}
    \end{subfigure}
    \begin{subfigure}{\depthWidth}
        \begin{tikzpicture}[spy using outlines={green,magnification=\ssmag,size=\ssizz},inner sep=0]
            \node [align=center, img] {\includegraphics[width=\textwidth]{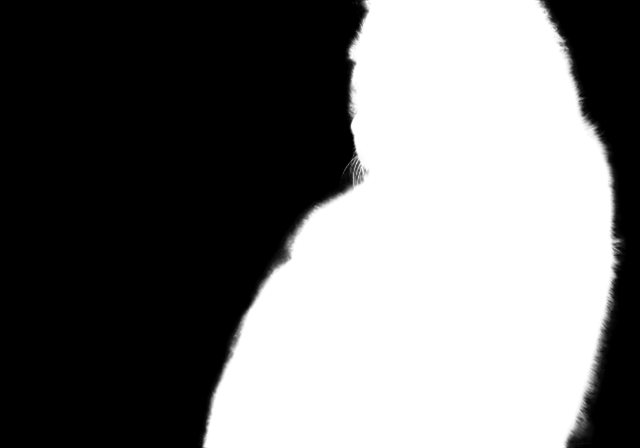}};
            \spy on \zoomeight in node [left] at \rebigone;
    	\end{tikzpicture}
      \end{subfigure}
    \begin{subfigure}{\depthWidth}
        \begin{tikzpicture}[spy using outlines={green,magnification=\ssmag,size=\ssizz},inner sep=0]
            \node [align=center, img] {\includegraphics[width=\textwidth]{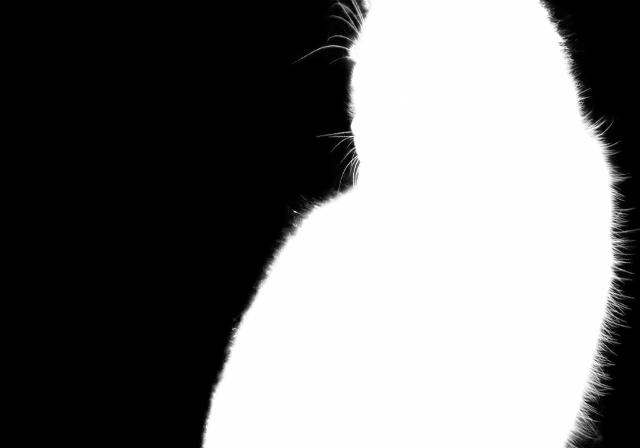}};
            \spy on \zoomeight in node [left] at \rebigone;
    	\end{tikzpicture}
      \end{subfigure}
    \begin{subfigure}{\depthWidth}
        \begin{tikzpicture}[spy using outlines={green,magnification=\ssmag,size=\ssizz},inner sep=0]
            \node [align=center, img] {\includegraphics[width=\textwidth]{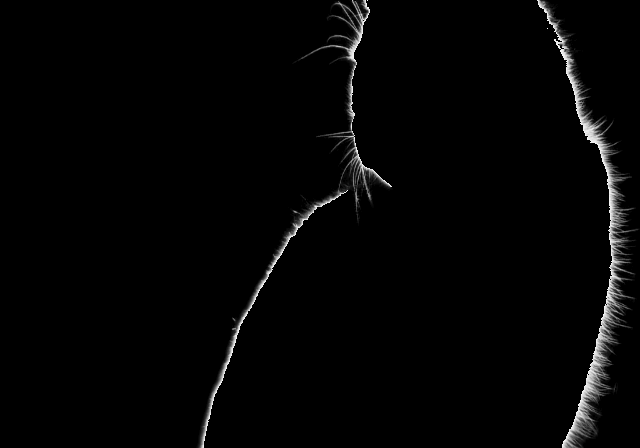}};
            \spy on \zoomeight in node [left] at \rebigone;
    	\end{tikzpicture}
      \end{subfigure}
    \end{subfigure}
    \begin{subfigure}{\linewidth}
    \centering
    \begin{subfigure}{\depthWidth}
        \begin{tikzpicture}[spy using outlines={green,magnification=\ssmag,size=\ssizz},inner sep=0]
            \node [align=center, img] {\includegraphics[width=\textwidth]{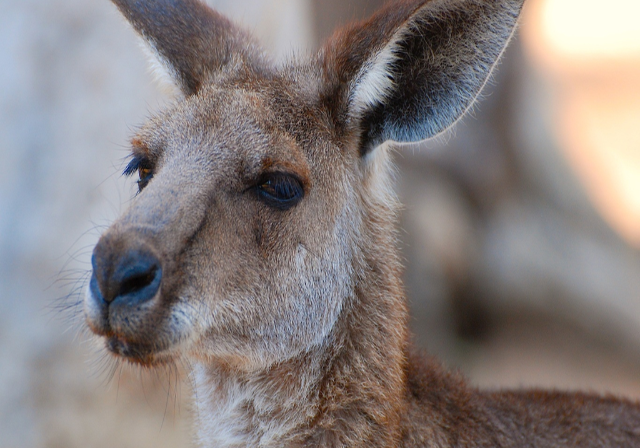}};
            \spy on \zoomnine in node [left] at \rebigtwo;
    	\end{tikzpicture}
        \caption*{Input Image}
    \end{subfigure}
    \begin{subfigure}{\depthWidth}
		\begin{tikzpicture}[spy using outlines={green,magnification=\ssmag,size=\ssizz},inner sep=0]
            \node [align=center, img] {\includegraphics[width=\textwidth]{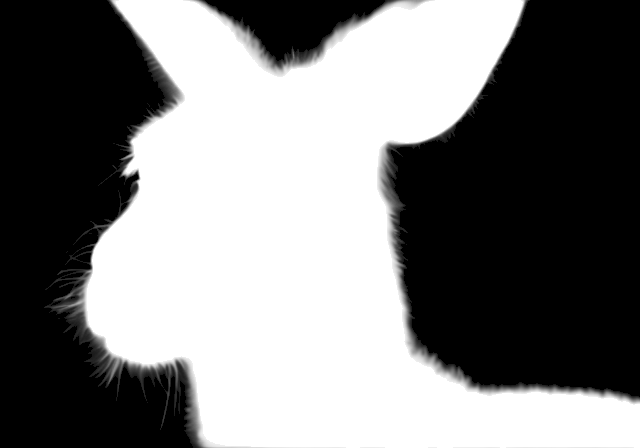}};
            \spy on \zoomnine in node [left] at \rebigtwo;
    	\end{tikzpicture}
        \caption*{GVM~\cite{ge2025GVM}}
    \end{subfigure}
    \begin{subfigure}{\depthWidth}
        \begin{tikzpicture}[spy using outlines={green,magnification=\ssmag,size=\ssizz},inner sep=0]
            \node [align=center, img] {\includegraphics[width=\textwidth]{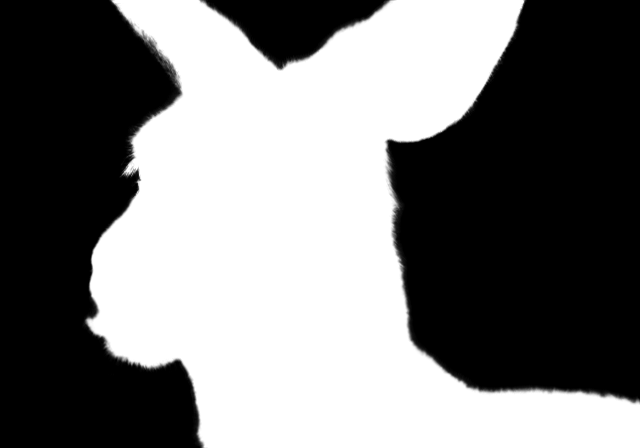}};
            \spy on \zoomnine in node [left] at \rebigtwo;
    	\end{tikzpicture}
        \caption*{MatAnyone 2~\cite{yang2025matanyone2}}
      \end{subfigure}
    \begin{subfigure}{\depthWidth}
        \begin{tikzpicture}[spy using outlines={green,magnification=\ssmag,size=\ssizz},inner sep=0]
            \node [align=center, img] {\includegraphics[width=\textwidth]{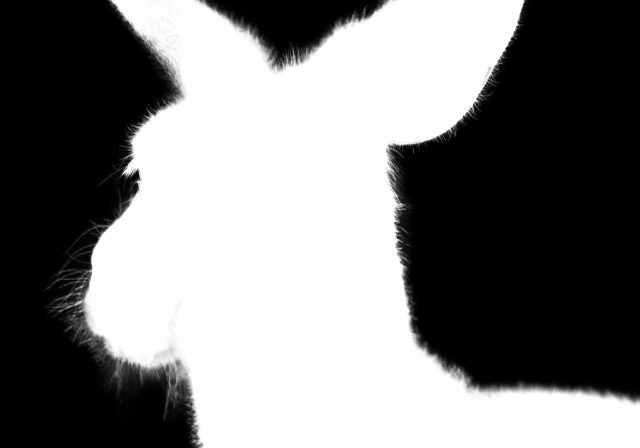}};
            \spy on \zoomnine in node [left] at \rebigtwo;
    	\end{tikzpicture}
        \caption*{ViTMatte~\cite{yao2024vitmatte}}
      \end{subfigure}
    \begin{subfigure}{\depthWidth}
        \begin{tikzpicture}[spy using outlines={green,magnification=\ssmag,size=\ssizz},inner sep=0]
            \node [align=center, img] {\includegraphics[width=\textwidth]{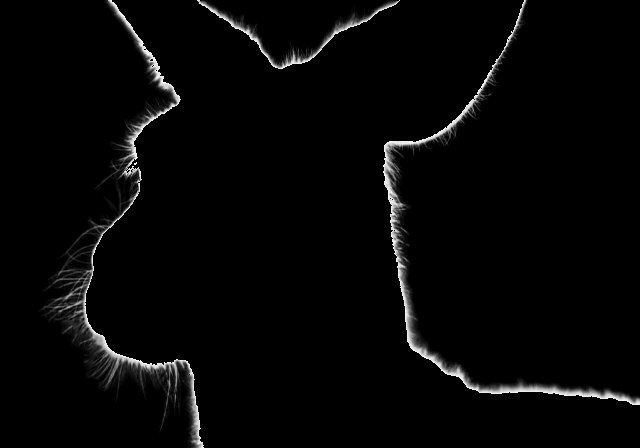}};
            \spy on \zoomnine in node [left] at \rebigtwo;
    	\end{tikzpicture}
        \caption*{\textbf{\myname\ (Ours)}}
      \end{subfigure}
    \end{subfigure}
    \caption{\textbf{Visual comparisons with alpha matting methods.} }
    \label{fig:main-visual-alpha}
\end{figure}

\begin{table*}[t]
\centering
\caption{\textbf{Alpha matting performance}. The \textcolor{red}{best} and \textcolor{blue}{second best} results are marked.}
\label{tab:main-alpha-performance}
\footnotesize
\setlength\tabcolsep{3pt}
\begin{tabular}{lccccc c ccc}
\toprule[0.15em]
\rowcolor{color3} & \textbf{Guidance} & \textbf{Average} & \multicolumn{3}{c}{\textbf{AIM-500}} & & \multicolumn{3}{c}{\textbf{P3M-10K}} \\
\cline{4-6} \cline{8-10}
\rowcolor{color3} \multirow{-2}{*}{\textbf{Method}} & \textbf{Type} & \textbf{Rank} & SAD $\downarrow$  & Grad $\downarrow$ & Conn $\downarrow$ & & SAD $\downarrow$ & Grad $\downarrow$ & Conn $\downarrow$ \\ \midrule
ViTMatte~\cite{yao2024vitmatte} & Trimap & \textcolor{blue}{2.17} & 7.44 & {\color{red} 17.82} & 4.07 & & {\color{blue} 4.06} & {\color{blue} 11.27} & {\color{blue} 2.29} \\
GVM~\cite{ge2025GVM}    & Auxiliary-Free & 4.00 & 9.64 & 34.04 & 5.35 & & 5.56 & 18.30 & 3.11 \\
MatAnyone 2~\cite{yang2025matanyone2} & Segmentation Mask & \textcolor{red}{1.67} & {\color{blue} 7.42} & 21.61 & {\color{blue} 4.06} & & {\color{red} 3.63} & {\color{red} 10.10} & {\color{red} 2.08} \\
\rowcolor{color3} \textbf{\myname\ (Ours)} & Depth & \textcolor{blue}{2.17} & {\color{red} 7.24} & {\color{blue} 20.20} & {\color{red} 3.89} & & 4.09 & 11.88 & 2.31 \\
\bottomrule[0.15em]
\end{tabular}
\end{table*}

\subsection{Alpha Matting}\label{sec:main-exp-matting}
We evaluate our Circular Alpha Representation (CAR) against trimap-based (ViTMatte~\cite{yao2024vitmatte}), mask-based (MatAnyone 2~\cite{yang2025matanyone2}), and auxiliary-free (GVM~\cite{ge2025GVM}) baselines. Since CAR predicts alpha values at local soft boundaries without defining global foreground/background, we apply circular alpha encoding Eq.~(\ref{eq:circular_alpha_encoding}) to ground-truth labels and the predictions of all baselines for fair comparisons (please see Sec.~\ref{supp-sec:matting-eval-details} for more details). Following ViTMatte~\cite{yao2024vitmatte}, we compute standard alpha metrics (SAD, Grad, Conn) exclusively within the unknown regions of trimaps. As shown in Tab.~\ref{tab:main-alpha-performance} and Fig.~\ref{fig:main-visual-alpha}, CAR performs comparably to state-of-the-art methods without requiring manual guidance. 
Furthermore, in contrast to prior matting methods, our design is able to handle complex multi-target scenes (Fig.~\ref{fig:main-alpha-valley}) and capture intra-object soft boundaries (\eg, Figs.~\ref{fig:main-visual-alpha} top and \ref{fig:main-teaser} top).



\begin{table}[!t]
\centering
\footnotesize
\caption{\textbf{Ablation study}. (a) Effects of Alpha Estimation (AE), Circular Alpha Representation (CAR), and Layered Representation (LR) on warping performance. (b) Impacts of different strategies on alpha matting. The \textcolor{red}{best} and \textcolor{blue}{second best} results are marked. Please see Sec.~\ref{supp-sec:vis-ablation} for visual results.}
\label{tab:main-ablation}

\begin{subtable}[t]{0.45\linewidth}
\centering
\caption{\myname\ ablation. }
\vspace{-0.5em}
\label{tab:main-ablation-warping}
\setlength\tabcolsep{3pt}
\begin{tabular}{lcccccc}
\toprule[0.15em]
\rowcolor{color3}
                     &                                      &                                 &                                 & \multicolumn{3}{c}{\textbf{Marvel-10K}}                                \\
\cline{5-7} \rowcolor{color3}
\multirow{-2}{*}{\textbf{ID}} & \multirow{-2}{*}{\textbf{AE}} & \multirow{-2}{*}{\textbf{CAR}} & \multirow{-2}{*}{\textbf{LR}} & S-SSIM $\uparrow$                     & LPIPS $\downarrow$                      & DISTS $\downarrow$                      \\
\midrule
A\#1                  &                                  &                             &                             & 0.6718                        & 0.1337                        & 0.0910                        \\
A\#2                  & $\checkmark$                       &                           &                             & 0.7165                        & 0.1346                        & 0.0869                        \\
A\#3                  & $\checkmark$                       & $\checkmark$             &                              & \textcolor{blue}{0.7214}      & \textcolor{blue}{0.1280}      & \textcolor{blue}{0.0819}      \\
A\#4                  & $\checkmark$                       & $\checkmark$             & $\checkmark$                  & \textcolor{red}{0.7636}       & \textcolor{red}{0.1190}       & \textcolor{red}{0.0738}       \\
\bottomrule[0.15em]
\end{tabular}
\end{subtable}
\hfill
\begin{subtable}[t]{0.5\linewidth}
\centering
\caption{CAR ablation.}
\vspace{-0.5em}
\label{tab:main-ablation-alpha}
\setlength\tabcolsep{2.3pt}
\begin{tabular}{lcccc}
\toprule[0.15em]
\rowcolor{color3}
                     &                                      & \multicolumn{3}{c}{\textbf{P3M-10K}} \\
\cline{3-5} \rowcolor{color3}
\multirow{-2}{*}{\textbf{ID}} & \multirow{-2}{*}{\textbf{Strategies}} & SAD $\downarrow$ & Grad $\downarrow$ & Conn $\downarrow$ \\
\midrule
B\#1                 & Vanilla Alpha Representation & 4.47                 & 12.41                 & \textcolor{blue}{2.42} \\
B\#2                 & $\ell_1$ Loss Only               & 4.36                 & 12.54                 & 2.48                 \\
B\#3                 & Single-Stage Training           & \textcolor{blue}{4.29} & \textcolor{red}{11.77}  & 2.43                 \\
B\#4                 & \textbf{CAR (Ours)}             & \textcolor{red}{4.09}  & \textcolor{blue}{11.88} & \textcolor{red}{2.31}  \\
\bottomrule[0.15em]
\end{tabular}
\end{subtable}
\end{table}

\subsection{Ablation Study}
To validate our design choices, we analyze both warping and alpha matting performance in Tab.~\ref{tab:main-ablation}. 
\textbf{(i) \myname\ Ablation}. Compared with the baseline directly using VDA depth~\cite{chen2025videodepthanything} for warping (A\#1), estimating only foreground information (\ie, foreground color, depth, and alpha) at soft boundaries mitigates background bleeding and improves structural recovery (S-SSIM of A\#2 in Tab.~\ref{tab:main-ablation-warping}). Replacing the vanilla alpha representation with our proposed CAR better handles complex scenes, yielding consistent gains (A\#3 \vs\ A\#2). By additionally estimating background information, our layered \myname\ representation achieves the best warping performance via soft boundary decomposition (A\#4). \textbf{(ii) CAR Ablation}. Tab.~\ref{tab:main-ablation-alpha} validates the contributions of matting loss $\mathcal{L}_{\mathrm{m}}$ (B\#2 \vs\ B\#4) and two-stage training (B\#3 \vs\ B\#4) in alpha matting performance. Unlike vanilla alpha representation that often suffers from unstable predictions due to alpha valley issues (Fig.~\ref{fig:main-alpha-valley}), our CAR focuses on local soft boundaries and delivers the best matting performance (B\#1 \vs\ B\#4).



\begin{table*}[t]
    \centering
    \caption{\textbf{Synergy between circular alpha and layered representation}. We follow the experimental setup in Tab.~\ref{tab:main-ablation-warping} and compare our circular alpha against vanilla alpha in terms of warping performance. The \textcolor{red}{best} and \textcolor{blue}{second best} results are marked. }
    \label{tab:synergy}
    \footnotesize
    \begin{tabular}{cccc}
    \toprule[0.15em]
    \rowcolor{color3}
                                                              & \multicolumn{3}{c}{\textbf{Marvel-10K}} \\
    \cline{2-4} \rowcolor{color3}
      \multirow{-2}{*}{\textbf{Strategies}} & S-SSIM $\uparrow$ & LPIPS $\downarrow$ & DISTS $\downarrow$ \\
    \midrule
     Vanilla Alpha & 0.7165                & 0.1346                 & 0.0869 \\
     Vanilla Alpha + Layered Representation               & 0.7207                 & 0.1294                & 0.0832                 \\
     Circular Alpha           & \textcolor{blue}{0.7214} & \textcolor{blue}{0.1280}  & \textcolor{blue}{0.0819}                 \\
     Circular Alpha + Layered Representation            & \textcolor{red}{0.7636}  & \textcolor{red}{0.1190} & \textcolor{red}{0.0738}  \\
    \bottomrule[0.15em]
    \end{tabular}
\end{table*}

\begin{table*}[t]
\centering
\caption{\textbf{Impact of alpha maps on stereo conversion}. We conduct experiments on Marvel-10K using MatAnyone 2~\protect\cite{yang2025matanyone2} in place of our circular alpha estimation. Note that MatAnyone 2 requires masks as guidance, while $\alpha$Depth operates without any manual guidance. \textcolor{red}{Best} results are marked.}
\label{tab:alphadepth-matanyone-comparison}
\footnotesize
\setlength\tabcolsep{5pt}
\begin{tabular}{lccc c cccc c}
\toprule[0.15em]
\rowcolor{color3} \textbf{Stage} & \textbf{Alpha Source} & PSNR $\uparrow$ & SSIM $\uparrow$ &  & S-PSNR $\uparrow$ & S-SSIM $\uparrow$ & LPIPS $\downarrow$ & DISTS $\downarrow$ & FVD $\downarrow$ \\ \midrule
\multirow{2}{*}{Warping Only} & MatAnyone 2 & 27.11 & 0.9055 &  & 24.90 & 0.6962 & 0.1305 & 0.0865 & 9.41 \\
 & \textbf{$\alpha$Depth} & {\color{red} 27.38} & {\color{red} 0.9094} &  & {\color{red} 28.68} & {\color{red} 0.7636} & {\color{red} 0.1190} & {\color{red} 0.0738} & {\color{red} 7.13} \\ \midrule
\multirow{2}{*}{Stereo Conversion} & MatAnyone 2 & 33.04 & 0.9312 &  & 27.22 & 0.8122 & 0.0730 & 0.0317 & 1.97 \\
 & \textbf{$\alpha$Depth} & {\color{red} 33.31} & {\color{red} 0.9326} &  & {\color{red} 28.46} & {\color{red} 0.8373} & {\color{red} 0.0699} & {\color{red} 0.0297} & {\color{red} 1.72} \\ \bottomrule[0.15em]
\end{tabular}
\end{table*}
\section{Method Analysis}
This section investigates the synergistic effect of our core components and the impact of alpha representation in stereo conversion.

\par \textbf{Synergistic Design}. Tab.~\ref{tab:synergy} shows that our circular alpha and layered representation are mutually indispensable: recovering accurate soft boundary details requires both clean layered decomposition (color and depth, provided by layered representation) and a precise blending weight (enabled by circular alpha). Adding layered representation to a vanilla alpha baseline yields only marginal gains (\eg, 0.0042 S-SSIM gain in Tab.~\ref{tab:synergy}), whereas incorporating layered representation alongside circular alpha produces a substantially larger improvement (\eg, 0.0422 S-SSIM gain). This contrast arises because the alpha valley issue degrades vanilla alpha estimation, causing blending errors that propagate through the compositing process regardless of the accuracy of the underlying layered content. Consequently, the performance benefit of layered representation only fully manifests once circular alpha resolves the alpha manifold discontinuity, making their interaction synergistic rather than additive.

\par \textbf{Alpha Representation}.  A straightforward baseline to our automatic decomposition is providing the layered pipeline with alpha mattes from a state-of-the-art matting method.
To test this, we manually annotate segmentation masks on Marvel-10K as guidance for MatAnyone 2~\cite{yang2025matanyone2} and substitute its mattes into our conversion pipeline, evaluating both the warped intermediates and the final stereo results. As shown in Tab.~\ref{tab:alphadepth-matanyone-comparison}, even with manually provided guidance, the complex cinematic scenes of Marvel-10K degrade the matting quality of MatAnyone 2, which in turn hinders accurate soft boundary synthesis in the target view. By contrast, $\alpha$Depth consistently outperforms this baseline across all metrics in both stages, despite requiring no user guidance. 

\section{Conclusion}

This paper proposes \myname, a layered representation designed to resolve depth ambiguity and color mixing at soft boundaries for stereo conversion. By leveraging our Circular Alpha Representation (CAR), \myname\ bypasses the discontinuities of vanilla alpha in complex scenes, enabling automatic scene-level decomposition in a single forward pass. Extensive experiments verify the effectiveness of CAR, showing state-of-the-art boundary fidelity of \myname\ in stereo conversion.

\small
\bibliographystyle{plain}
\bibliography{main}

\newpage
\appendix
\section*{Appendix}


We provide more technical details, experimental results, ablation studies, and qualitative visualizations to support the contributions of our \myname\ approach. Detailed contents are listed as follows:

\setlength{\cftbeforesecskip}{0.5em}
\cftsetindents{section}{0em}{1.8em}
\cftsetindents{subsection}{1em}{2.5em}
\etoctoccontentsline{part}{Appendix}
\localtableofcontents
\hypersetup{linkbordercolor=red,linkcolor=red}


\section{More Implementation Details}\label{supp-sec:implementation}


\subsection{Layered Warping with \myname\ Representation}~\label{supp-sec:layered-warping}
Previous stereo conversion approaches often employ techniques like softmax splatting~\cite{niklaus2020softmaxsplat} for view transformation~\cite{mehl2024stereoconv,shvetsova2025m2svid,mono2stereo,hairguard}. 
To support the layered representation of \myname, we extend the softmax splatting to first jointly project the foreground layer with premultiplied alpha, \ie,
\begin{equation}
    \tilde{I}_{\alpha \mathrm{FG}}, \tilde{\alpha} = \operatorname{Project}(\{\hat{\alpha} \odot \hat{I}_{\mathrm{FG}}, \hat{\alpha}\}, \hat{D}_{\mathrm{FG}}),
\end{equation}
where $\operatorname{Project}(\cdot)$ represents depth-guided softmax splatting~\cite{mehl2024stereoconv} to handle occlusions.
The joint projection ensures that $\tilde{\alpha}$ aligns with the foreground color $\tilde{I}_{\mathrm{FG}}$.
We then separately project the background layer
\begin{equation}
    \tilde{I}_{\mathrm{BG}} = \operatorname{Project}(\hat{I}_{\mathrm{BG}}, \hat{D}_{\mathrm{BG}}).
\end{equation}
Finally, we generate the warped view $\tilde{I}$ via alpha compositing on soft boundary regions:
\begin{equation}\label{eq:supp_warped_alpha_composition}
\tilde{I} = \tilde{I}_{\alpha \mathrm{FG}} + (1 - \tilde{\alpha})\tilde{I}_{\mathrm{BG}}.
\end{equation}
Since our \myname\ model only estimates layered information for soft boundary regions, with zero $\tilde{\alpha}$ for opaque regions,
Eq.~(\ref{eq:supp_warped_alpha_composition}) performs alpha composition only on soft boundaries. Thus, the warped image $\tilde{I}$ preserves the geometry estimated from state-of-the-art depth models~\cite{xu2025ppd,wang2025moge2,lin2025depthanything3}, while recovering high-fidelity structures on soft boundaries.

\subsection{Matting Evaluation Details}~\label{supp-sec:matting-eval-details}
We provide more details for the matting evaluation protocol used in Sec.~\ref{sec:main-exp-matting}. Since our \myname\ representation uses zero to represent opaque regions without differentiating foreground or background, directly computing alpha metrics cannot reflect our detail extraction performance on soft boundary regions. Thus, we first apply circular alpha encoding (Eq.~(\ref{eq:circular_alpha_encoding})) to project ground-truth alpha labels and the alpha estimation results of all baselines into the continuous trigonometric space. This maps the foreground ($\alpha=1$) and background ($\alpha=0$) to the same coordinate $(0, 1)$ in the $(\alpha_{\mathrm{sin}}, \alpha_{\mathrm{cos}})$ space, benefiting the evaluation on soft boundary regions. Then, we compute alpha metrics on the trigonometric space, \ie,
\begin{equation}
\begin{aligned}
   \operatorname{SAD} &= 0.5\cdot\operatorname{SAD}(\hat{\alpha}_{\mathrm{sin}},\alpha_{\mathrm{sin}}) + 0.5\cdot \operatorname{SAD}(\hat{\alpha}_{\mathrm{cos}},\alpha_{\mathrm{cos}}), \\
   \operatorname{Grad} &= 0.5\cdot\operatorname{Grad}(\hat{\alpha}_{\mathrm{sin}},\alpha_{\mathrm{sin}}) + 0.5\cdot \operatorname{Grad}(\hat{\alpha}_{\mathrm{cos}},\alpha_{\mathrm{cos}}), \\
   \operatorname{Conn} &= 0.5\cdot\operatorname{Conn}(\hat{\alpha}_{\mathrm{sin}},\alpha_{\mathrm{sin}}) + 0.5\cdot \operatorname{Conn}(\hat{\alpha}_{\mathrm{cos}},\alpha_{\mathrm{cos}}),
\end{aligned}
\end{equation}
where $(\hat{\alpha}_{\mathrm{sin}}, \hat{\alpha}_{\mathrm{cos}})$ and $({\alpha}_{\mathrm{sin}}, {\alpha}_{\mathrm{cos}})$ represent the estimated result and ground-truth label, respectively. $\operatorname{SAD}(\cdot), \operatorname{Grad}(\cdot),\operatorname{Conn}(\cdot)$ indicate the commonly used Sum of Absolute Differences (SAD), Gradient loss (Grad), and Connectivity loss (Conn)~\cite{yao2024vitmatte}. We also follow ViTMatte to only compute the metrics on the unknown regions of the official trimap in the matting datasets~\cite{yao2024vitmatte}. 
\par
For the input guidance to matting baselines, we employ the official trimaps for the trimap-based method ViTMatte~\cite{yao2024vitmatte}. The recent approach MatAnyone 2 requires binary segmentation masks as guidance, which are not available in AIM-500~\cite{aim_500} and P3M-10K~\cite{li2021p3m_10k}. Thus, we generate binary masks by thresholding the ground-truth alpha maps, \ie, $\mathbb{I}(0.5 \le \alpha)$. Compared with previous methods, our \myname\ model directly infers soft boundary regions from image semantics and geometry layout, achieving comparable matting performance (see Tab.~\ref{tab:main-alpha-performance}) without requiring manual guidance. 

\subsection{Training Data Curation}\label{supp-sec:train-data-curation}
We provide more implementation details in training data curation (Fig.~\ref{fig:main-train-pipeline}). Given the original foreground image $I^{\mathrm{ori}}_{\mathrm{FG}}$ (unpremultiplied, \ie, not mixed with background colors) from matting datasets and the original background image $I^{\mathrm{ori}}_{\mathrm{BG}}$ from image datasets, we first apply alpha composition to generate the input image,
\begin{equation}
    I_{\mathrm{IN}} = \alpha I^{\mathrm{ori}}_{\mathrm{FG}} + (1-\alpha) I^{\mathrm{ori}}_{\mathrm{BG}}.
\end{equation}
For the ground-truth layered color $I_{\mathrm{FG}},I_{\mathrm{BG}}$, we perform masked blending based on the binary foreground and background masks $M_{\mathrm{FG}}=\mathbb{I}(\alpha \ge \alpha_{\mathrm{th}}),M_{\mathrm{BG}}=\mathbb{I}(\alpha \ge 1-\alpha_{\mathrm{th}})$, \ie,
\begin{equation}
\begin{aligned}
    I_{\mathrm{FG}} &= M_{\mathrm{FG}} \odot I^{\mathrm{ori}}_{\mathrm{FG}} + (1 - M_{\mathrm{FG}}) \odot I^{\mathrm{ori}}_{\mathrm{BG}}, \\
    I_{\mathrm{BG}} &= M_{\mathrm{BG}} \odot I^{\mathrm{ori}}_{\mathrm{FG}} + (1 - M_{\mathrm{BG}}) \odot I^{\mathrm{ori}}_{\mathrm{BG}}.
\end{aligned}
\end{equation}
Regarding depth data generation, we first follow HairGuard to obtain high-quality depth $D^{\mathrm{ori}}_{\mathrm{FG}}, D^{\mathrm{ori}}_{\mathrm{BG}}$ using pre-trained depth models, and then generate the input depth map $D_{\mathrm{IN}}$ via depth composition~\cite{hairguard}. For the ground-truth depth labels $D_{\mathrm{FG}},D_{\mathrm{BG}}$, we perform masked blending based on the same foreground and background masks $M_{\mathrm{FG}},M_{\mathrm{BG}}$, \ie,
\begin{equation}
    \begin{aligned}
    D_{\mathrm{FG}} &= M_{\mathrm{FG}} \odot D^{\mathrm{ori}}_{\mathrm{FG}} + (1 - M_{\mathrm{FG}}) \odot D^{\mathrm{ori}}_{\mathrm{BG}}, \\
    D_{\mathrm{BG}} &= M_{\mathrm{BG}} \odot D^{\mathrm{ori}}_{\mathrm{FG}} + (1 - M_{\mathrm{BG}}) \odot D^{\mathrm{ori}}_{\mathrm{BG}}.
\end{aligned}
\end{equation}
As shown in Fig.~\ref{fig:main-train-pipeline}, the generated $I_{\mathrm{FG}}/D_{\mathrm{FG}}$ and $I_{\mathrm{BG}}/D_{\mathrm{BG}}$ preserve foreground and background information on soft boundary regions, respectively.

\subsection{Matting Loss $\mathcal{L}_{\mathrm{m}}$}
To enhance the detail extraction performance of \myname, we adopt the matting loss $\mathcal{L}_{\mathrm{m}}$ introduced in ViTMatte~\cite{yao2024vitmatte}. Specifically, $\mathcal{L}_{\mathrm{m}}$ is composed of three terms:
\begin{equation}
\mathcal{L}_{\mathrm{m}} = \mathcal{L}_{1} + \mathcal{L}_{lap} + \mathcal{L}_{gp},
\end{equation}
where $\mathcal{L}_{lap}$ and $\mathcal{L}_{gp}$ denote the Laplacian loss~\cite{lischke2020laploss} and the gradient loss~\cite{dai2022gploss}, respectively. Since this matting loss is designed for single-channel data (\eg, alpha mattes), it can be directly applied to supervise our depth predictions. For the foreground and background color outputs, we compute $\mathcal{L}_{\mathrm{m}}$ independently across each color channel and average the results.

\section{More Experimental Results}\label{supp-sec:more-exp}

\begin{table*}[t]
\centering
\caption{\textbf{Pixel-level metrics on full image}. The \textcolor{red}{best} and \textcolor{blue}{second best} results are marked.}
\vspace{-0.5em}
\label{tab:supp-full-pixel}
\footnotesize
\begin{tabular}{lcc c cc}
\toprule[0.15em]
\rowcolor{color3} & \multicolumn{2}{c}{\textbf{Mono2Stereo~\cite{mono2stereo}}} &  & \multicolumn{2}{c}{\textbf{Marvel-10K~\cite{hairguard}}} \\
\cline{2-3} \cline{5-6}
\rowcolor{color3} \multirow{-2}{*}{\textbf{Method}} & PSNR $\uparrow$ & SSIM $\uparrow$ &  & PSNR $\uparrow$ & SSIM $\uparrow$ \\ \midrule
\rowcolor{color3} \textbf{Stereo Conversion} & & & & &
\\
StereoDiffusion~\cite{stereodiffusion} & 26.34 & 0.7541 &  & 26.04 & 0.7656 \\
Mono2Stereo~\cite{mono2stereo} & 28.15 & 0.7980 &  & 28.19 & 0.8132 \\
StereoCrafter~\cite{stereocrafter} & 28.06 & 0.7855 &  & 29.27 & 0.8340 \\
SplatDiff~\cite{splatdiff} & 30.40 & 0.8909 &  & 32.37 & 0.9284 \\
HairGuard~\cite{hairguard} & {\color{blue} 30.88} & {\color{blue} 0.8911} &  & {\color{blue} 33.09} & {\color{blue} 0.9316} \\
\textbf{\myname\ (Ours)} & {\color{red} 31.03} & {\color{red} 0.8942} &  & {\color{red} 33.31} & {\color{red} 0.9326} \\ \midrule
\rowcolor{color3} \textbf{Warping Performance} & & & & &
\\
DAv2~\cite{yang2024depthanythingv2} / VDA~\cite{chen2025videodepthanything} & {\color{blue} 29.35} & 0.8888  & & 27.04 & 0.9055 \\
DAv2 / VDA + HairGuard~\cite{hairguard} & {\color{blue} 29.35} & {\color{blue} 0.8889} & & {\color{blue} 27.06} & {\color{blue} 0.9068} \\
\textbf{DAv2 / VDA + \myname\ (Ours)} & {\color{red} 29.38} & {\color{red} 0.8891} & &{\color{red} 27.38} & {\color{red} 0.9094} \\ 
\bottomrule[0.15em]
\end{tabular}
\end{table*}

\begin{table*}[t]
\centering
\caption{\textbf{Statistical confidence analysis of full-image metrics on stereo conversion. } $\bar{\Delta}$ represents the mean difference with its 95\% paired-bootstrap Confidence Interval (CI). We also report the one-sided Wilcoxon signed-rank p-value, and the fraction of samples where our method is better, \ie, the win rate (Win\%).}
\label{tab:alphadepth-statistical-confidence}
\footnotesize
\renewcommand{\arraystretch}{1.15}
\begin{tabular}{lccc c}
\toprule[0.15em]
\rowcolor{color3} \textbf{Dataset} & \textbf{Metric} & $\bar{\Delta}$ (95\% CI) & $p$ (Wilcoxon) & Win\% \\
\midrule
\multirow{4}{*}{Marvel-10K}
& PSNR$\uparrow$  & $+0.2280\;[+0.1360, +0.3220]$  & $9.0\times10^{-10}$  & 62\% \\
& SSIM$\uparrow$  & $+0.0010\;[+0.0006, +0.0013]$ & $8.7\times10^{-9}$   & 62\% \\
& LPIPS$\downarrow$ & $-0.0035\;[-0.0040, -0.0031]$ & $6.3\times10^{-43}$  & 78\% \\
& DISTS$\downarrow$ & $-0.0025\;[-0.0028, -0.0021]$ & $5.9\times10^{-39}$  & 78\% \\
\midrule
\multirow{4}{*}{Mono2Stereo}
& PSNR$\uparrow$  & $+0.1420\;[+0.1210, +0.1620]$  & $3.6\times10^{-129}$ & 76\% \\
& SSIM$\uparrow$  & $+0.0031\;[+0.0029, +0.0032]$ & $4.9\times10^{-281}$ & 86\% \\
& LPIPS$\downarrow$ & $-0.0031\;[-0.0033, -0.0029]$ & $9.2\times10^{-191}$ & 76\% \\
& DISTS$\downarrow$ & $-0.0027\;[-0.0029, -0.0025]$ & $1.5\times10^{-186}$ & 75\% \\
\bottomrule[0.15em]
\end{tabular}
\end{table*}

\subsection{Pixel-Level Metrics on Full Image}
Following the experimental settings in Tabs.~\ref{tab:main-stereo-conv-full} and \ref{tab:main-warping-performance}, we additionally provide pixel-level metrics computed on the full image.
Tab.~\ref{tab:supp-full-pixel} verifies the state-of-the-art performance of \myname\ in both warping and stereo conversion. To assess the reliability of the reported improvements, we further perform a paired statistical analysis against HairGuard on stereo conversion. Since all methods are evaluated on identical inputs, we form per-sample paired differences $\Delta_i = \operatorname{Metric}_i(\text{Ours}) - \operatorname{Metric}_i(\text{HairGuard})$ for each metric and report three complementary statistics: (i)~a 95\% confidence interval on the mean difference $\bar{\Delta}$, obtained by the percentile paired bootstrap with $10{,}000$ resamples drawn over samples
with replacement; (ii)~a one-sided Wilcoxon signed-rank
test, which is distribution-free and robust to the heavy-tailed difference distributions; and (iii)~the win rate, \ie, the fraction of samples on which our method outperforms HairGuard. All analyses use a fixed random seed, and the direction
of the one-sided tests follows each metric's polarity (higher is better for PSNR/SSIM, lower for LPIPS/DISTS). As shown in Tab.~\ref{tab:alphadepth-statistical-confidence}, all 95\% confidence intervals exclude zero, all Wilcoxon p-values are below $10^{-8}$, and our method wins on 62–86\% of individual samples, confirming that the improvements are statistically reliable and consistent.

\begin{table*}[t]
\centering
\caption{\textbf{Comparison of temporal stability and stereo comfort metrics} on the Marvel-10K dataset. The \textcolor{red}{best} and \textcolor{blue}{second best} results are marked.}
\label{tab:alphadepth-temporal-stereo}
\footnotesize
\begin{tabular}{lcc c ccc}
\toprule[0.15em]
\rowcolor{color3} {\textbf{Method}} & $E_{\text{warp}}$ $\downarrow$ & RFlick-B $\downarrow$ &  & $E_{\text{disp}}$ $\downarrow$ & TDV $\downarrow$ & VDisp $\downarrow$ \\ \midrule
StereoCrafter & 0.681 & 0.547 &  & 18.56 & 5.309 & 0.285 \\
SplatDiff & 0.713 & 0.466 &  & 7.29 & 2.242 & 0.078 \\
HairGuard & {\color{blue} 0.589} & {\color{blue} 0.464} &  & {\color{blue} 6.87} & {\color{blue} 2.044} & {\color{blue} 0.077} \\
\rowcolor{color3} \textbf{\myname\ (Ours)} & {\color{red} 0.588} & {\color{red} 0.445} &  & {\color{red} 6.58} & {\color{red} 2.029} & {\color{red} 0.075} \\ \bottomrule[0.15em]
\end{tabular}
\end{table*}
\subsection{Temporal Stability and Stereo Comfort}

Beyond per-frame quality, temporal stability and stereo viewing comfort are crucial for high-quality stereo conversion. Let $\hat{I}_t$ and $I_t$ denote the synthesized view and the ground-truth view of frame $t$, respectively. We first assess temporal stability using video-oriented metrics:
\begin{itemize}
    \item \textbf{Warping Error ($E_{\text{warp}}$)}. Temporal flicker is evaluated via the warping error, which measures photometric agreement between consecutive predicted frames after motion compensation, \ie,
    \begin{equation}
    E_{\text{warp}} \;=\; \frac{1}{|\mathcal{V}|} \sum_{\mathbf{p} \in \mathcal{V}} \big\| \hat{I}_{t+1}(\mathbf{p}) - \overrightarrow{\hat{I}_{t}}(\mathbf{p}) \big\|_1,
    \label{eq:warp-err}
    \end{equation}
    where $\overrightarrow{\hat{I}_{t}}$ denotes frame $t$ warped to frame $t{+}1$ with optical flow estimated on the reference sequence, $\mathbf{p} = (u, v)^\top$ indicates pixel coordinates, and $\mathcal{V}$ excludes occluded pixels.
    \item \textbf{Boundary Residual Flicker (RFlick-B)}. Boundary regions of synthesized videos are prone to temporal flicker, and thus we define a boundary residual flicker metric RFlick-B for boundary stability evaluation. Let $B_t$ denote the reference disparity edge region of frame $t$, and let
    \begin{equation}
        r_t \;=\; \frac{1}{|B_t|} \sum_{\mathbf{p} \in B_t} \big\| \hat{I}_t(\mathbf{p}) - I_t(\mathbf{p}) \big\|_1
        \label{eq:boundary-residual}
    \end{equation}
    be the mean boundary residual at frame $t$. RFlick-B is then the temporal standard deviation of these residuals,
    \begin{equation}
        \operatorname{RFlick\text{-}B} \;=\; \sqrt{ \frac{1}{T} \sum_{t=1}^{T} \big( r_t - \bar{r} \big)^2 },
        \quad
        \bar{r} \;=\; \frac{1}{T} \sum_{t=1}^{T} r_t,
        \label{eq:rflickb}
    \end{equation}
    where $T$ is the number of frames. The metric focuses on the temporal instability of boundary errors, which often introduces perceptible visual artifacts, such as flickering hair.
\end{itemize}
\par 
Since image-level metrics are largely insensitive to disparity errors (\eg, a view with a plausible appearance but wrong parallax scores well on PSNR), we employ stereo metrics that extract disparity from the synthesized pair with an off-the-shelf stereo matcher~\cite{wen2025foundationstereo} and assess it directly. In the following, $\hat{\delta}_t$ and $\delta_t$ denote the disparities estimated from the generated and reference stereo pairs of frame $t$, respectively.
\begin{itemize}
    \item \textbf{Disparity Error ($E_{\text{disp}}$)}. We follow Elastic3D~\cite{metzger2025elastic3d} to measure the geometric accuracy of the induced parallax as the mean absolute deviation between the two, \ie,
    \begin{equation}
        E_{\text{disp}} \;=\; \frac{1}{THW} \sum_{t=1}^{T} \sum_{v=1}^{H} \sum_{u=1}^{W} \big| \hat{\delta}_t(\mathbf{p}) - \delta_t(\mathbf{p}) \big|,
        \label{eq:edisp}
    \end{equation}
    directly quantifying whether the synthesized view places content at the intended depth. 
    \item \textbf{Temporal Disparity Variation (TDV)}. Temporal fluctuations of the overall depth placement are known to induce visual discomfort~\cite{speranza2006effect}, and thus we quantify them as the temporal disparity variation TDV, \ie,
    \begin{equation}
        \operatorname{TDV} \;=\; \frac{1}{T - 1} \sum_{t=2}^{T} \big| \bar{\delta}_t - \bar{\delta}_{t-1} \big|,
        \label{eq:tdv}
    \end{equation}
    where $\bar{\delta}_t$ denotes the spatial mean of $\hat{\delta}_t$. 
    \item \textbf{Vertical Disparity (VDisp)}. A rectified stereo pair must contain purely horizontal parallax, and any vertical component violates epipolar geometry and is a common source of eye strain~\cite{kooi2004visual}. We measure it as the mean vertical disparity VDisp:
    \begin{equation}
        \operatorname{VDisp} \;=\; \frac{1}{THW} \sum_{t=1}^{T} \sum_{v=1}^{H} \sum_{u=1}^{W} \big| F^{y}_{t}(\mathbf{p}) \big|,
        \label{eq:vdisp}
    \end{equation}
    where $F^{y}_{t}$ denotes the vertical component of the optical flow estimated between the left and right views of frame $t$.
\end{itemize}
As presented in Tab.~\ref{tab:alphadepth-temporal-stereo}, $\alpha$Depth achieves superior performance across all five metrics. The margin over HairGuard is particularly pronounced on $\text{RFlick-B}$ and $E_{\text{disp}}$, demonstrating that our layered representation effectively stabilizes high-frequency boundary content and maps it to more plausible disparities.


\begin{figure}[t]
  \centering
  \begin{subfigure}[b]{0.165\textwidth}
    \centering
    \includegraphics[width=\linewidth]{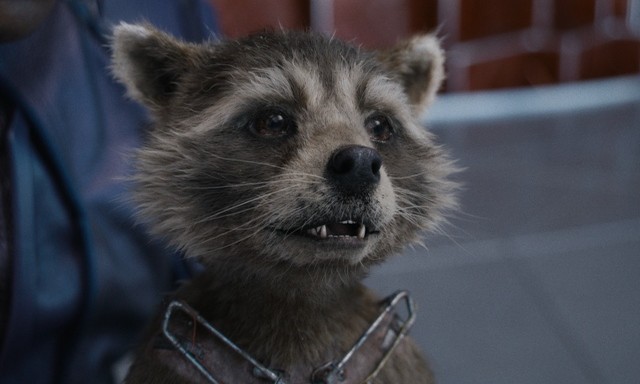}
  \end{subfigure}\hfill
  \begin{subfigure}[b]{0.165\textwidth}
    \centering
    \includegraphics[width=\linewidth]{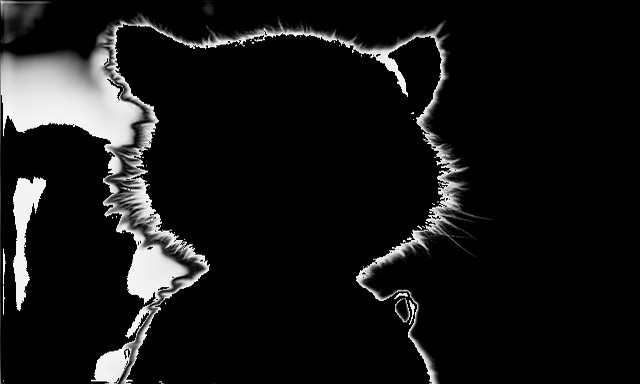}
  \end{subfigure}\hfill
  \begin{subfigure}[b]{0.165\textwidth}
    \centering
    \includegraphics[width=\linewidth]{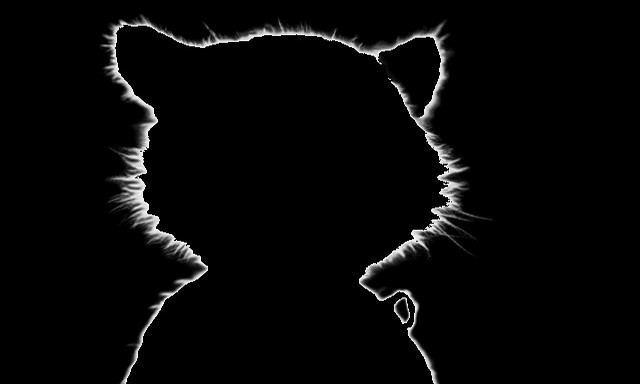}
  \end{subfigure}\hfill
  \begin{subfigure}[b]{0.165\textwidth}
    \centering
    \includegraphics[width=\linewidth]{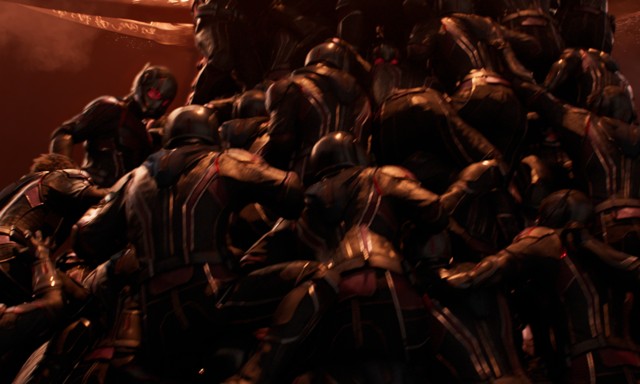}
  \end{subfigure}\hfill
  \begin{subfigure}[b]{0.165\textwidth}
    \centering
    \includegraphics[width=\linewidth]{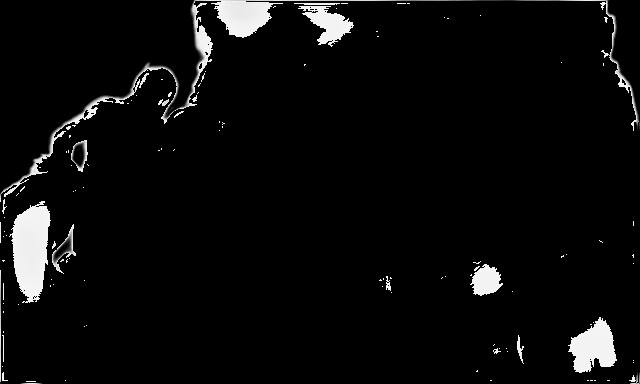}
  \end{subfigure}\hfill
  \begin{subfigure}[b]{0.165\textwidth}
    \centering
    \includegraphics[width=\linewidth]{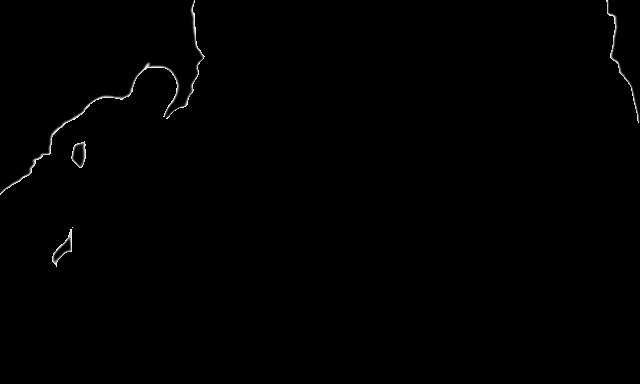}
  \end{subfigure}
  \begin{subfigure}[b]{0.165\textwidth}
    \centering
    \includegraphics[width=\linewidth]{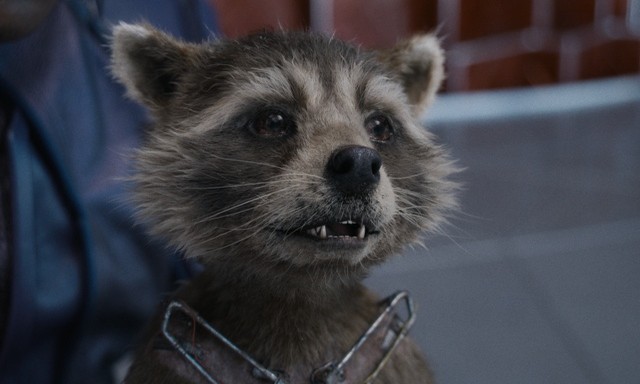}
  \end{subfigure}\hfill
  \begin{subfigure}[b]{0.165\textwidth}
    \centering
    \includegraphics[width=\linewidth]{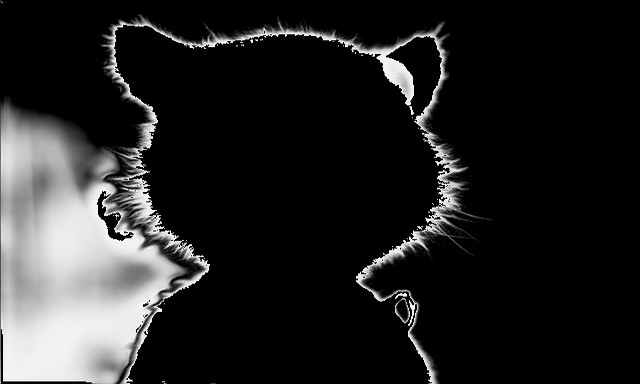}
  \end{subfigure}\hfill
  \begin{subfigure}[b]{0.165\textwidth}
    \centering
    \includegraphics[width=\linewidth]{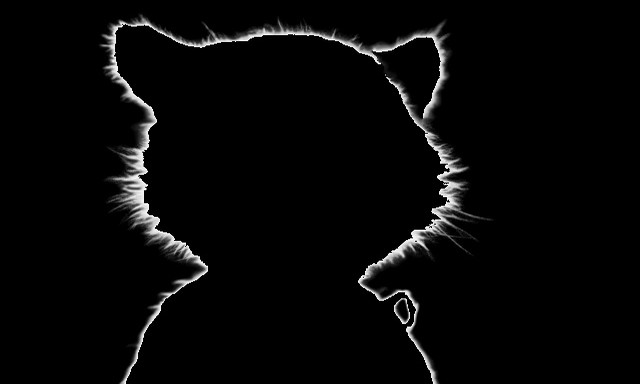}
  \end{subfigure}\hfill
  \begin{subfigure}[b]{0.165\textwidth}
    \centering
    \includegraphics[width=\linewidth]{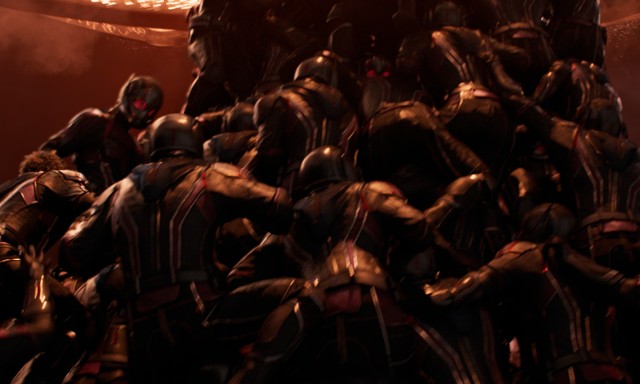}
  \end{subfigure}\hfill
  \begin{subfigure}[b]{0.165\textwidth}
    \centering
    \includegraphics[width=\linewidth]{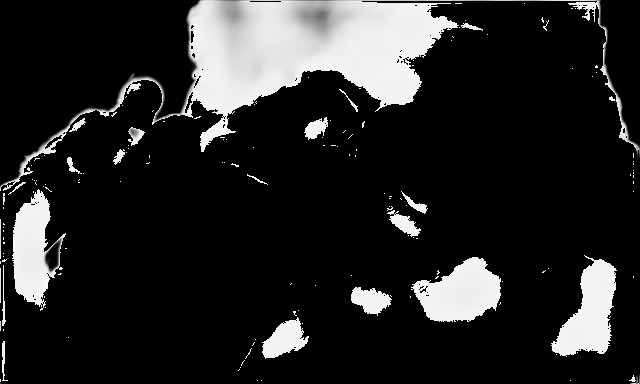}
  \end{subfigure}\hfill
  \begin{subfigure}[b]{0.165\textwidth}
    \centering
    \includegraphics[width=\linewidth]{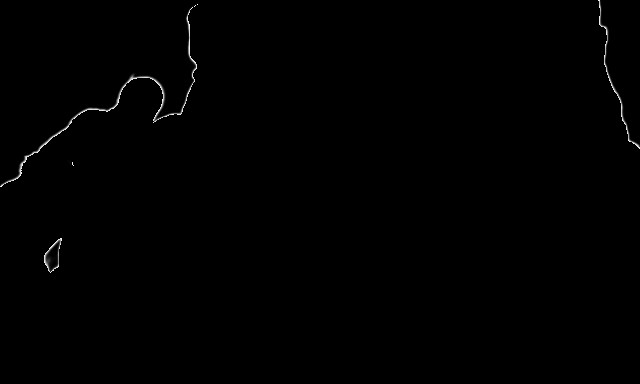}
  \end{subfigure}
  \begin{subfigure}[b]{0.165\textwidth}
    \centering
    \includegraphics[width=\linewidth]{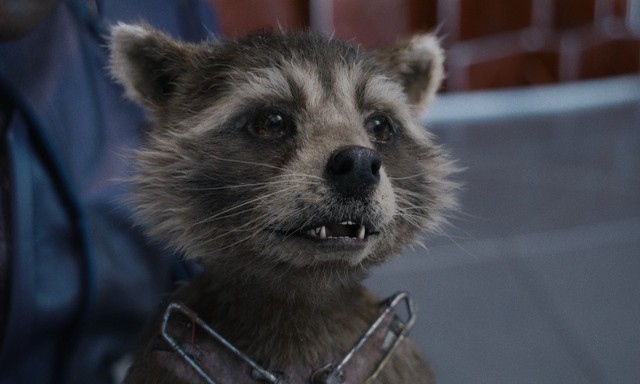}
  \end{subfigure}\hfill
  \begin{subfigure}[b]{0.165\textwidth}
    \centering
    \includegraphics[width=\linewidth]{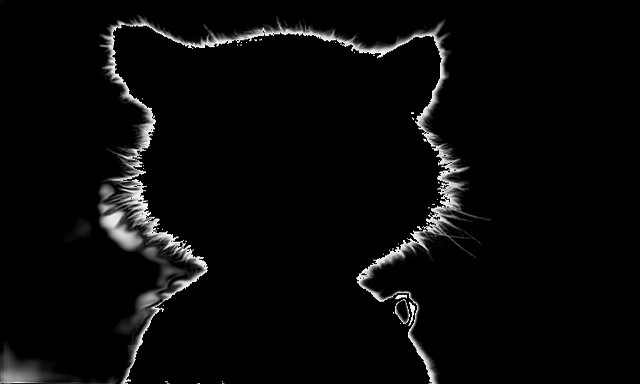}
  \end{subfigure}\hfill
  \begin{subfigure}[b]{0.165\textwidth}
    \centering
    \includegraphics[width=\linewidth]{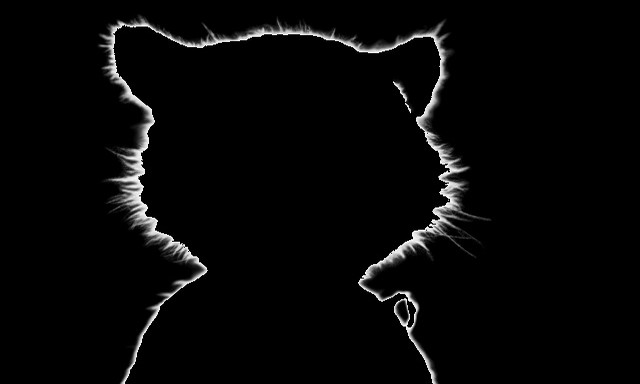}
  \end{subfigure}\hfill
  \begin{subfigure}[b]{0.165\textwidth}
    \centering
    \includegraphics[width=\linewidth]{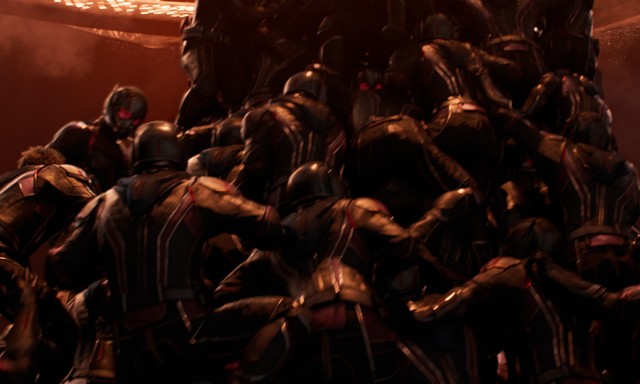}
  \end{subfigure}\hfill
  \begin{subfigure}[b]{0.165\textwidth}
    \centering
    \includegraphics[width=\linewidth]{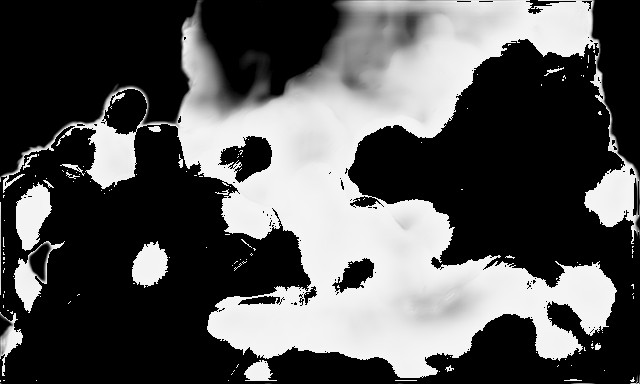}
  \end{subfigure}\hfill
  \begin{subfigure}[b]{0.165\textwidth}
    \centering
    \includegraphics[width=\linewidth]{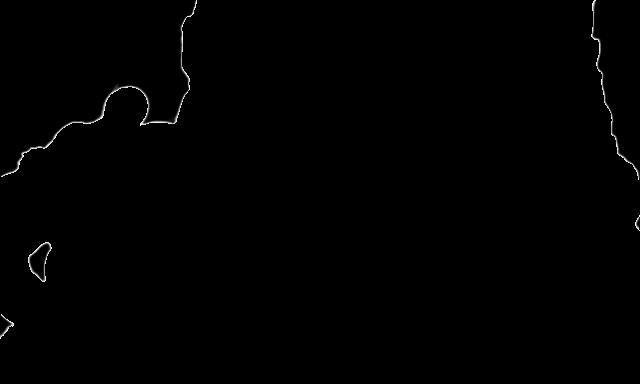}
  \end{subfigure}
  \begin{subfigure}[b]{0.165\textwidth}
    \centering
    \includegraphics[width=\linewidth]{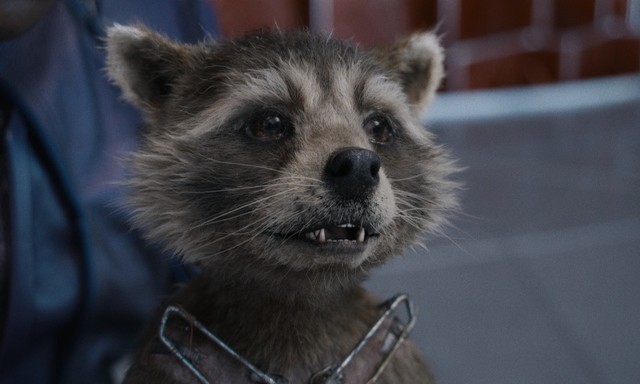}
  \end{subfigure}\hfill
  \begin{subfigure}[b]{0.165\textwidth}
    \centering
    \includegraphics[width=\linewidth]{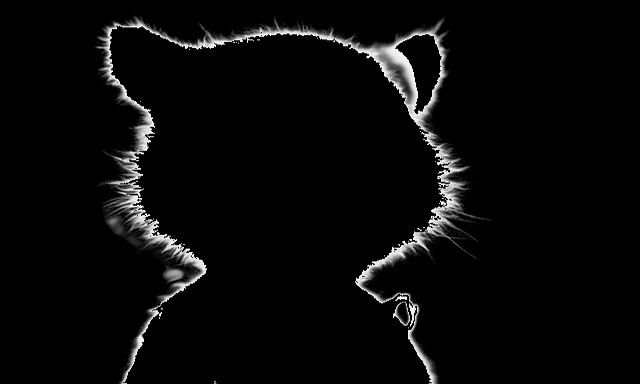}
  \end{subfigure}\hfill
  \begin{subfigure}[b]{0.165\textwidth}
    \centering
    \includegraphics[width=\linewidth]{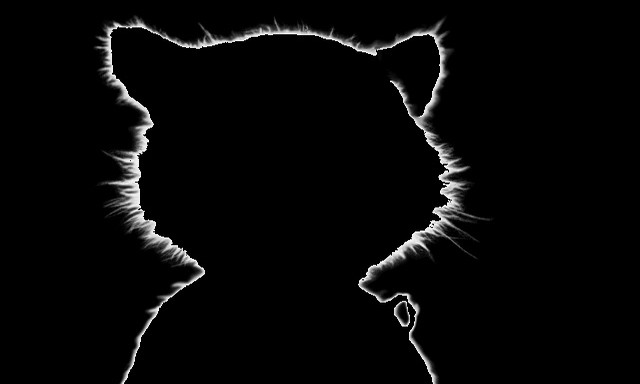}
  \end{subfigure}\hfill
  \begin{subfigure}[b]{0.165\textwidth}
    \centering
    \includegraphics[width=\linewidth]{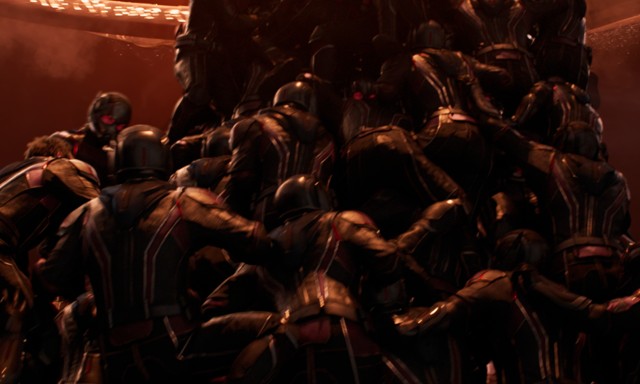}
  \end{subfigure}\hfill
  \begin{subfigure}[b]{0.165\textwidth}
    \centering
    \includegraphics[width=\linewidth]{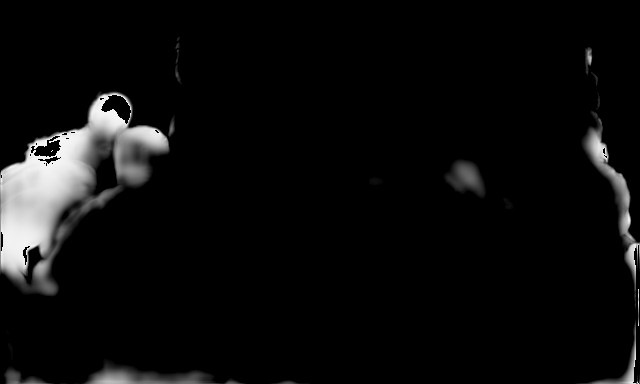}
  \end{subfigure}\hfill
  \begin{subfigure}[b]{0.165\textwidth}
    \centering
    \includegraphics[width=\linewidth]{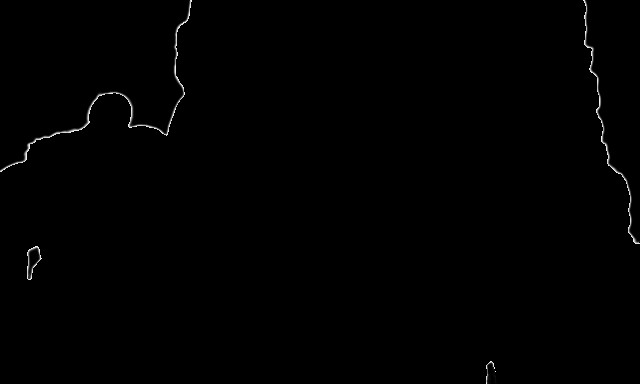}
  \end{subfigure}
  \begin{subfigure}[b]{0.165\textwidth}
    \centering
    \includegraphics[width=\linewidth]{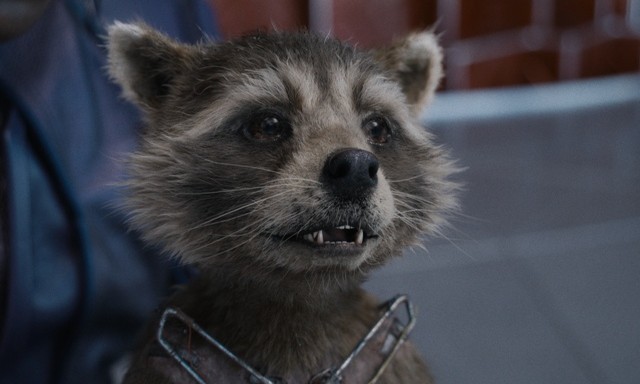}
  \end{subfigure}\hfill
  \begin{subfigure}[b]{0.165\textwidth}
    \centering
    \includegraphics[width=\linewidth]{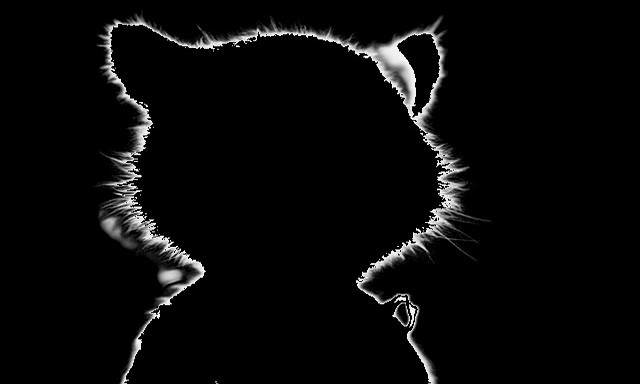}
  \end{subfigure}\hfill
  \begin{subfigure}[b]{0.165\textwidth}
    \centering
    \includegraphics[width=\linewidth]{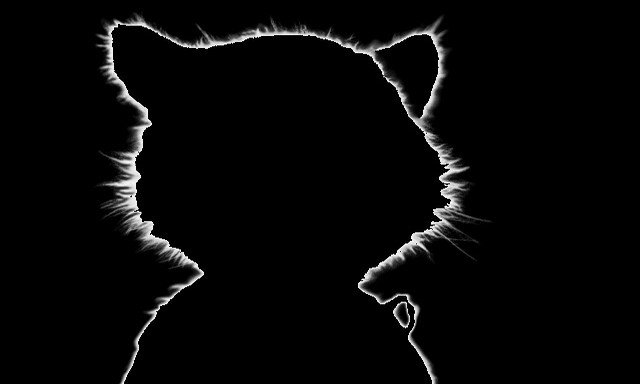}
  \end{subfigure}\hfill
  \begin{subfigure}[b]{0.165\textwidth}
    \centering
    \includegraphics[width=\linewidth]{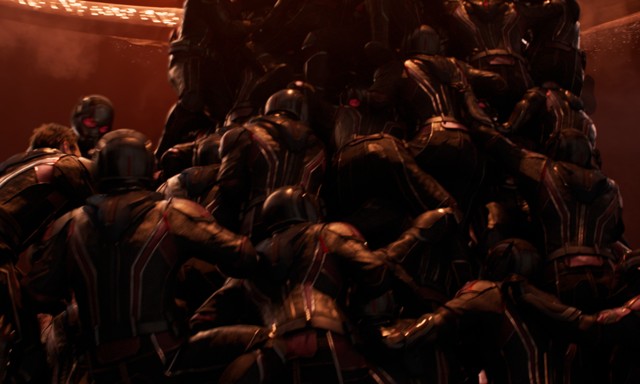}
  \end{subfigure}\hfill
  \begin{subfigure}[b]{0.165\textwidth}
    \centering
    \includegraphics[width=\linewidth]{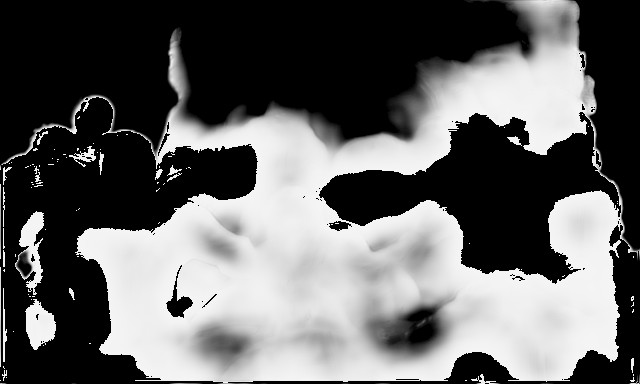}
  \end{subfigure}\hfill
  \begin{subfigure}[b]{0.165\textwidth}
    \centering
    \includegraphics[width=\linewidth]{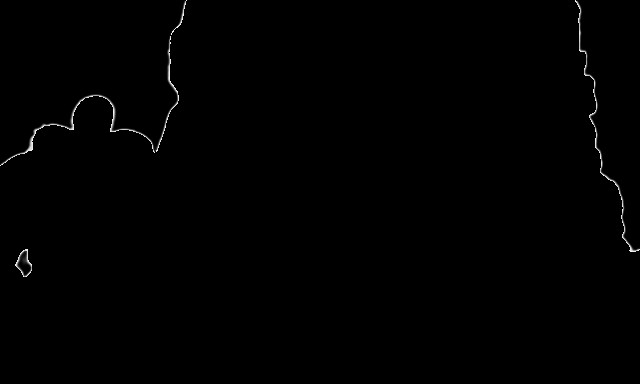}
  \end{subfigure}
  \begin{subfigure}[b]{0.165\textwidth}
    \centering
    \includegraphics[width=\linewidth]{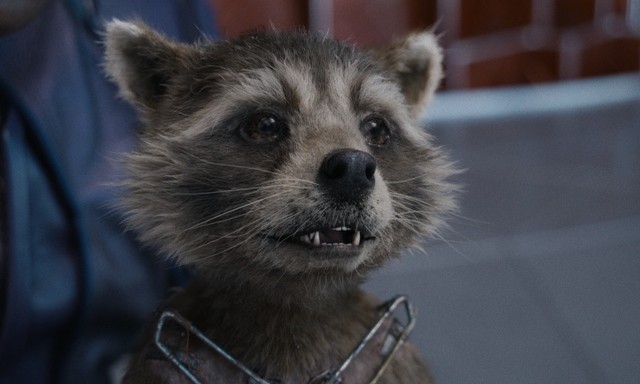}
  \end{subfigure}\hfill
  \begin{subfigure}[b]{0.165\textwidth}
    \centering
    \includegraphics[width=\linewidth]{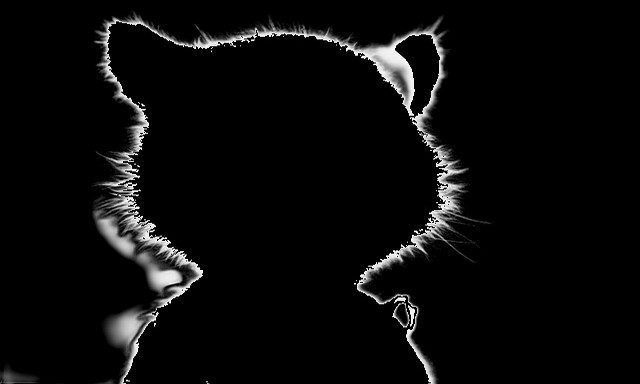}
  \end{subfigure}\hfill
  \begin{subfigure}[b]{0.165\textwidth}
    \centering
    \includegraphics[width=\linewidth]{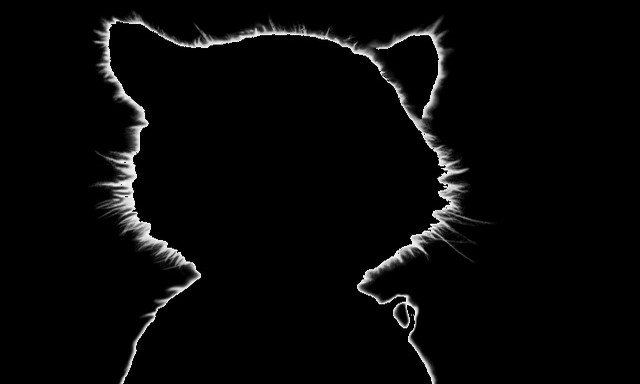}
  \end{subfigure}\hfill
  \begin{subfigure}[b]{0.165\textwidth}
    \centering
    \includegraphics[width=\linewidth]{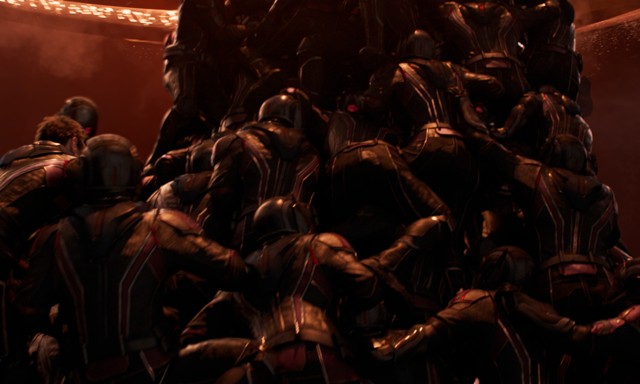}
  \end{subfigure}\hfill
  \begin{subfigure}[b]{0.165\textwidth}
    \centering
    \includegraphics[width=\linewidth]{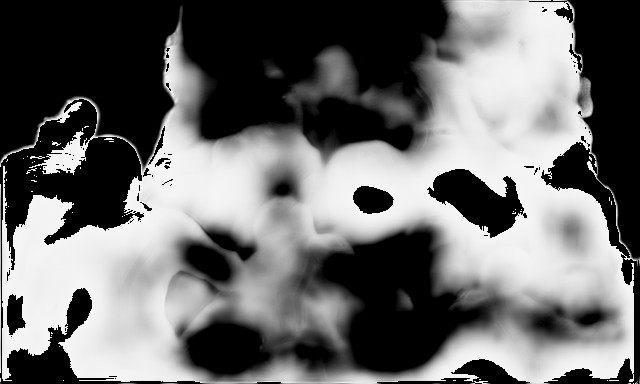}
  \end{subfigure}\hfill
  \begin{subfigure}[b]{0.165\textwidth}
    \centering
    \includegraphics[width=\linewidth]{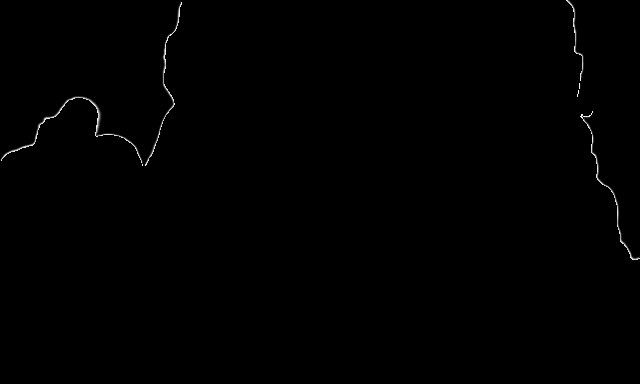}
  \end{subfigure}
  \begin{subfigure}[b]{0.165\textwidth}
    \centering
    \includegraphics[width=\linewidth]{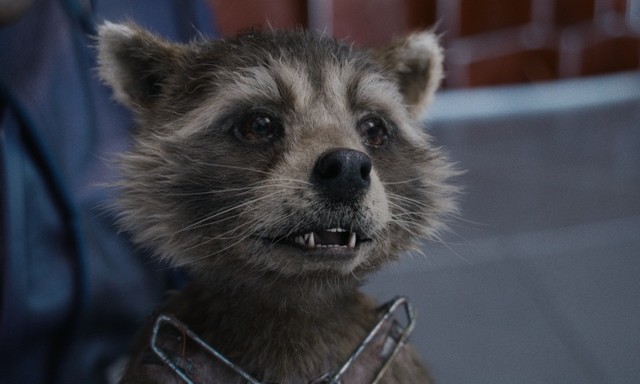}
  \end{subfigure}\hfill
  \begin{subfigure}[b]{0.165\textwidth}
    \centering
    \includegraphics[width=\linewidth]{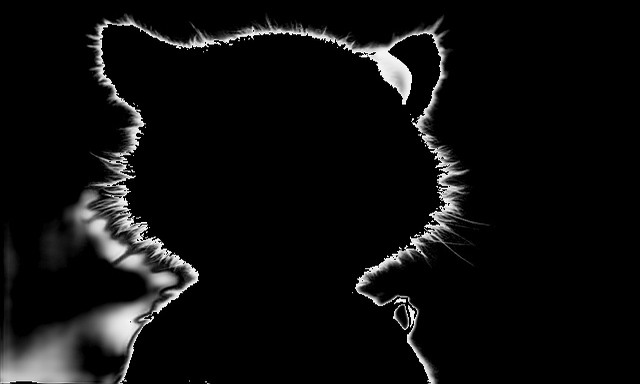}
  \end{subfigure}\hfill
  \begin{subfigure}[b]{0.165\textwidth}
    \centering
    \includegraphics[width=\linewidth]{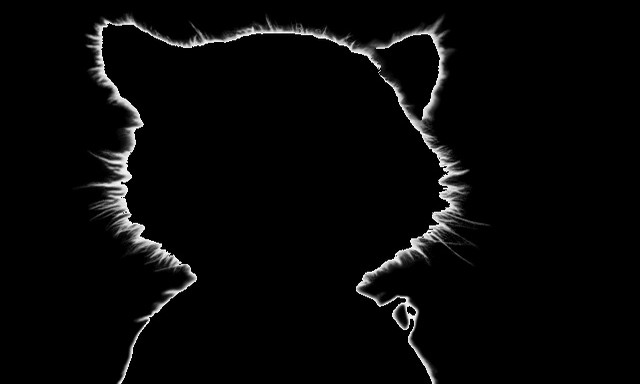}
  \end{subfigure}\hfill
  \begin{subfigure}[b]{0.165\textwidth}
    \centering
    \includegraphics[width=\linewidth]{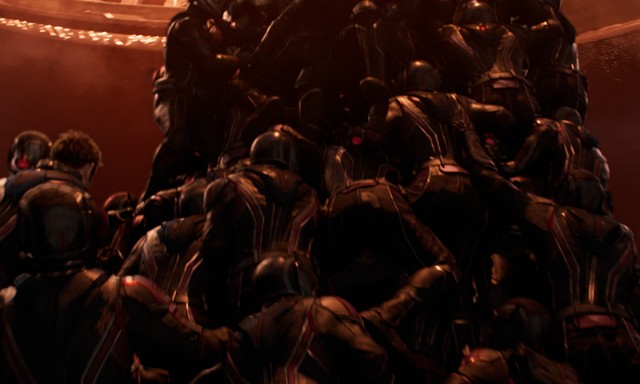}
  \end{subfigure}\hfill
  \begin{subfigure}[b]{0.165\textwidth}
    \centering
    \includegraphics[width=\linewidth]{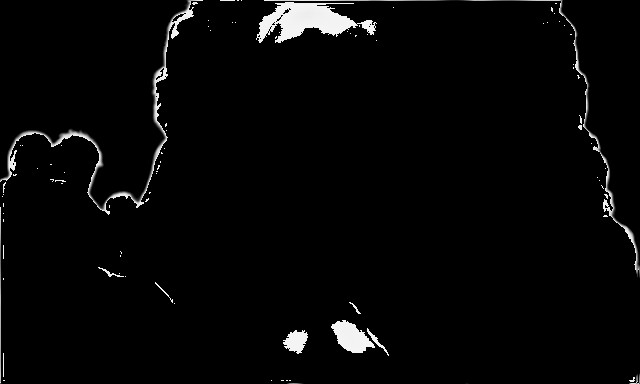}
  \end{subfigure}\hfill
  \begin{subfigure}[b]{0.165\textwidth}
    \centering
    \includegraphics[width=\linewidth]{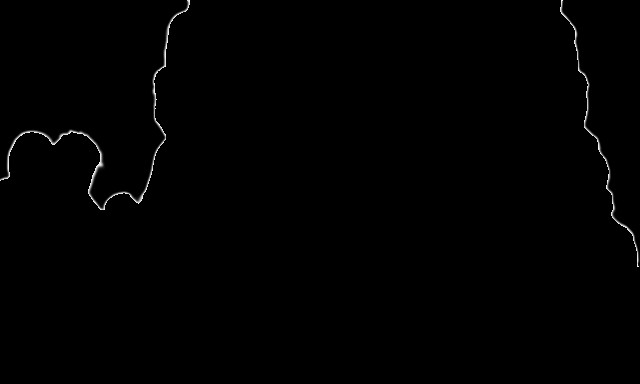}
  \end{subfigure}
  \begin{subfigure}[b]{0.165\textwidth}
    \centering
    \includegraphics[width=\linewidth]{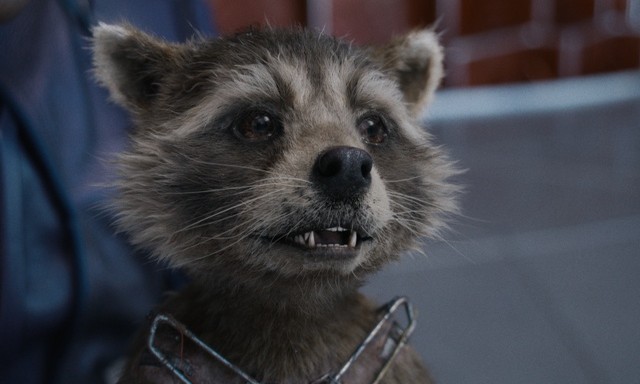}
  \end{subfigure}\hfill
  \begin{subfigure}[b]{0.165\textwidth}
    \centering
    \includegraphics[width=\linewidth]{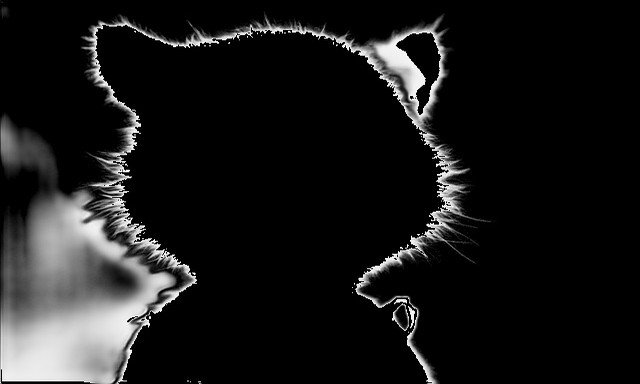}
  \end{subfigure}\hfill
  \begin{subfigure}[b]{0.165\textwidth}
    \centering
    \includegraphics[width=\linewidth]{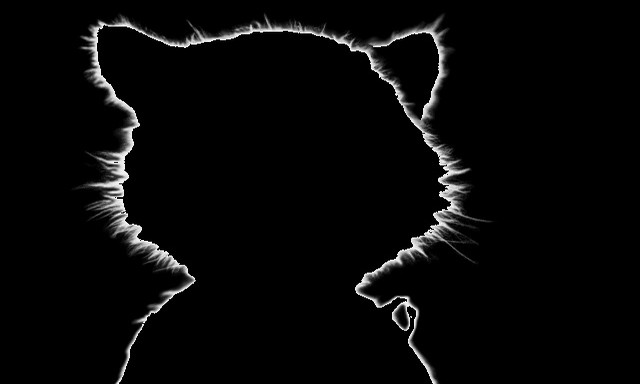}
  \end{subfigure}\hfill
  \begin{subfigure}[b]{0.165\textwidth}
    \centering
    \includegraphics[width=\linewidth]{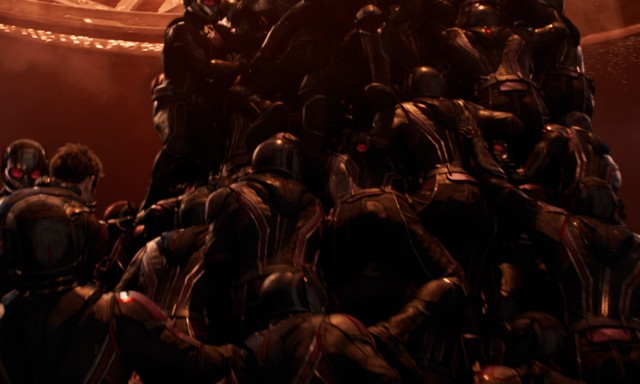}
  \end{subfigure}\hfill
  \begin{subfigure}[b]{0.165\textwidth}
    \centering
    \includegraphics[width=\linewidth]{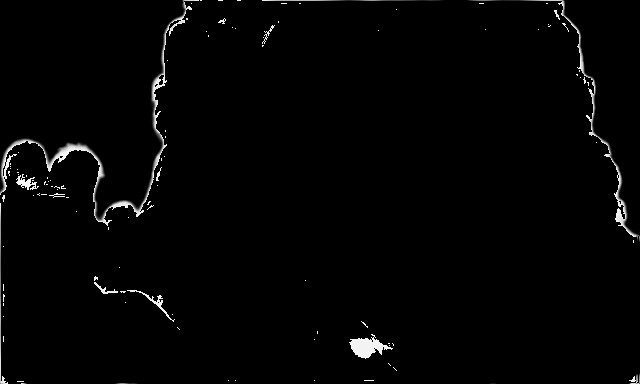}
  \end{subfigure}\hfill
  \begin{subfigure}[b]{0.165\textwidth}
    \centering
    \includegraphics[width=\linewidth]{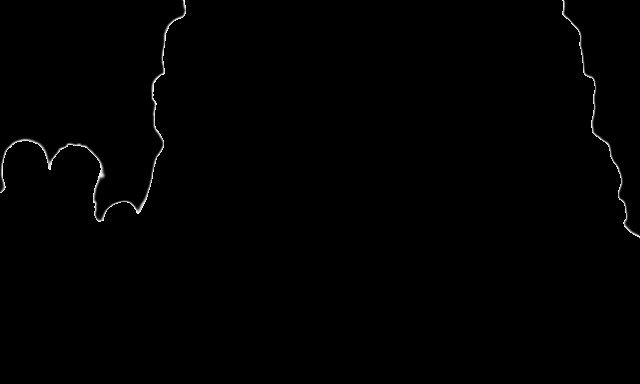}
  \end{subfigure}
  \begin{subfigure}[b]{0.165\textwidth}
    \centering
    \includegraphics[width=\linewidth]{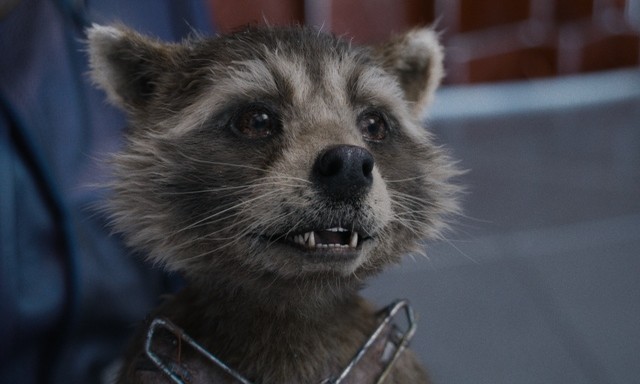}
  \end{subfigure}\hfill
  \begin{subfigure}[b]{0.165\textwidth}
    \centering
    \includegraphics[width=\linewidth]{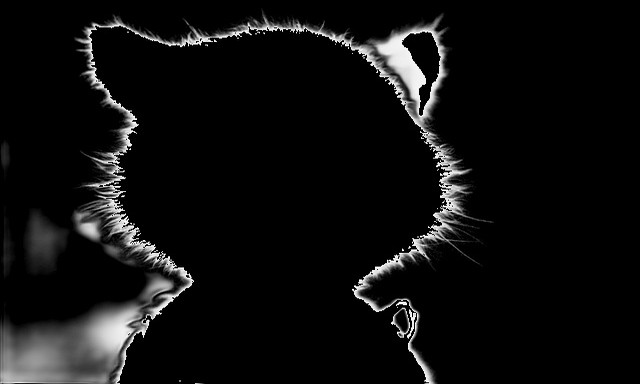}
  \end{subfigure}\hfill
  \begin{subfigure}[b]{0.165\textwidth}
    \centering
    \includegraphics[width=\linewidth]{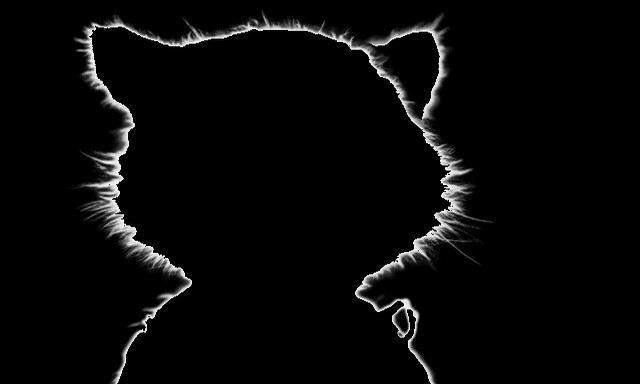}
  \end{subfigure}\hfill
  \begin{subfigure}[b]{0.165\textwidth}
    \centering
    \includegraphics[width=\linewidth]{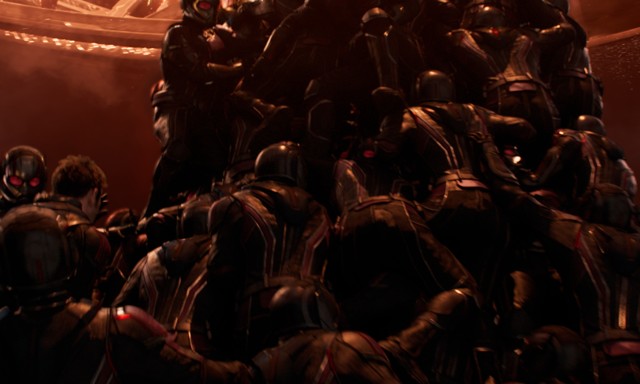}
  \end{subfigure}\hfill
  \begin{subfigure}[b]{0.165\textwidth}
    \centering
    \includegraphics[width=\linewidth]{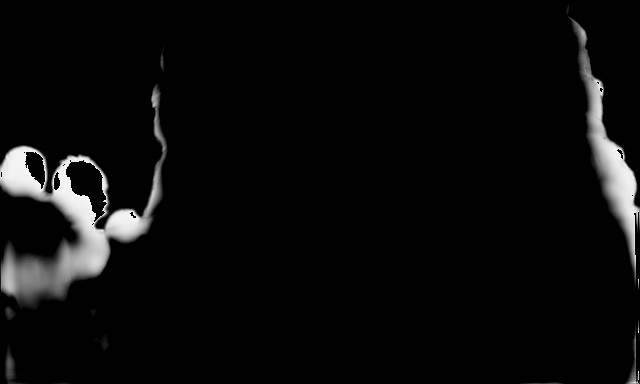}
  \end{subfigure}\hfill
  \begin{subfigure}[b]{0.165\textwidth}
    \centering
    \includegraphics[width=\linewidth]{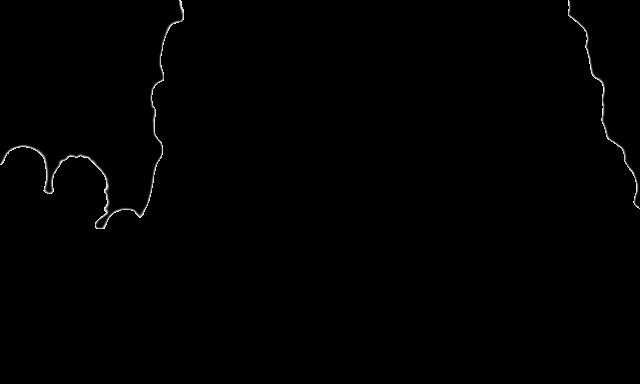}
  \end{subfigure}
  \begin{subfigure}[b]{0.165\textwidth}
    \centering
    \includegraphics[width=\linewidth]{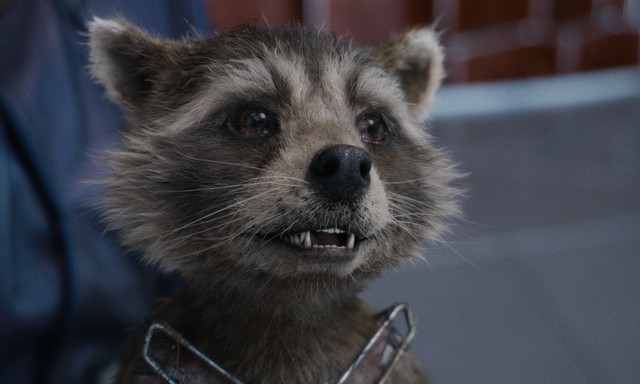}
    \caption*{Input Video}
  \end{subfigure}\hfill
  \begin{subfigure}[b]{0.165\textwidth}
    \centering
    \includegraphics[width=\linewidth]{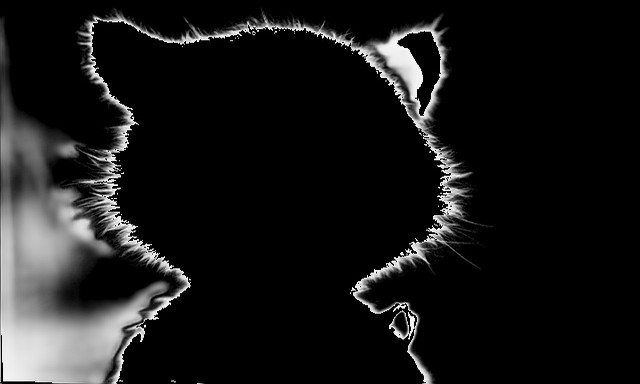}
    \caption*{Vanilla Alpha}
  \end{subfigure}\hfill
  \begin{subfigure}[b]{0.165\textwidth}
    \centering
    \includegraphics[width=\linewidth]{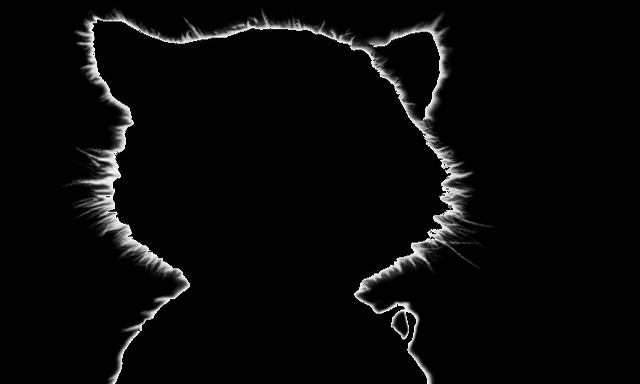}
    \caption*{\textbf{CAR (Ours)}}
  \end{subfigure}\hfill
  \begin{subfigure}[b]{0.165\textwidth}
    \centering
    \includegraphics[width=\linewidth]{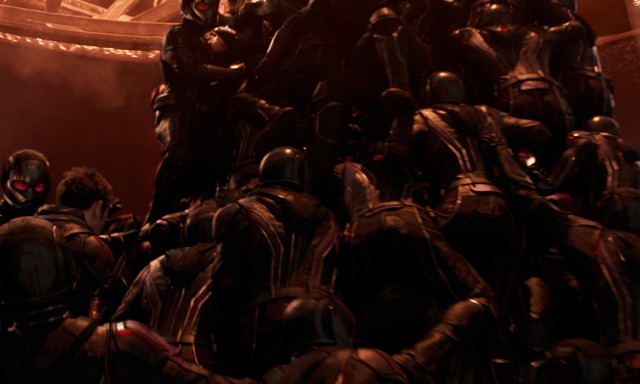}
    \caption*{Input Video}
  \end{subfigure}\hfill
  \begin{subfigure}[b]{0.165\textwidth}
    \centering
    \includegraphics[width=\linewidth]{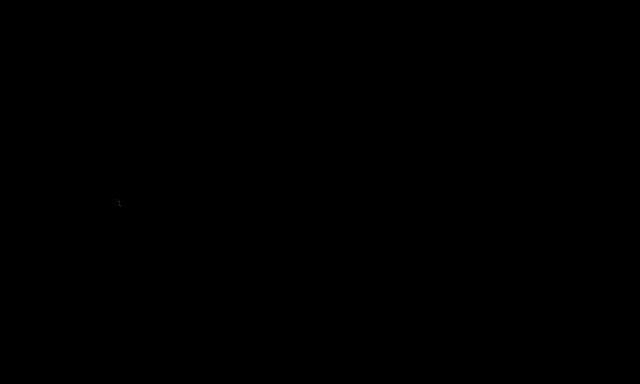}
    \caption*{Vanilla Alpha}
  \end{subfigure}\hfill
  \begin{subfigure}[b]{0.165\textwidth}
    \centering
    \includegraphics[width=\linewidth]{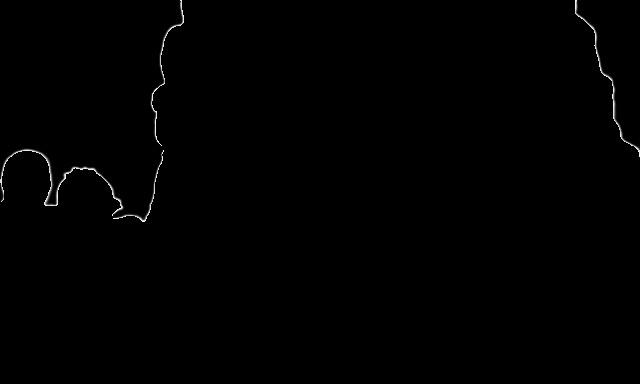}
    \caption*{\textbf{CAR (Ours)}}
  \end{subfigure}
  \caption{\textbf{Stability comparisons} between vanilla alpha representation and circular alpha representation (CAR). Vanilla alpha representation often suffers from alpha valley issues and thus produces unstable results. By contrast, our CAR shows consistent performance when processing video inputs. Regions outside $\alpha\in[0.02,0.98]$ are masked out for better comparison.}
  \label{fig:supp-stability}
\end{figure}

\begin{table*}[t]
\centering
\caption{\textbf{Ablation on the circular alpha representation for video matting} on the VideoMatte~\cite{lin2021VideoMatte} and YouTubeMatte~\cite{yang2025matanyone} datasets. \textcolor{red}{Best} results are marked.}
\label{tab:alphadepth-video-alpha}
\footnotesize
\begin{tabular}{lcccc c cccc}
\toprule[0.15em]
\rowcolor{color3} & \multicolumn{4}{c}{\textbf{VideoMatte}} & & \multicolumn{4}{c}{\textbf{YouTubeMatte}} \\
\cline{2-5} \cline{7-10}
\rowcolor{color3} \multirow{-2}{*}{\textbf{Method}} & SAD $\downarrow$ & Grad $\downarrow$ & Conn $\downarrow$ & dtSSD $\downarrow$ & & SAD $\downarrow$ & Grad $\downarrow$ & Conn $\downarrow$ & dtSSD $\downarrow$ \\ \midrule
Vanilla Alpha & 3.07 & 3.24 & 0.73 & 7.40 & & 2.01 & 4.67 & 0.95 & 8.76 \\
\rowcolor{color3} \textbf{Circular Alpha (Ours)} & {\color{red} 2.42} & {\color{red} 1.94} & {\color{red} 0.54} & {\color{red} 6.23} & & {\color{red} 1.02} & {\color{red} 1.96} & {\color{red} 0.55} & {\color{red} 6.69} \\ \bottomrule[0.15em]
\end{tabular}
\end{table*}
\subsection{Vanilla Alpha Representation \vs\ Circular Alpha Representation}
Due to the discontinuity in alpha labels (\eg, see the left part of Fig.~\ref{fig:main-circular-alpha}), learning to predict vanilla alpha representation in complex scenes often suffers from alpha valley issues (Fig.~\ref{fig:main-alpha-valley}), resulting in inaccurate alpha estimation. To verify this, we employ the \myname\ network and train a variant model for vanilla alpha prediction under the same training settings. Fig.~\ref{fig:supp-stability} compares the performance of vanilla alpha representation and our Circular Alpha Representation (CAR) under video inputs. It is evident that vanilla alpha representation often struggles at the intersections of multiple foreground targets, leading to unstable performance. In contrast, by focusing on local soft boundaries, our CAR achieves remarkable temporal consistency despite relying on an image-based model. 
We further conduct quantitative evaluation on two widely used video matting benchmarks, VideoMatte~\cite{lin2021VideoMatte} and YouTubeMatte~\cite{yang2025matanyone}, and additionally report the temporal consistency metric dtSSD~\cite{erofeev2015dtssd}. As shown in Tab.~\ref{tab:alphadepth-video-alpha}, CAR consistently outperforms vanilla alpha representation across all boundary precision and temporal consistency metrics on both benchmarks. In particular, it reduces dtSSD from 7.40 to 6.23 on VideoMatte and from 8.76 to 6.69 on YouTubeMatte. These results demonstrate that resolving the manifold discontinuity directly translates into significantly reduced temporal flicker and higher spatio-temporal boundary stability.

\begin{figure}
\centering
\pgfplotsset{
  myblue/.style={draw=blue, mark options={solid}, very thick},
  myred/.style={draw=red, mark options={solid}, very thick},
}
\begin{subfigure}[b]{0.38\linewidth}
       \begin{tikzpicture}[scale=0.92]
		\begin{axis}
			[
             height=0.75\linewidth,
             width=\linewidth,
             ymin=1, ymax=9,
             title={\textbf{VideoMatte}},
                title style={
                    at={(0.5,1.05)},
                    anchor=north,
                },
              xlabel={Training stage},
              xtick={1,2,3,4},
              xticklabels={S1-E25,S1-E50,S2-E25,S2-E50},
              scaled x ticks=false,
			 axis y line=left,
			 axis x line=bottom,
              grid,
              grid style={color=white, line width=0.5pt},
              axis background/.style={fill=color3},
              ylabel={SAD $\downarrow$},
              xlabel style={yshift=5pt},
              ylabel style={yshift=-20pt},
              line width=0.8pt,
              every axis/.append style={font=\scriptsize},
              legend style={fill=none,draw=none,font=\scriptsize, at={(0.98,0.98)},anchor=north east},
			 enlarge x limits=0.15,
              ]
			\addplot+ [myblue,mark=square] coordinates {
      (1, 5.95)
      (2, 5.55)
      (3, 3.25)
      (4, 3.07)
    };
    \addlegendentry{Vanilla alpha}

			\addplot+ [myred,mark=o] coordinates {
      (1, 2.46)
      (2, 2.49)
      (3, 2.45)
      (4, 2.42)
    };
    \addlegendentry{Circular alpha}
		\end{axis}
	\end{tikzpicture}
\end{subfigure}
\hspace{0.02\linewidth}
\begin{subfigure}[b]{0.38\linewidth}
       \begin{tikzpicture}[scale=0.92]
		\begin{axis}
			[
             height=0.75\linewidth,
             width=\linewidth,
             ymin=0, ymax=8,
             title={\textbf{YouTubeMatte}},
                title style={
                    at={(0.5,1.05)},
                    anchor=north,
                },
              xlabel={Training stage},
              xtick={1,2,3,4},
              xticklabels={S1-E25,S1-E50,S2-E25,S2-E50},
              scaled x ticks=false,
			 axis y line=left,
			 axis x line=bottom,
              grid,
              grid style={color=white, line width=0.5pt},
              axis background/.style={fill=color3},
              ylabel={SAD $\downarrow$},
              xlabel style={yshift=5pt},
              ylabel style={yshift=-20pt},
              line width=0.8pt,
              every axis/.append style={font=\scriptsize},
              legend style={fill=none,draw=none,font=\scriptsize, at={(0.98,0.98)},anchor=north east},
			 enlarge x limits=0.15,
              ]
			\addplot+ [myblue,mark=square] coordinates {
      (1, 5.15)
      (2, 4.45)
      (3, 2.28)
      (4, 2.01)
    };
    \addlegendentry{Vanilla alpha}

			\addplot+ [myred,mark=o] coordinates {
      (1, 1.14)
      (2, 1.13)
      (3, 1.04)
      (4, 1.02)
    };
    \addlegendentry{Circular alpha}
		\end{axis}
	\end{tikzpicture}
\end{subfigure}
\caption{\textbf{Training convergence comparison} of vanilla alpha and circular alpha, measured by SAD every 25 epochs at each stage under the same training setup. SX-EY indicates training stage X and epoch Y.}
\label{fig:alphadepth-training-convergence}
\end{figure}
\par 
The necessity of CAR is further highlighted by its training convergence. In Fig.~\ref{fig:alphadepth-training-convergence}, we track SAD performance on both benchmarks every 25 epochs for each training stage (SX-EY means training stage X, epoch Y), under the same training setup. By projecting the linear alpha scale onto a continuous circular manifold, CAR eliminates the non-smooth gradient landscape inherent to vanilla representations. As a result, CAR achieves accelerated convergence early in training, reaching a low SAD score of 2.46 on VideoMatte at Stage 1-Epoch 25 compared to 5.95 for vanilla alpha. Furthermore, while vanilla alpha struggles to escape local minima during early stages, CAR maintains smooth gradient flow throughout, demonstrating that resolving the alpha valley issue is critical for stable and efficient training.


\begin{figure}[t]
  \centering

  \begin{minipage}[c]{0.195\textwidth}
    \centering
    \vfill
    \includegraphics[width=\linewidth]{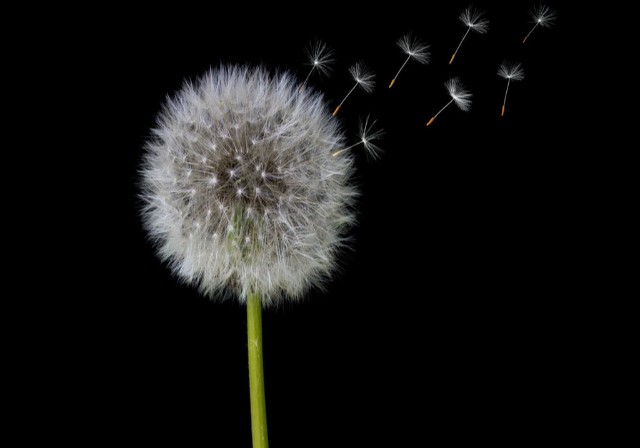}
    \caption*{Input Image}
    \vfill
  \end{minipage}\hfill
  \begin{minipage}[c]{0.8\textwidth}
    \centering
    \setlength{\tabcolsep}{1pt}
    \renewcommand{\arraystretch}{1.0}
    \begin{tabular}{@{}cccc@{}}
      \begin{minipage}[t]{0.25\textwidth}\centering
        \includegraphics[width=\linewidth]{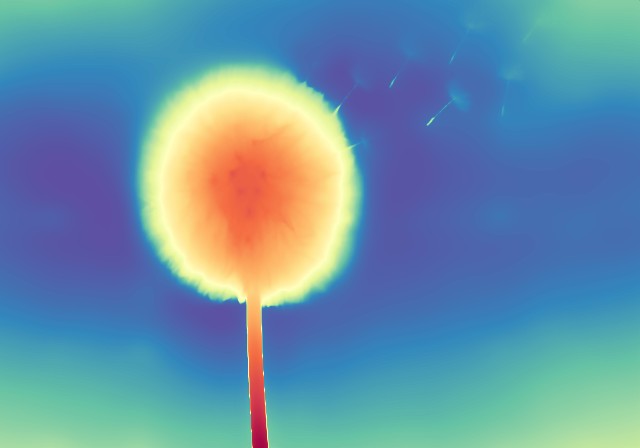}\\[-2pt]
        \caption*{\centering Depth Input\\
        (DAv2~\cite{yang2024depthanythingv2})}
      \end{minipage} &
      \begin{minipage}[t]{0.25\textwidth}\centering
        \includegraphics[width=\linewidth]{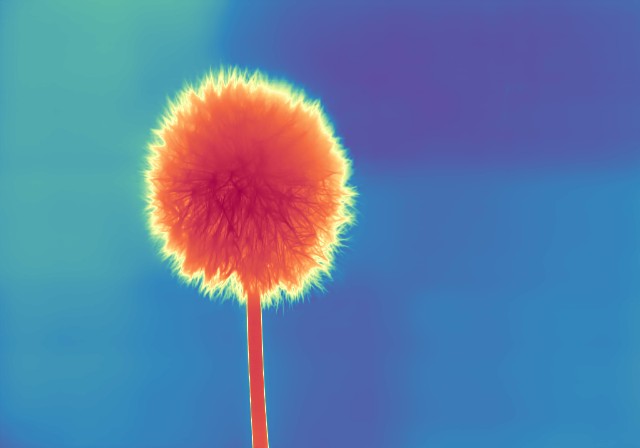}\\[-2pt]
        \caption*{\centering Depth Input\\ (DPro~\cite{depthpro})}
      \end{minipage} &
      \begin{minipage}[t]{0.25\textwidth}\centering
        \includegraphics[width=\linewidth]{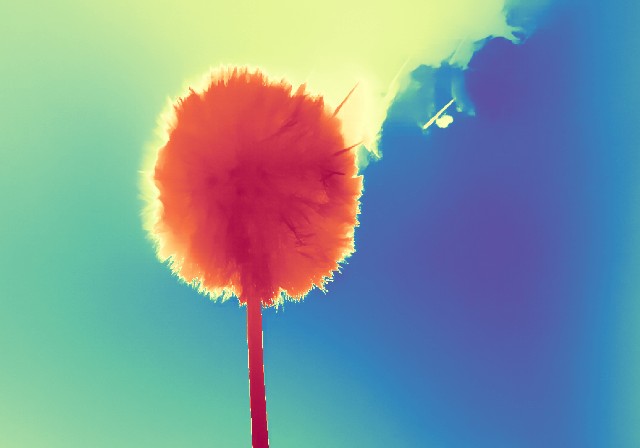}\\[-2pt]
        \caption*{\centering Depth Input\\ (PPD~\cite{xu2025ppd})}
      \end{minipage} &
      \begin{minipage}[t]{0.25\textwidth}\centering
        \includegraphics[width=\linewidth]{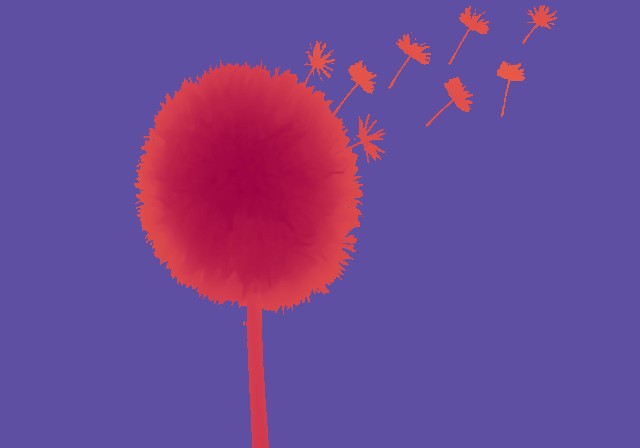}\\[-2pt]
        \caption*{\centering Depth Input\\ (MoGe-2~\cite{wang2025moge2})}
      \end{minipage} \\

      \begin{minipage}[t]{0.25\textwidth}\centering
        \includegraphics[width=\linewidth]{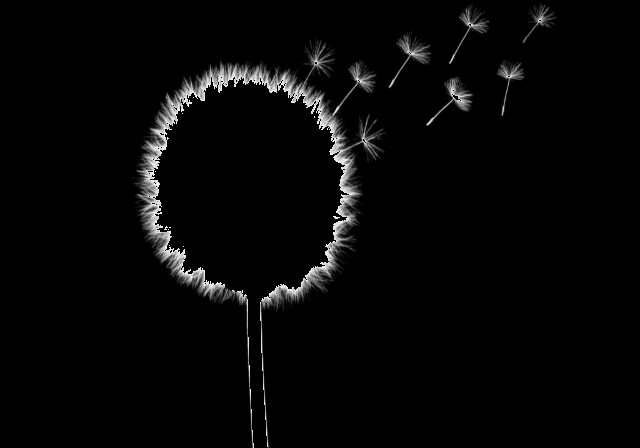}\\[-2pt]
        \caption*{\centering Alpha \\(+ DAv2~\cite{yang2024depthanythingv2})}
      \end{minipage} &
      \begin{minipage}[t]{0.25\textwidth}\centering
        \includegraphics[width=\linewidth]{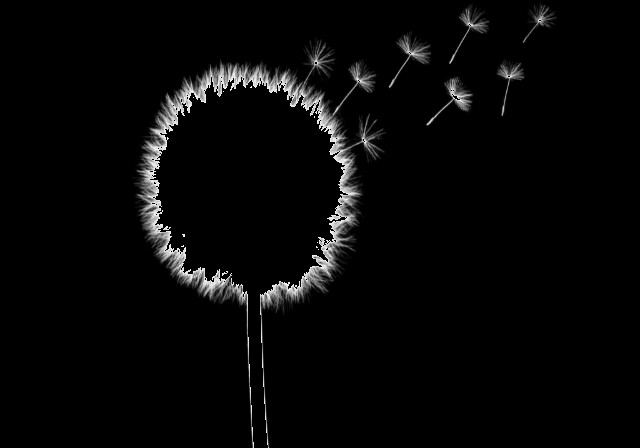}\\[-2pt]
        \caption*{\centering Alpha \\(+ DPro~\cite{depthpro})}
      \end{minipage} &
      \begin{minipage}[t]{0.25\textwidth}\centering
        \includegraphics[width=\linewidth]{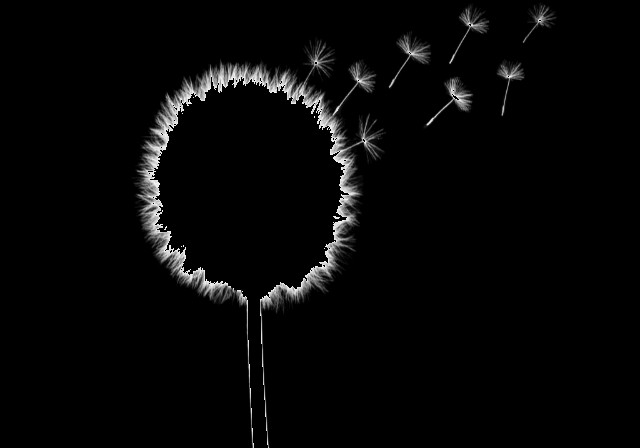}\\[-2pt]
        \caption*{\centering Alpha \\(+ PPD~\cite{xu2025ppd})}
      \end{minipage} &
      \begin{minipage}[t]{0.25\textwidth}\centering
        \includegraphics[width=\linewidth]{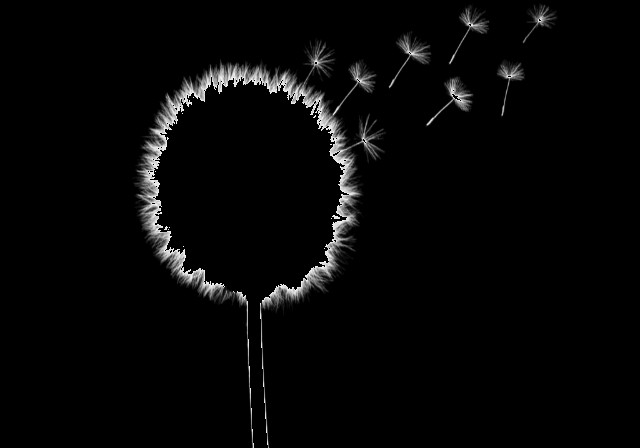}\\[-2pt]
        \caption*{\centering Alpha \\(+ MoGe-2~\cite{wang2025moge2})}
      \end{minipage}
    \end{tabular}
  \end{minipage}
\\
  \begin{minipage}[c]{0.195\textwidth}
    \centering
    \vfill
    \includegraphics[width=\linewidth]{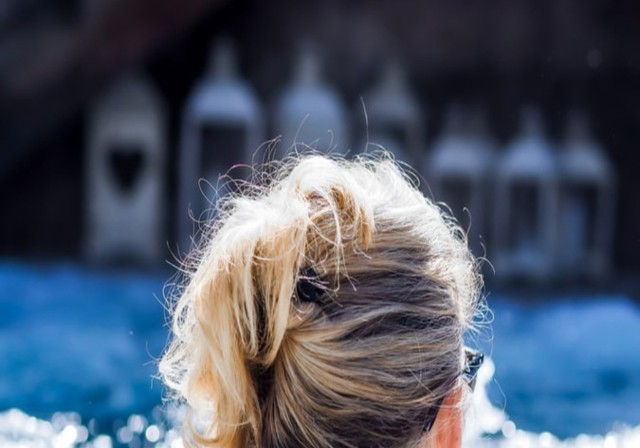}
    \caption*{Input Image}
    \vfill
  \end{minipage}\hfill
  \begin{minipage}[c]{0.8\textwidth}
    \centering
    \setlength{\tabcolsep}{1pt}
    \renewcommand{\arraystretch}{1.0}
    \begin{tabular}{@{}cccc@{}}
      \begin{minipage}[t]{0.25\textwidth}\centering
        \includegraphics[width=\linewidth]{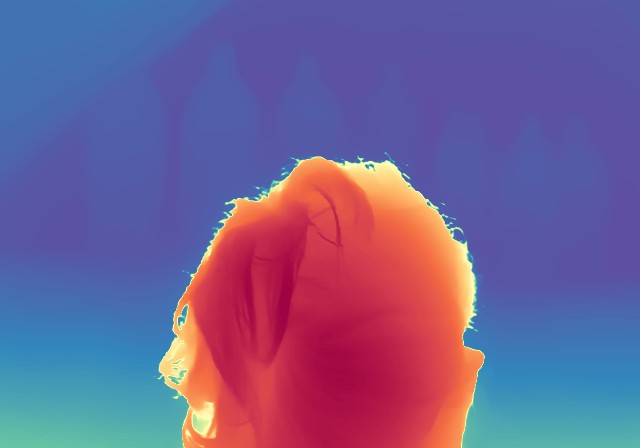}\\[-2pt]
        \caption*{\centering Depth Input\\
        (DAv2~\cite{yang2024depthanythingv2})}
      \end{minipage} &
      \begin{minipage}[t]{0.25\textwidth}\centering
        \includegraphics[width=\linewidth]{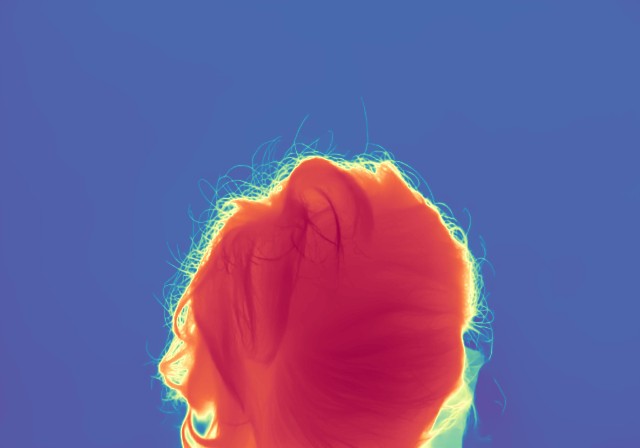}\\[-2pt]
        \caption*{\centering Depth Input\\ (DPro~\cite{depthpro})}
      \end{minipage} &
      \begin{minipage}[t]{0.25\textwidth}\centering
        \includegraphics[width=\linewidth]{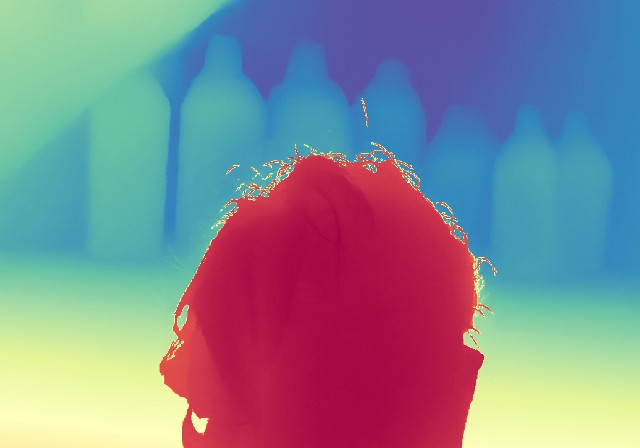}\\[-2pt]
        \caption*{\centering Depth Input\\ (PPD~\cite{xu2025ppd})}
      \end{minipage} &
      \begin{minipage}[t]{0.25\textwidth}\centering
        \includegraphics[width=\linewidth]{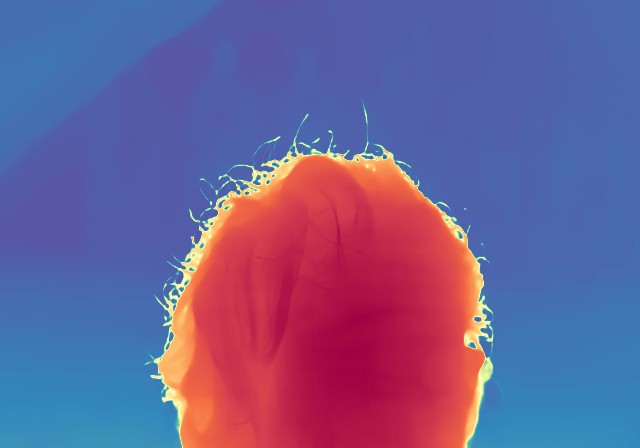}\\[-2pt]
        \caption*{\centering Depth Input\\ (MoGe-2~\cite{wang2025moge2})}
      \end{minipage} \\

      \begin{minipage}[t]{0.25\textwidth}\centering
        \includegraphics[width=\linewidth]{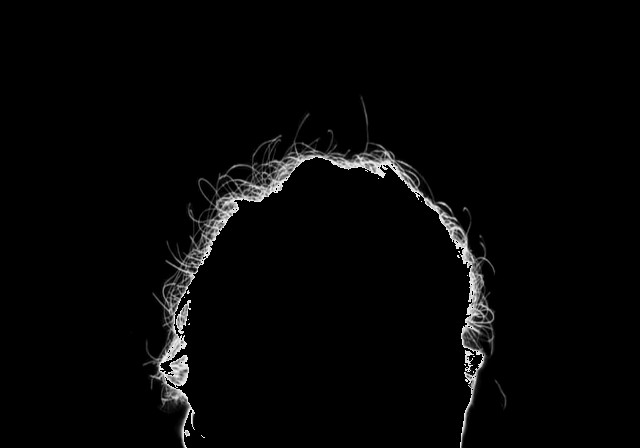}\\[-2pt]
        \caption*{\centering Alpha \\(+ DAv2~\cite{yang2024depthanythingv2})}
      \end{minipage} &
      \begin{minipage}[t]{0.25\textwidth}\centering
        \includegraphics[width=\linewidth]{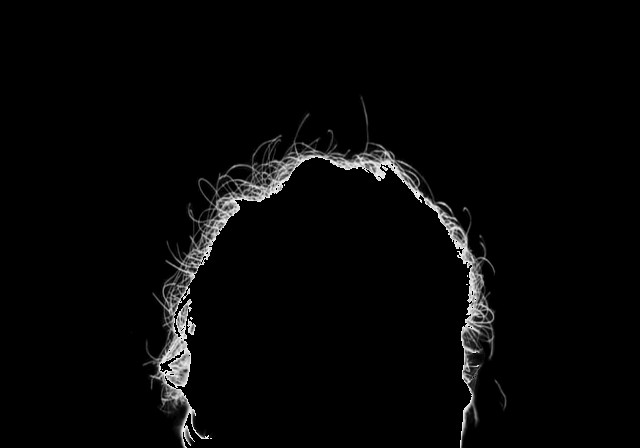}\\[-2pt]
        \caption*{\centering Alpha \\(+ DPro~\cite{depthpro})}
      \end{minipage} &
      \begin{minipage}[t]{0.25\textwidth}\centering
        \includegraphics[width=\linewidth]{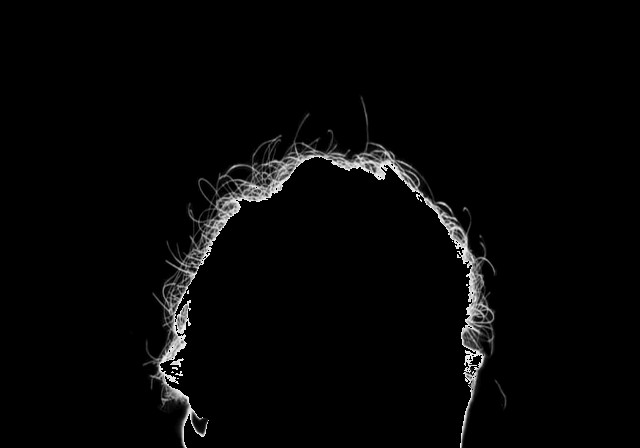}\\[-2pt]
        \caption*{\centering Alpha \\(+ PPD~\cite{xu2025ppd})}
      \end{minipage} &
      \begin{minipage}[t]{0.25\textwidth}\centering
        \includegraphics[width=\linewidth]{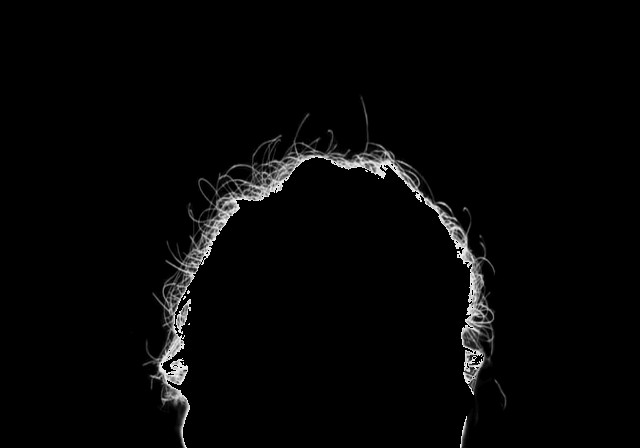}\\[-2pt]
        \caption*{\centering Alpha \\(+ MoGe-2~\cite{wang2025moge2})}
      \end{minipage}
    \end{tabular}
  \end{minipage}

  \caption{\textbf{Alpha estimation performance} under different depth inputs. We generate input depth using state-of-the-art models, including Depth Anything V2 (DAv2)~\cite{yang2024depthanythingv2}, Depth Pro (DPro)~\cite{depthpro}, Pixel-Perfect Depth (PPD)~\cite{xu2025ppd}, and MoGe-2~\cite{wang2025moge2}. Despite different characteristics exhibited in depth inputs, our \myname\ shows stable performance in alpha estimation and soft boundary detail extraction.}
  \label{fig:supp-depth-impact}
\end{figure}

\begin{table*}[t]
\centering
\caption{\textbf{Matting performance of \myname} with different depth models. The \textcolor{red}{best} and \textcolor{blue}{second best} results are marked.}
\vspace{-0.5em}
\label{tab:supp-matting-diff-depth}
\footnotesize
\setlength\tabcolsep{4pt}
\begin{tabular}{lccc c ccc}
\toprule[0.15em]
\rowcolor{color3} & \multicolumn{3}{c}{\textbf{AIM-500}} & & \multicolumn{3}{c}{\textbf{P3M-10K}} \\
\cline{2-4} \cline{6-8}
\rowcolor{color3} \multirow{-2}{*}{\textbf{Depth Model}} & SAD $\downarrow$  & Grad $\downarrow$ & Conn $\downarrow$ & & SAD $\downarrow$ & Grad $\downarrow$ & Conn $\downarrow$ \\ \midrule
Depth Anything V2~\cite{yang2024depthanythingv2} & {\color{blue} 7.24} & {\color{red} 20.20} & {\color{blue} 3.89} & & {\color{red} 4.09} & {\color{red} 11.88} & {\color{red} 2.31} \\
Depth Pro~\cite{depthpro} & {\color{red} 7.22} & {\color{blue} 20.35} & {\color{red} 3.88} & & {\color{blue} 4.15} & {\color{blue} 12.20} & {\color{blue} 2.34} \\
MoGe-2~\cite{wang2025moge2} & 7.34 & 21.09 & 3.95 & & 4.19 & 12.47 & 2.36 \\
Pixel-Perfect Depth~\cite{xu2025ppd} & 7.37 & 20.76 & 3.95 & & 4.18 & 12.24 & 2.36 \\
\bottomrule[0.15em]
\end{tabular}

\end{table*}

\begin{table}[t]
\centering
\caption{\textbf{Robustness to temporally inconsistent depth} on the VideoMatte~\cite{lin2021VideoMatte} and YouTubeMatte~\cite{yang2025matanyone} datasets. We compare video-consistent depth (VDA~\protect\cite{chen2025videodepthanything}) against per-frame depth (DAv2~\protect\cite{yang2024depthanythingv2}) as input to $\alpha$Depth. The temporal alpha metric dtSSD~\cite{erofeev2015dtssd} is additionally adopted for evaluation, and \textcolor{red}{best} results are marked.}
\label{tab:alphadepth-depth-robustness-video}
\footnotesize
\begin{tabular}{lcccc c cccc}
\toprule[0.15em]
\rowcolor{color3} & \multicolumn{4}{c}{\textbf{VideoMatte}} & & \multicolumn{4}{c}{\textbf{YouTubeMatte}} \\ \cline{2-5} \cline{7-10}
\rowcolor{color3} \multirow{-2}{*}{\textbf{Depth Input}} & SAD $\downarrow$ & Grad $\downarrow$ & Conn $\downarrow$ & dtSSD $\downarrow$ & & SAD $\downarrow$ & Grad $\downarrow$ & Conn $\downarrow$ & dtSSD $\downarrow$ \\ \midrule
Video Depth (VDA) & 2.42 & {\color{red} 1.94} & 0.54 & 6.23 & & 1.02 & {\color{red} 1.96} & 0.55 & 6.69 \\
Image Depth (DAv2) & {\color{red} 2.41} & 1.98 & {\color{red} 0.51} & {\color{red} 6.09} & & {\color{red} 1.01} & 1.99 & {\color{red} 0.54} & {\color{red} 6.61} \\ \bottomrule[0.15em]
\end{tabular}
\end{table}

\subsection{Matting Performance under Different Depth Inputs}

Depth estimation performance is essential for high-quality stereo conversion because it directly influences geometry and parallax of the synthesized views. Since $\alpha$Depth focuses on soft boundary decomposition while preserving the original geometry and texture of opaque regions, it can integrate seamlessly with state-of-the-art depth estimators in a plug-and-play manner. To evaluate the robustness of \myname\ against various depth inputs, we conduct experiments on both image and video matting benchmarks, and our multimodal design remains robust to depth errors across spatial and temporal dimensions.
\textbf{(i) Spatial Robustness}: As shown in Tab.~\ref{tab:supp-matting-diff-depth}, $\alpha$Depth achieves consistent performance across four different depth backbones. Fig.~\ref{fig:supp-depth-impact} (top rows) highlights three characteristic failure modes: over-smoothed discontinuities from Depth Anything V2 (DAv2), missing fine structures from Depth Pro (DPro), and incorrect depth ordering from Pixel-Perfect Depth (PPD). In all cases, $\alpha$Depth yields consistent decompositions and captures soft boundary details even where the depth inputs fail (\eg, the fine dandelions in Fig.~\ref{fig:supp-depth-impact}). This is because our dual-path encoder utilizes image priors to compensate for unreliable depth cues.
\textbf{(ii) Temporal Robustness}: Tab.~\ref{tab:alphadepth-depth-robustness-video} compares per-frame depth from DAv2 against temporally consistent depth from VDA on two video matting datasets VideoMatte~\cite{lin2021VideoMatte} and YouTubeMatte~\cite{yang2025matanyone}, and the performance using per-frame depth matches that of video-consistent depth across all metrics. This confirms that temporal depth flicker does not propagate into our predictions, as $\alpha$Depth anchors its estimation on the temporally stable visual sequence.

\usetikzlibrary{calc,spy}

\def\suppSpySize{1cm}

\def\suppSpyMagInput{4}
\def\suppSpyMagFone{4}
\def\suppSpyMagFtwo{4}
\def\suppSpyMagFthree{4}

\def\suppSpyXInput{0.65}
\def\suppSpyYInput{0.82}

\def\suppSpyXFone{0.59}
\def\suppSpyYFone{0.82}

\def\suppSpyXFtwo{0.55}
\def\suppSpyYFtwo{0.82}

\def\suppSpyXFthree{0.5}
\def\suppSpyYFthree{0.82}

\def\suppSpyShiftX{-0.55cm}
\def\suppSpyShiftY{0.55cm}

\newcommand{\suppimgwithspy}[4]{%
  \begin{tikzpicture}[spy using outlines={green,magnification=#2,size=\suppSpySize},inner sep=0]
    \node (img) {\includegraphics[width=\linewidth]{#1}};
    \coordinate (bot) at ($(img.south west)!#3!(img.south east)$);
    \coordinate (top) at ($(img.north west)!#3!(img.north east)$);
    \coordinate (spyp) at ($(bot)!#4!(top)$);
    \spy on (spyp) in node at ($(img.south east)+(\suppSpyShiftX,\suppSpyShiftY)$);
  \end{tikzpicture}%
}

\begin{figure*}[t]
  \centering

  \begin{minipage}[c]{0.248\textwidth}
    \centering
    \begin{subfigure}[t]{\linewidth}
      \centering
      \suppimgwithspy{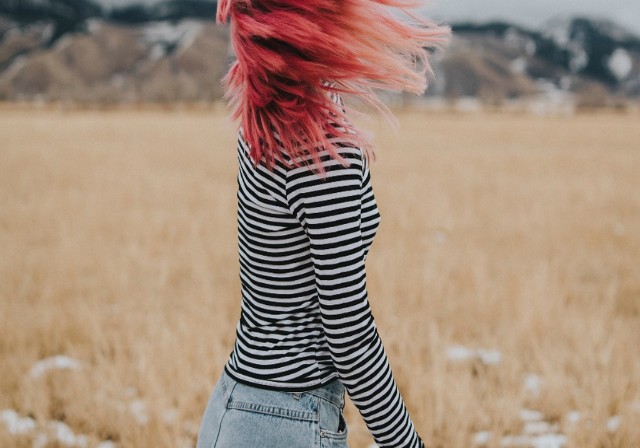}{\suppSpyMagInput}{\suppSpyXInput}{\suppSpyYInput}
      \caption*{Input Image}
    \end{subfigure}\\[4pt]
    \begin{subfigure}[t]{\linewidth}
      \centering
      \includegraphics[width=\linewidth]{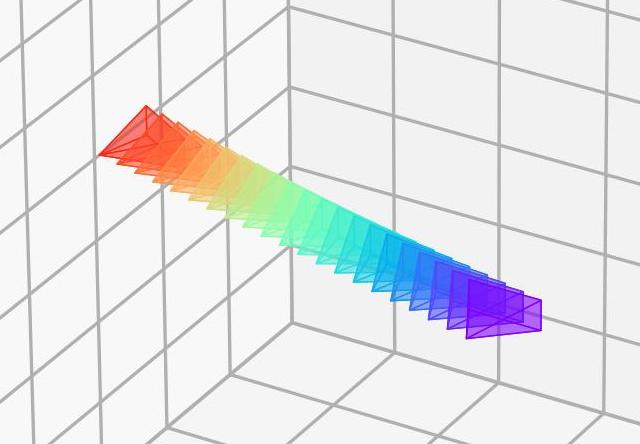}
      \caption*{Camera Motion}
    \end{subfigure}
  \end{minipage}%
  \hfill
  \begin{minipage}[c]{0.75\textwidth}
    \centering
    \renewcommand{\arraystretch}{1.0}

    \begin{subfigure}[t]{0.32\linewidth}
      \centering
      \suppimgwithspy{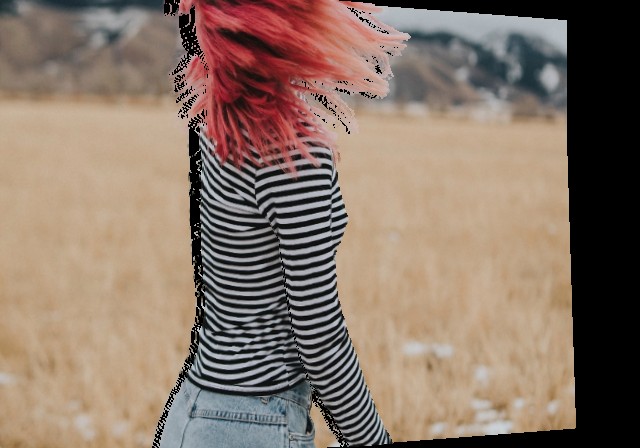}{\suppSpyMagFone}{\suppSpyXFone}{\suppSpyYFone}
      \caption*{\centering Frame\#1 \\ (Original~\cite{yang2024depthanythingv2})}
    \end{subfigure}
    \begin{subfigure}[t]{0.32\linewidth}
      \centering
      \suppimgwithspy{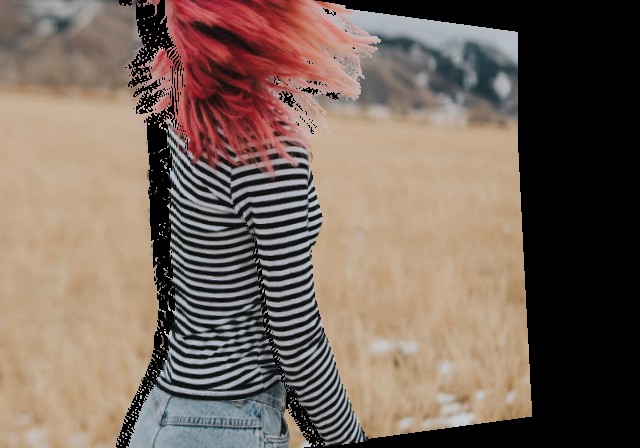}{\suppSpyMagFtwo}{\suppSpyXFtwo}{\suppSpyYFtwo}
      \caption*{\centering Frame\#2 \\ (Original~\cite{yang2024depthanythingv2})}
    \end{subfigure}
    \begin{subfigure}[t]{0.32\linewidth}
      \centering
      \suppimgwithspy{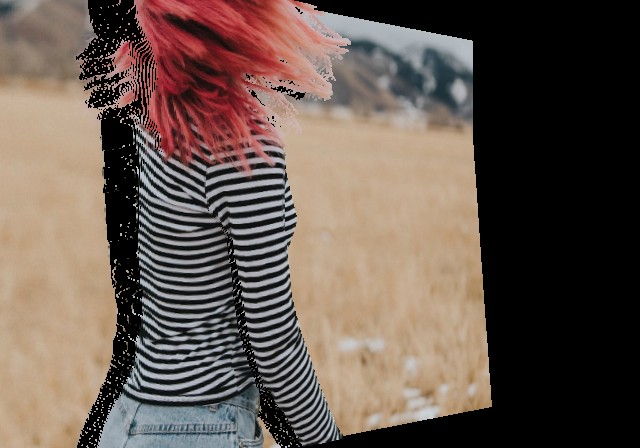}{\suppSpyMagFthree}{\suppSpyXFthree}{\suppSpyYFthree}
      \caption*{\centering Frame\#3 \\ (Original~\cite{yang2024depthanythingv2})}
    \end{subfigure}\\[6pt]

    \begin{subfigure}[t]{0.32\linewidth}
      \centering
      \suppimgwithspy{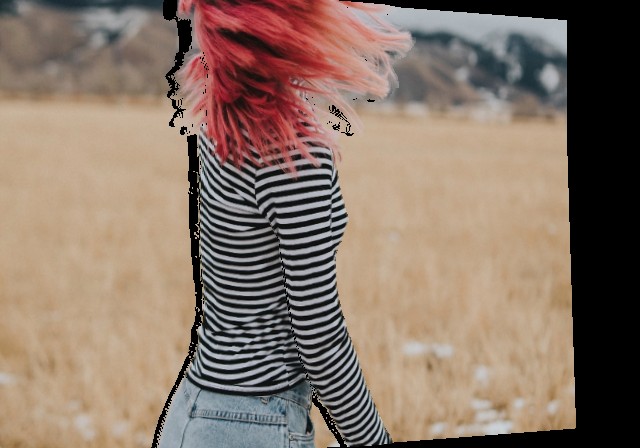}{\suppSpyMagFone}{\suppSpyXFone}{\suppSpyYFone}
      \caption*{\centering Frame\#1 \\ (+ HairGuard~\cite{hairguard})}
    \end{subfigure}
    \begin{subfigure}[t]{0.32\linewidth}
      \centering
      \suppimgwithspy{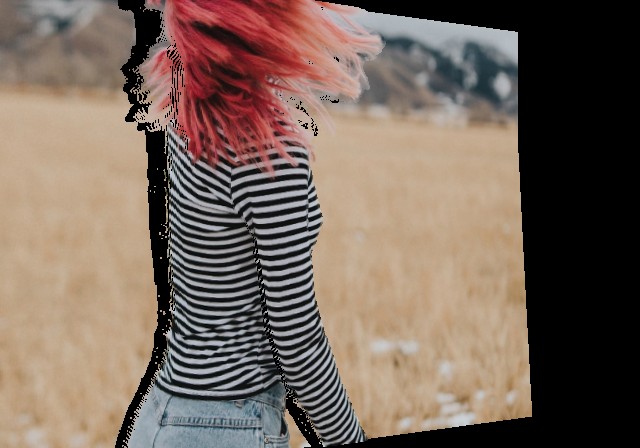}{\suppSpyMagFtwo}{\suppSpyXFtwo}{\suppSpyYFtwo}
      \caption*{\centering Frame\#2 \\ (+ HairGuard~\cite{hairguard})}
    \end{subfigure}
    \begin{subfigure}[t]{0.32\linewidth}
      \centering
      \suppimgwithspy{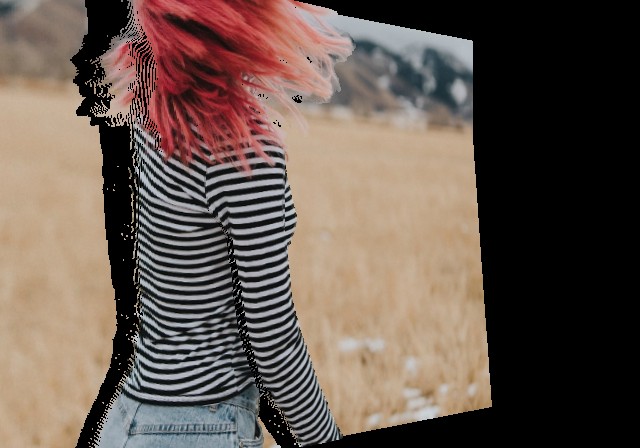}{\suppSpyMagFthree}{\suppSpyXFthree}{\suppSpyYFthree}
      \caption*{\centering Frame\#3 \\ (+ HairGuard~\cite{hairguard})}
    \end{subfigure}\\[6pt]

    \begin{subfigure}[t]{0.32\linewidth}
      \centering
      \suppimgwithspy{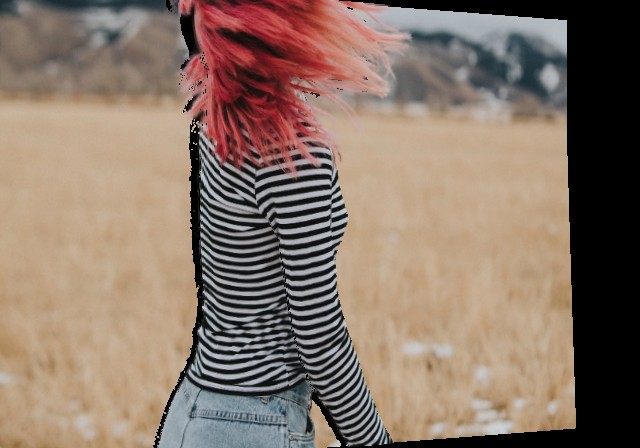}{\suppSpyMagFone}{\suppSpyXFone}{\suppSpyYFone}
      \caption*{\centering Frame\#1 \\(+ Ours)}
    \end{subfigure}
    \begin{subfigure}[t]{0.32\linewidth}
      \centering
      \suppimgwithspy{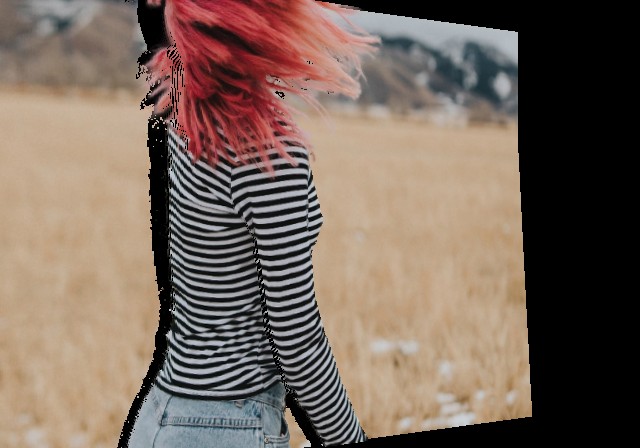}{\suppSpyMagFtwo}{\suppSpyXFtwo}{\suppSpyYFtwo}
      \caption*{\centering Frame\#2 \\(+ Ours)}
    \end{subfigure}
    \begin{subfigure}[t]{0.32\linewidth}
      \centering
      \suppimgwithspy{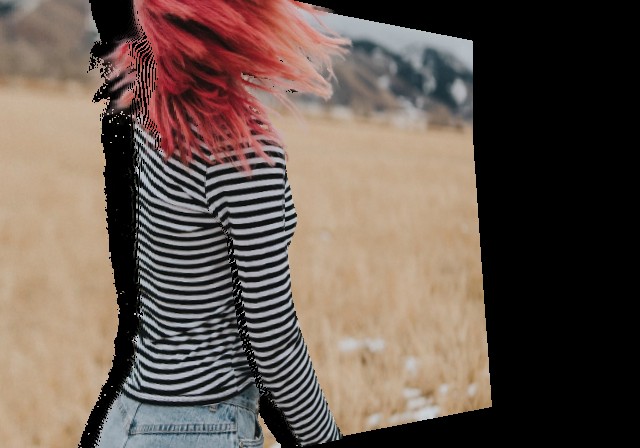}{\suppSpyMagFthree}{\suppSpyXFthree}{\suppSpyYFthree}
      \caption*{\centering Frame\#3 \\(+ Ours)}
    \end{subfigure}
  \end{minipage}

  \caption{\textbf{Warping performance under large viewpoint changes}. This example employs the camera motion (arc left with rotation) from ReCamMaster~\cite{recammaster}. Due to depth ambiguity in soft boundary regions, the warping results using the original depth from Depth Anything V2~\cite{yang2024depthanythingv2} often contain broken structures. Although HairGuard refines depth to better preserve soft boundary details~\cite{hairguard}, its results often suffer from background bleeding. The proposed \myname\ achieves the best warping performance with high-fidelity soft boundary details. }
  \label{fig:supp-cam-traj}
\end{figure*}

\begin{table*}[th]
\centering
\caption{\textbf{Warping performance (FID $\downarrow$)} under different camera trajectories on natural image matting datasets. We apply horizontal swing motion to simulate different baseline lengths in stereo conversion. We also employ 10 different camera trajectories from the evaluation protocol of ReCamMaster~\cite{recammaster} to compare the warping performance under larger and more flexible viewpoint changes. The \textcolor{red}{best} and \textcolor{blue}{second best} results are marked. }
\label{tab:supp-diff-cam-motion}
\footnotesize
\setlength\tabcolsep{4pt}

\begin{tabular}{lcc c cc}
\toprule[0.15em]
\rowcolor{color3} & \multicolumn{2}{c}{\textbf{AIM-500}} & & \multicolumn{2}{c}{\textbf{P3M-10K}} \\
\cline{2-3} \cline{5-6}
\rowcolor{color3} \multirow{-2}{*}{\textbf{Method}} & Horizontal Swing & ReCamMaster Motion & & Horizontal Swing & ReCamMaster Motion \\ \midrule
DAv2~\cite{yang2024depthanythingv2} & 41.75 & 36.99 & & 49.50 & 48.47 \\
DAv2 + HairGuard~\cite{hairguard}  & {\color{blue} 38.49} & {\color{blue} 36.00} & & {\color{blue} 45.71} & {\color{blue} 47.17} \\
\rowcolor{color3} \textbf{DAv2 + \myname\ (Ours)} & {\color{red} 34.43} & {\color{red} 34.26} & & {\color{red} 35.47} & {\color{red} 43.58} \\
\bottomrule[0.15em]
\end{tabular}

\end{table*}

\subsection{Performance under Different Camera Trajectories}
We test the performance of our \myname\ under different camera trajectories with larger viewpoint changes. Two categories of camera motions are employed: (i) We first apply a horizontal swing motion to simulate different baseline lengths in stereo conversion settings. (ii) For more flexible camera motion, we follow ReCamMaster~\cite{recammaster} to test the robustness of our \myname\ method with 10 types of different trajectories. Tab.~\ref{tab:supp-diff-cam-motion} verifies the state-of-the-art performance of \myname\ under different camera motions. Existing depth models often struggle at soft boundaries due to depth ambiguity, leading to broken structures at soft boundary regions. Although the recent approach HairGuard~\cite{hairguard} refines depth to capture soft boundary details, the improved warping still suffers from background bleeding artifacts, as illustrated in Fig.~\ref{fig:supp-cam-traj}. By decomposing soft boundaries via \myname, our method achieves the best performance in soft boundary preservation.


\begin{table*}[t]
\centering
\caption{\textbf{Number of parameters for each component in \myname\ network.}}
\vspace{-0.5em}
\label{tab:supp-parameters}
\footnotesize
\begin{tabular}{cc c ccc c c}
\toprule[0.15em]
\rowcolor{color3} \multicolumn{2}{c}{\textbf{Dual-Path Encoder}} & & \multicolumn{3}{c}{\textbf{Multi-Branch Decoder}} & & \\
\cline{1-2} \cline{4-6}
\rowcolor{color3} Detail Encoder & Semantic Encoder & & Alpha Decoder & Color Decoder & Depth Decoder & & \multirow{-2}{*}{\textbf{Total}}  \\ \midrule
1.17 M & 319.46 M & & 2.65 M & 2.65 M & 2.65 M & & 328.58 M \\
\bottomrule[0.15em]
\end{tabular}

\end{table*}

\subsection{Computational Complexity}
Tab.~\ref{tab:supp-parameters} breaks down the number of parameters for each component of our \myname\ network. Additionally, for an input size of $448\times640$, the model requires 1.94 GB of peak GPU memory and achieves an inference speed of 0.0153 seconds per image on an NVIDIA GeForce RTX 4090 GPU.

\section{More Ablation Studies}\label{supp-sec:more-ablation}

\subsection{Impact of Multi-Branch Decoder}
As shown in Fig.~\ref{fig:main-network}, we employ a multi-branch decoder to estimate different modalities, \ie, alpha, depth, and color, in our \myname\ representation. This benefits the network by explicitly decoupling the distinct structural and textural characteristics of each modality, preventing feature interference during task-specific predictions. To verify this, we train an additional variant to directly estimate \myname\ representation via a unified decoder. As demonstrated in Tab.~\ref{tab:supp-multi-decoder}, removing the multi-branch decoder leads to a consistent drop in alpha matting performance across all metrics. Specifically, the full model with the multi-branch decoder improves the SAD metric from 8.22 to 7.24 on the AIM-500 dataset~\cite{aim_500}, and from 4.36 to 4.09 on the P3M-10K dataset~\cite{li2021p3m_10k}, validating its effectiveness in accurately extracting soft boundary details.

\begin{table*}[t]
\centering
\caption{\textbf{Ablation of multi-branch decoder} on matting datasets. The \textcolor{red}{best} results are marked.}
\vspace{-0.5em}
\label{tab:supp-multi-decoder}
\footnotesize
\begin{tabular}{lccc c ccc}
\toprule[0.15em]
\rowcolor{color3} & \multicolumn{3}{c}{\textbf{AIM-500}} & & \multicolumn{3}{c}{\textbf{P3M-10K}} \\
\cline{2-4} \cline{6-8}
\rowcolor{color3} \multirow{-2}{*}{\textbf{Method}} & SAD $\downarrow$  & Grad $\downarrow$ & Conn $\downarrow$ & & SAD $\downarrow$ & Grad $\downarrow$ & Conn $\downarrow$ \\ \midrule
w/o Multi-Branch Decoder & 8.22 & 25.95 & 4.55 & & 4.36 & 14.05 & 2.51 \\ 
\rowcolor{color3} \textbf{w/ Multi-Branch Decoder (Ours)} & {\color{red} 7.24} & {\color{red} 20.20} & {\color{red} 3.89} & & {\color{red} 4.09} & {\color{red} 11.88} & {\color{red} 2.31} \\
\bottomrule[0.15em]
\end{tabular}

\end{table*}

\subsection{Impact of Semantic Encoder}
In the proposed \myname\ network, we employ a semantic encoder to extract high-level semantics and a detail encoder to capture soft boundary details (Fig.~\ref{fig:main-network}). The semantic encoder leverages pre-trained image priors to extract deep semantic features, which are essential for high-level soft boundary reasoning. Tab.~\ref{tab:supp-edge-and-encoder} presents the ablation study of the semantic encoder on the Marvel-10K dataset~\cite{hairguard}, following the same experimental setups in Tab.~\ref{tab:main-ablation-warping}. Removing the semantic encoder results in a noticeable performance drop in warping performance, with S-PSNR decreasing from 28.68 to 27.35 and S-SSIM dropping from 0.7636 to 0.7147. This degradation indicates that high-level contextual cues and semantic understanding are vital for the network to resolve depth ambiguity and correctly decompose soft boundaries in complex scenes.

\begin{table*}[t]
\centering
\caption{\textbf{Ablation of edge extraction and semantic encoder} on Marvel-10K~\cite{hairguard}. The \textcolor{red}{best} and \textcolor{blue}{second best} results are marked.}
\vspace{-0.5em}
\label{tab:supp-edge-and-encoder}
\footnotesize
\begin{tabular}{lcccc}
\toprule[0.15em]
\rowcolor{color3} & \multicolumn{4}{c}{\textbf{Marvel-10K}} \\
\cline{2-5}
\rowcolor{color3} \multirow{-2}{*}{\textbf{Method}} & S-PSNR $\uparrow$ & S-SSIM $\uparrow$ & LPIPS $\downarrow$ & DISTS $\downarrow$ \\ \midrule
w/o Edge Extraction & 26.14 & 0.6718 & 0.1337 & 0.0910 \\
w/o Semantic Encoder & {\color{blue} 27.35} & {\color{blue} 0.7147} & {\color{blue} 0.1269} & {\color{blue} 0.0839} \\
\rowcolor{color3} \textbf{Full Model (Ours)} & {\color{red} 28.68} & {\color{red} 0.7636} & {\color{red} 0.1190} & {\color{red} 0.0738} \\
\bottomrule[0.15em]
\end{tabular}

\end{table*}

\subsection{Impact of Depth Edge Extraction}
Depth edge extraction provides explicit geometric priors by extracting depth gradients from the input depth map, which serve as strong cues for soft boundary localization (Fig.~\ref{fig:main-network}). We empirically found that depth edges are crucial for the convergence of the network during training. Removing this component can lead to training instability or even divergence. This is likely because depth edges provide a strong initialization for soft boundary localization and thus alleviate the difficulty of multi-task learning. As shown in Tab.~\ref{tab:supp-edge-and-encoder}, omitting the edge extraction module severely impacts the warping performance, leading to a 2.54 dB drop in S-PSNR. This emphasizes the necessity of explicit boundary cues in guiding the network to focus on soft boundary regions.

\section{Limitations and Discussions}\label{supp-sec:limitation}

Although \myname\ effectively resolves soft boundaries in stereo conversion, some limitations remain:
\par 
\textbf{Dependence on Initial Depth Maps}. Our method adopts a plug-and-play design that can be integrated with various state-of-the-art monocular depth estimation models. Although $\alpha$Depth effectively decomposes local foreground and background at soft boundary regions, the global scene structure and scale remain heavily dependent on the quality of the initial input depth map. If the underlying depth model fails in extreme scenarios (\eg, severe geometric distortions), our approach may struggle to fully correct the errors in the base geometry. 
To address this issue, future research could explore end-to-end joint optimization strategies by integrating our module with foundation depth models. This would allow the explicit boundary priors extracted by \myname\ to back-propagate and iteratively correct global geometric distortions. Alternatively, introducing a confidence-aware fusion mechanism could enable the network to selectively rely on deep image semantics when the input depth exhibits low reliability. 

\par 
\textbf{Two-Layer Representation}. 
Our approach models local soft boundaries based on the observation that a two-layer decomposition (foreground and background) is generally sufficient to resolve local occlusions.
Thus, phenomena that are locally two-layer, including hair, fur, thin structures, and defocus or motion blur, can be well approximated by this formulation (see Fig.~\ref{fig:supp-alphadepth-vis}), as the alpha value corresponds to the point spread of defocus, the temporal coverage of motion, or the material transparency of the foreground layer. However, our approach may fail in cases that violate the local two-layer assumption. For example, volumetric media accumulate opacity continuously along the viewing ray and exhibit no well-defined boundary region. In addition, in some complex scenes where multiple semi-transparent boundaries overlap at the same pixel (\ie, three or more overlapping layers, such as the intersecting points in the top example of Fig.~\ref{fig:supp-alphadepth-vis}), the current two-layer model might not fully capture all layered information. In future work, extending our framework to support an arbitrary number of overlapping layers presents an exciting research avenue. This could be achieved through an iterative layer-peeling mechanism, where foreground layers are sequentially stripped away~\cite{wen2026seegroup}. Furthermore, integrating local volumetric representations (\eg, localized radiance fields~\cite{mildenhall2020nerf}) at intersecting boundaries could model complex semi-transparencies.

\par 
\textbf{Video Consistency}. Although the proposed Circular Alpha Representation (CAR) demonstrates remarkable temporal consistency when processing video inputs (\eg, see Fig.~\ref{fig:supp-stability}), the current \myname\ framework operates fundamentally as an image-based model. Due to the lack of explicit temporal constraints, \myname\ may produce results with flickering artifacts, especially in dynamic scenes where depth changes rapidly. 
Specifically, because \myname\ prioritizes capturing soft boundaries at depth discontinuities, it may fail to resolve boundaries at low depth gradients (\eg, when two targets move close together in depth).
Future extensions of this work could explore integrating spatio-temporal modules, \eg, spatio-temporal attention, to further stabilize the layered predictions across video sequences.

\section{Discussion of Societal Impacts}\label{sec:socialimpacts}
Our work on high-fidelity stereo conversion presents several positive societal impacts, primarily by democratizing the creation of immersive 3D content. By automating soft boundary decomposition in complex scenes, our \myname\ framework significantly lowers the barrier for creators in the Virtual and Augmented Reality (VR/AR), immersive education, and entertainment industries. This enables the efficient and low-cost transformation of legacy monocular media into engaging 3D experiences. However, we also acknowledge potential negative impacts associated with this technology. The ability to synthesize highly realistic stereo views could be misused to create immersive 3D disinformation, making fabricated content appear more physically credible. Additionally, the unauthorized stereo conversion of individuals or private scenes from casually captured 2D photos could raise privacy concerns.

\section{Visualization of Ablation Models}\label{supp-sec:vis-ablation}

Fig.~\ref{fig:supp-ablation} provides qualitative results for the ablation models detailed in Tab.~\ref{tab:main-ablation}. Due to depth ambiguity at soft boundaries, the baseline model (A\#1) often generates warped images exhibiting broken boundaries and flying pixels. Although incorporating alpha estimation improves the structure of soft boundaries, the alpha valley issue inherent to the vanilla alpha representation tends to degrade warping performance (\eg, see the bottom region of A\#2 in Fig.~\ref{fig:supp-ablation}). By contrast, the proposed circular alpha representation circumvents this issue, effectively preserving image structures and soft boundary details (A\#3). Finally, when combined with the layered representation, our \myname\ faithfully recovers background information at soft boundaries, achieving the best warping performance (A\#4).

\par 
Furthermore, the bottom row of Fig.~\ref{fig:supp-ablation} illustrates the qualitative impacts of the different alpha matting strategies evaluated in Tab.~\ref{tab:main-ablation-alpha}. When employing the vanilla alpha representation (B\#1), the network suffers from the alpha valley issue, which leads to unstable predictions and noticeable artifacts when extracting complex structures. Relying solely on the $\mathcal{L}_1$ loss without the matting loss $\mathcal{L}_{\mathrm{m}}$ (B\#2) fails to adequately capture fine-grained details, resulting in blurred and degraded soft boundaries. Similarly, utilizing single-stage training without the subsequent global refinement (B\#3) yields noisy alpha predictions. By contrast, our full model (B\#4) effectively overcomes these limitations and extracts intricate soft boundary details.

\input{figs/supp/fig-supp-ablation-vis}

\section{Visualization of \myname\ Results}\label{supp-sec:vis-alpha-depth}

\begin{figure*}[t]
  \centering

  \begin{subfigure}[t]{0.247\textwidth}
    \centering
    \includegraphics[width=\linewidth]{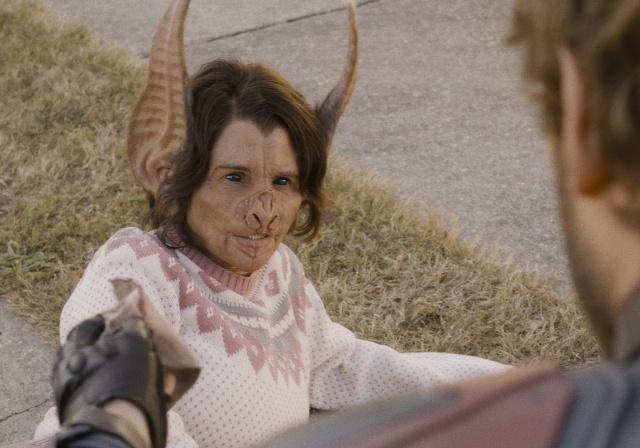}
    \caption*{Input Image $I_{\mathrm{IN}}$}
  \end{subfigure}\hfill
  \begin{subfigure}[t]{0.247\textwidth}
    \centering
    \includegraphics[width=\linewidth]{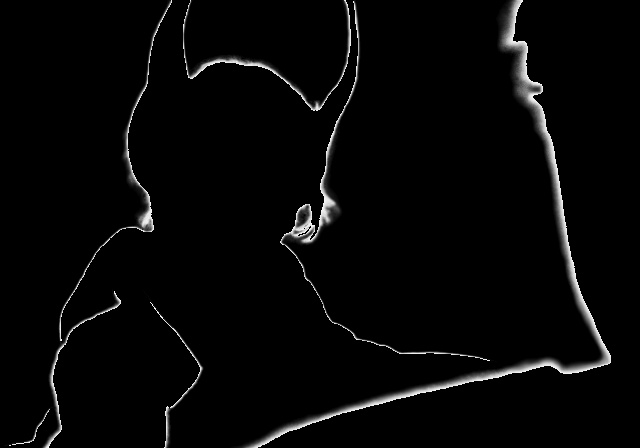}
    \caption*{Alpha $\hat{\alpha}$}
  \end{subfigure}\hfill
  \begin{subfigure}[t]{0.247\textwidth}
    \centering
    \includegraphics[width=\linewidth]{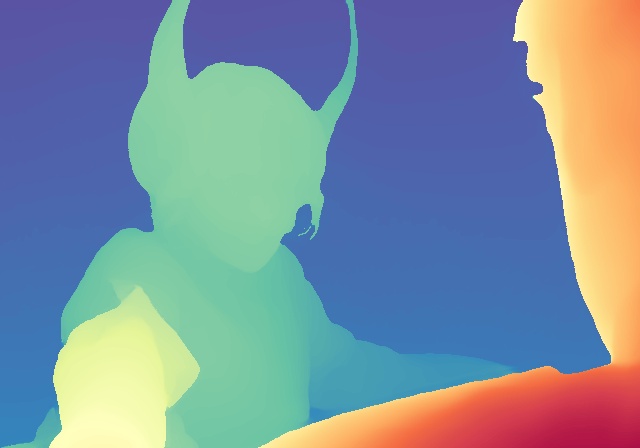}
    \caption*{Background Depth $\hat{D}_{\mathrm{BG}}$}
  \end{subfigure}\hfill
  \begin{subfigure}[t]{0.247\textwidth}
    \centering
    \includegraphics[width=\linewidth]{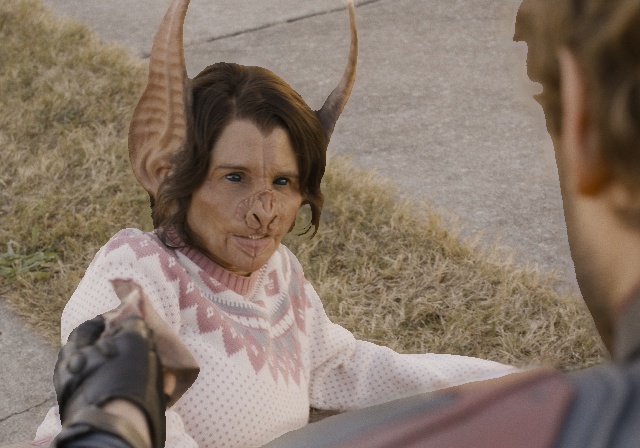}
    \caption*{Background Color $\hat{I}_{\mathrm{BG}}$}
  \end{subfigure}

  \begin{subfigure}[t]{0.247\textwidth}
    \centering
    \includegraphics[width=\linewidth]{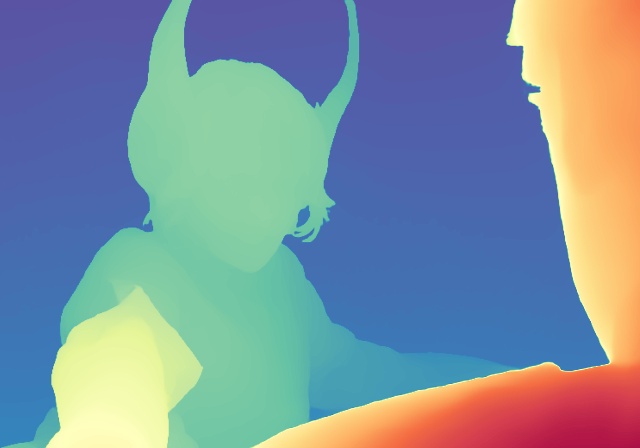}
    \caption*{Input Depth $D_\mathrm{IN}$}
  \end{subfigure}\hfill
  \begin{subfigure}[t]{0.247\textwidth}
    \centering
    \includegraphics[width=\linewidth]{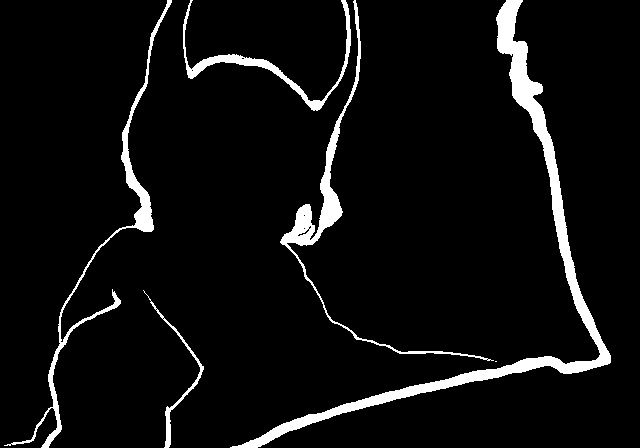}
    \caption*{Soft Boundary Region $\hat{S}$}
  \end{subfigure}\hfill
  \begin{subfigure}[t]{0.247\textwidth}
    \centering
    \includegraphics[width=\linewidth]{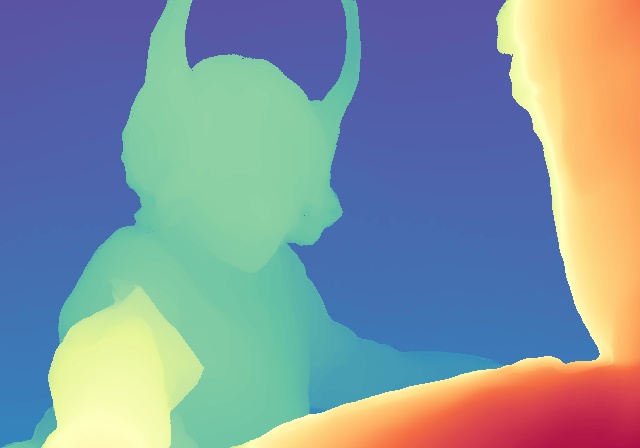}
    \caption*{Foreground Depth $\hat{D}_\mathrm{FG}$}
  \end{subfigure}\hfill
  \begin{subfigure}[t]{0.247\textwidth}
    \centering
    \includegraphics[width=\linewidth]{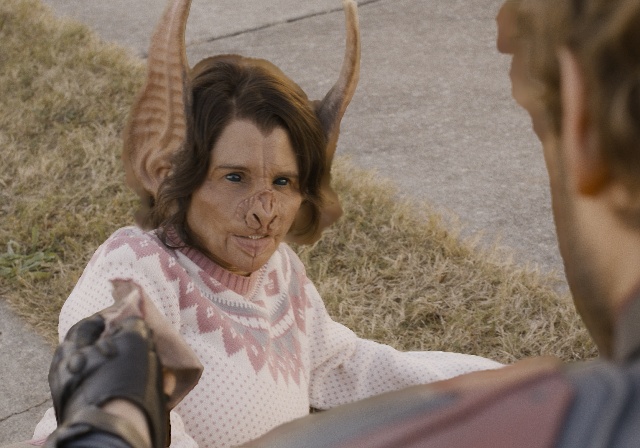}
    \caption*{Foreground Color $\hat{I}_\mathrm{FG}$}
  \end{subfigure}


  \begin{subfigure}[t]{0.247\textwidth}
    \centering
    \includegraphics[width=\linewidth]{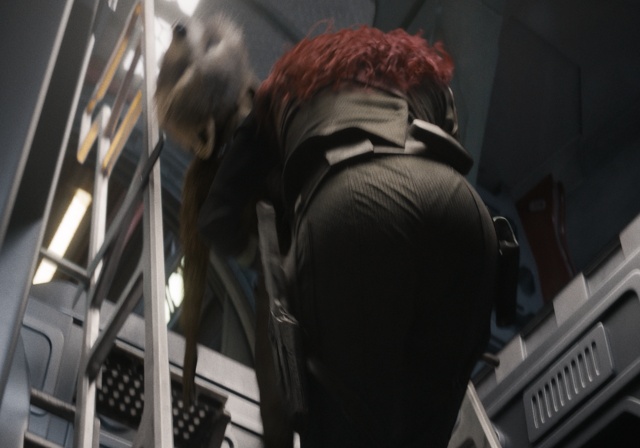}
    \caption*{Input Image $I_{\mathrm{IN}}$}
  \end{subfigure}\hfill
  \begin{subfigure}[t]{0.247\textwidth}
    \centering
    \includegraphics[width=\linewidth]{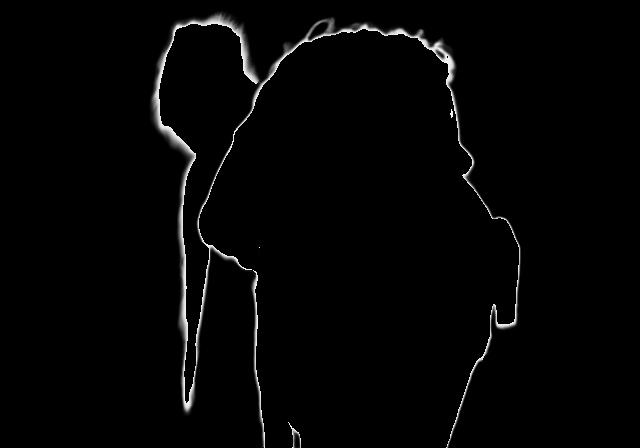}
    \caption*{Alpha $\hat{\alpha}$}
  \end{subfigure}\hfill
  \begin{subfigure}[t]{0.247\textwidth}
    \centering
    \includegraphics[width=\linewidth]{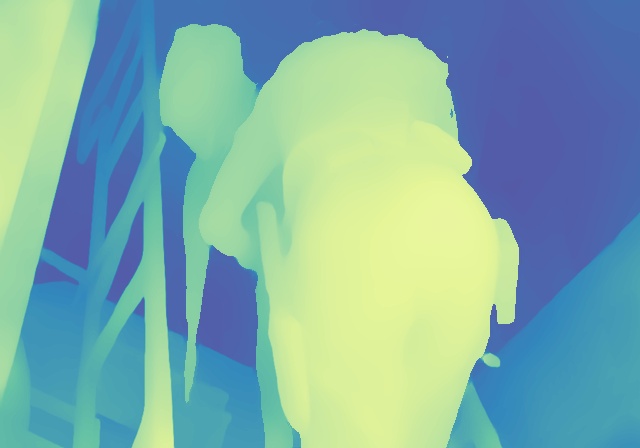}
    \caption*{Background Depth $\hat{D}_{\mathrm{BG}}$}
  \end{subfigure}\hfill
  \begin{subfigure}[t]{0.247\textwidth}
    \centering
    \includegraphics[width=\linewidth]{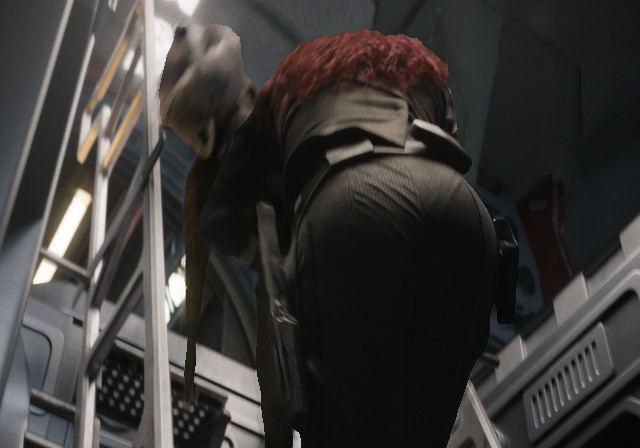}
    \caption*{Background Color $\hat{I}_{\mathrm{BG}}$}
  \end{subfigure}

  \begin{subfigure}[t]{0.247\textwidth}
    \centering
    \includegraphics[width=\linewidth]{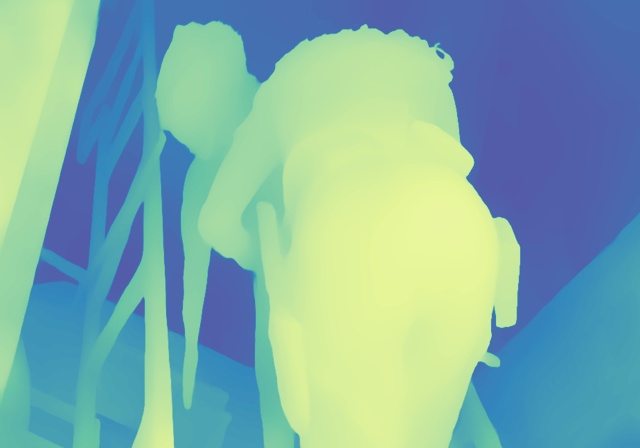}
    \caption*{Input Depth $D_\mathrm{IN}$}
  \end{subfigure}\hfill
  \begin{subfigure}[t]{0.247\textwidth}
    \centering
    \includegraphics[width=\linewidth]{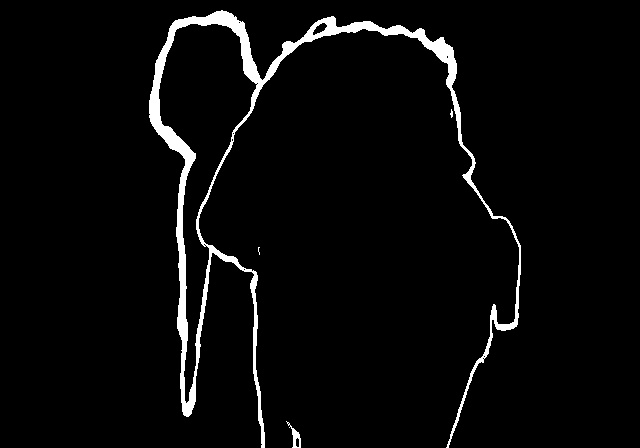}
    \caption*{Soft Boundary Region $\hat{S}$}
  \end{subfigure}\hfill
  \begin{subfigure}[t]{0.247\textwidth}
    \centering
    \includegraphics[width=\linewidth]{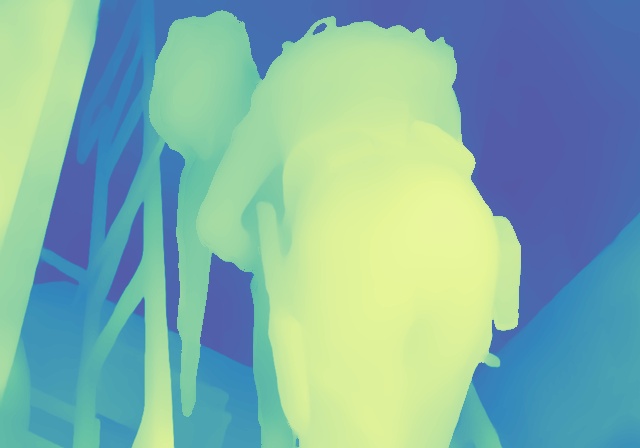}
    \caption*{Foreground Depth $\hat{D}_\mathrm{FG}$}
  \end{subfigure}\hfill
  \begin{subfigure}[t]{0.247\textwidth}
    \centering
    \includegraphics[width=\linewidth]{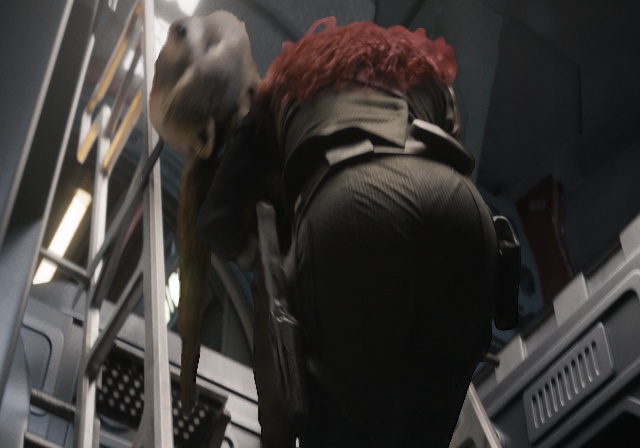}
    \caption*{Foreground Color $\hat{I}_\mathrm{FG}$}
  \end{subfigure}


  \begin{subfigure}[t]{0.247\textwidth}
    \centering
    \includegraphics[width=\linewidth]{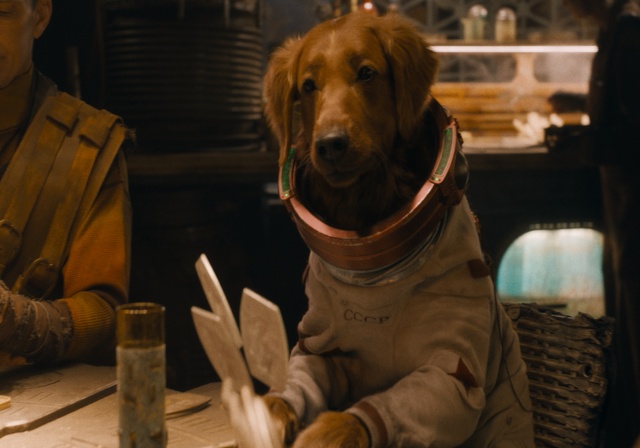}
    \caption*{Input Image $I_{\mathrm{IN}}$}
  \end{subfigure}\hfill
  \begin{subfigure}[t]{0.247\textwidth}
    \centering
    \includegraphics[width=\linewidth]{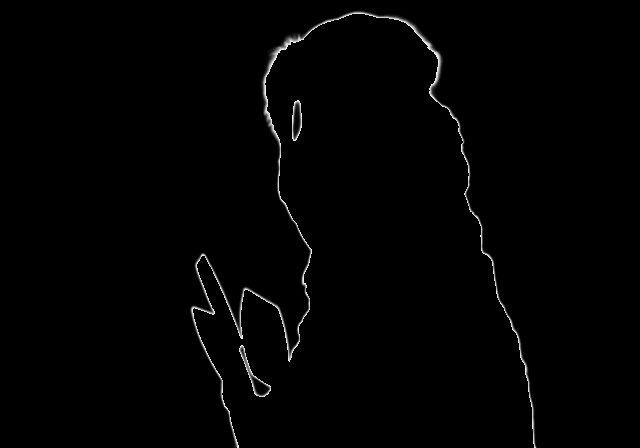}
    \caption*{Alpha $\hat{\alpha}$}
  \end{subfigure}\hfill
  \begin{subfigure}[t]{0.247\textwidth}
    \centering
    \includegraphics[width=\linewidth]{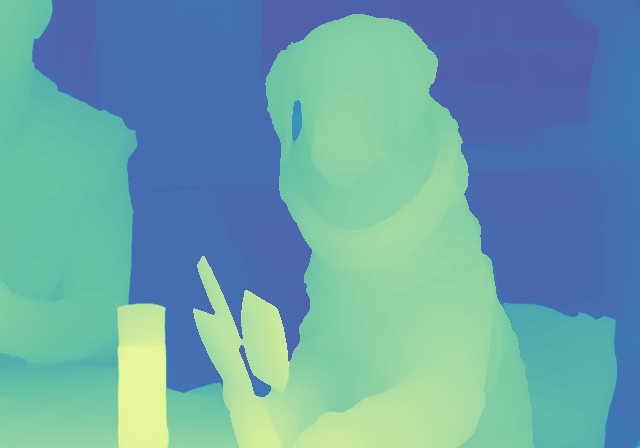}
    \caption*{Background Depth $\hat{D}_{\mathrm{BG}}$}
  \end{subfigure}\hfill
  \begin{subfigure}[t]{0.247\textwidth}
    \centering
    \includegraphics[width=\linewidth]{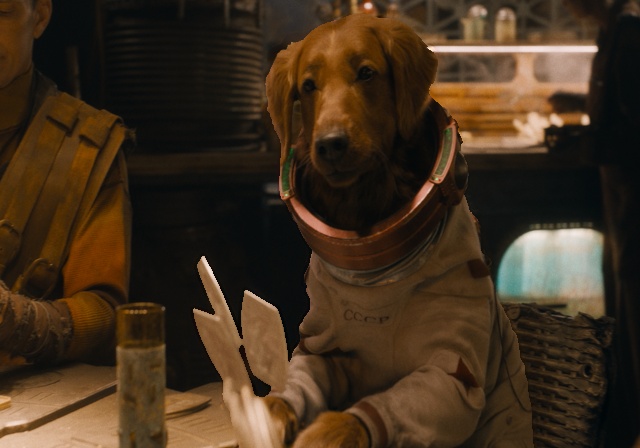}
    \caption*{Background Color $\hat{I}_{\mathrm{BG}}$}
  \end{subfigure}

  \begin{subfigure}[t]{0.247\textwidth}
    \centering
    \includegraphics[width=\linewidth]{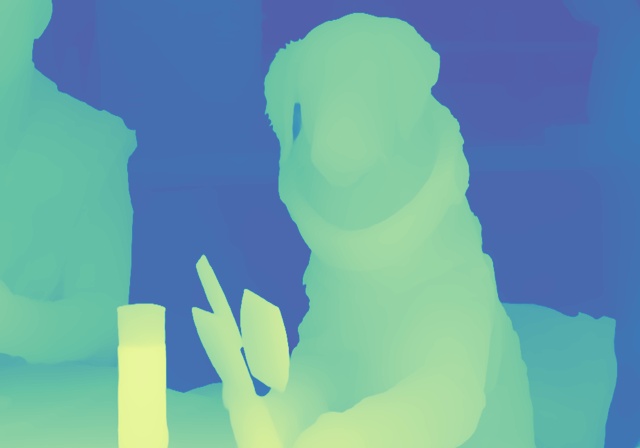}
    \caption*{Input Depth $D_\mathrm{IN}$}
  \end{subfigure}\hfill
  \begin{subfigure}[t]{0.247\textwidth}
    \centering
    \includegraphics[width=\linewidth]{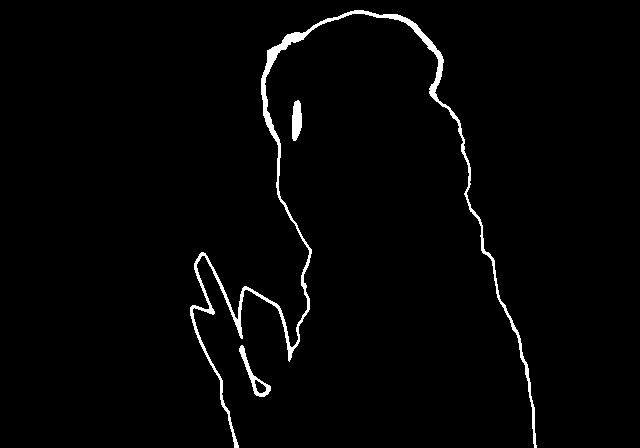}
    \caption*{Soft Boundary Region $\hat{S}$}
  \end{subfigure}\hfill
  \begin{subfigure}[t]{0.247\textwidth}
    \centering
    \includegraphics[width=\linewidth]{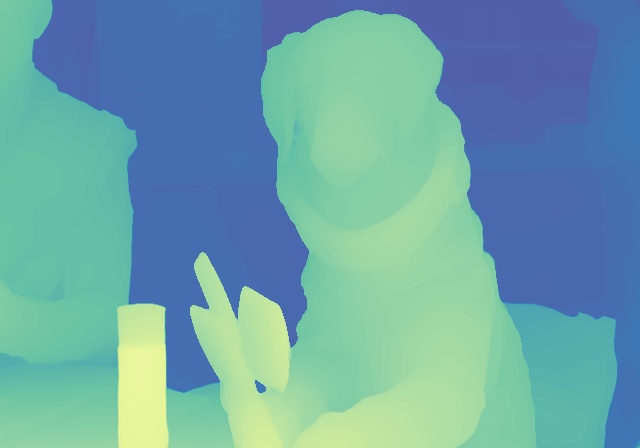}
    \caption*{Foreground Depth $\hat{D}_\mathrm{FG}$}
  \end{subfigure}\hfill
  \begin{subfigure}[t]{0.247\textwidth}
    \centering
    \includegraphics[width=\linewidth]{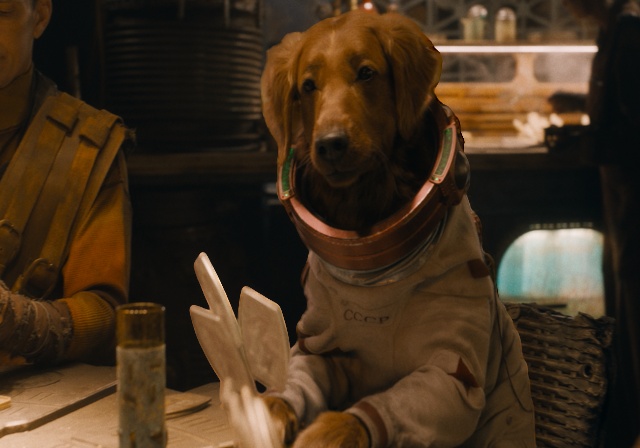}
    \caption*{Foreground Color $\hat{I}_\mathrm{FG}$}
  \end{subfigure}

  \caption{\textbf{Visualization of \myname\ results} on Marvel-10K dataset~\cite{hairguard}.}
  \label{fig:supp-alphadepth-vis}
\end{figure*}

Fig.~\ref{fig:supp-alphadepth-vis} illustrates the results estimated by \myname\ in typical stereo conversion scenarios. Our method demonstrates robust performance in challenging conditions, such as multiple targets with defocus blur (top example) and highly dynamic scenes with motion blur (middle example), highlighting its practical value for real-world applications.


\section{More Visual Comparisons}\label{supp-sec:more-visual}

\subsection{Stereo Conversion}
In Figs.~\ref{fig:supp-additional-stereo} and \ref{fig:supp-additional-stereo2}, we present additional qualitative comparisons evaluating warping and stereo conversion performance. When utilizing original depth maps estimated from Video Depth Anything~\cite{chen2025videodepthanything}, direct view transformation techniques struggle with the depth ambiguity at soft boundaries, frequently resulting in broken edges and flying pixels. While recent refinement approaches like HairGuard~\cite{hairguard} capture finer boundary details, their reliance on a single-layer depth representation still leads to visible background bleeding and aliasing artifacts during warping. Furthermore, when comparing final stereo conversion results, existing state-of-the-art methods often struggle to maintain structural consistency at these intricate boundaries. In contrast, our \myname\ framework explicitly addresses these limitations by employing a layered representation that decouples soft boundaries into local foreground and background. By disentangling the mixed colors and resolving depth ambiguities, our method achieves superior fidelity in warping and stereo conversion results. 

\input{figs/supp/fig-supp-stereo-conv}
\input{figs/supp/fig-supp-stereo-conv2}

\subsection{Alpha Matting}

Fig.~\ref{fig:supp-alpha} provides further visual comparisons of our Circular Alpha Representation (CAR) against state-of-the-art alpha matting baselines, including GVM~\cite{ge2025GVM}, MatAnyone 2~\cite{yang2025matanyone2}, and ViTMatte~\cite{yao2024vitmatte}. Conventional matting techniques generally rely on explicit global definitions of foreground and background, necessitating manual guidance such as user-provided trimaps (\eg, ViTMatte~\cite{yao2024vitmatte}) or segmentation masks (\eg, MatAnyone 2~\cite{yang2025matanyone2}) for instance-level inference. While auxiliary-free methods like GVM~\cite{ge2025GVM} reduce user effort, they are typically optimized for specific semantic categories and struggle to generalize to the diverse types of soft boundaries in complex scenes. As demonstrated in Fig.~\ref{fig:supp-alpha}, our approach achieves comparable performance in extracting intricate soft boundary details with state-of-the-art matting methods, without any user intervention. 

\input{figs/supp/fig-supp-alpha}







\end{document}